\pdfoutput=1
\documentclass[11pt]{article}
\usepackage[final]{acl}
\usepackage{makecell}
\usepackage{amsmath}
\usepackage{mdframed}
\usepackage{times}
\usepackage{latexsym}
\usepackage[T1]{fontenc}
\usepackage[utf8]{inputenc}
\usepackage{linguex}
\usepackage{graphicx}
\usepackage{tabularx}
\usepackage{xcolor}
\usepackage{booktabs}

\title{Language Models Largely Exhibit\\Human-like Constituent Ordering Preferences}

\author{
  \textbf{Ada Defne Tur\textsuperscript{$\dagger\ddagger$*}},
  \textbf{Gaurav Kamath\textsuperscript{$\dagger\ddagger$*}},
  \textbf{Siva Reddy\textsuperscript{$\dagger\ddagger\mathsection$}}
\\
\\
  \textsuperscript{$\dagger$}McGill University, Canada,
  \textsuperscript{$\ddagger$}Mila - Quebec AI Institute, Canada, \\
  \textsuperscript{$\mathsection$}Canada CIFAR AI Chair\\
  \textsuperscript{*}Equal Contribution
\\
  \small{
    \texttt{Correspondences to \href{mailto:ada.tur@mila.quebec}{ada.tur@mila.quebec}}
  }
}

\begin{document}
\maketitle

\begin{abstract}
Though English sentences are typically inflexible vis-à-vis word order, constituents often show far more variability in ordering. One prominent theory presents the notion that constituent ordering is directly correlated with constituent weight: a measure of the constituent's length or complexity. Such theories are interesting in the context of natural language processing (NLP), because while recent advances in NLP have led to significant gains in the performance of large language models (LLMs), much remains unclear about how these models process language, and how this compares to human language processing. In particular, the question remains whether LLMs display the same patterns with constituent movement, and may provide insights into existing theories on when and how the shift occurs in human language. We compare a variety of LLMs with diverse properties to evaluate broad LLM performance on four types of constituent movement: heavy NP shift, particle movement, dative alternation, and multiple PPs. Despite performing unexpectedly around particle movement, LLMs generally align with human preferences around constituent ordering.\footnote{All code and data is available at \href{https://github.com/McGill-NLP/Constituent-Movement}{https://github.com/McGill-NLP/Constituent-Movement}.}

\end{abstract}

\section{Introduction}
Despite the fact that word order in English is typically strict, constituents in post-verbal positions can be highly flexible in their ordering \cite{chomsky2002syntactic, Wasow_Arnold_2003}. A number of specific phenomena are prime examples of this movement; we show these in Figure~\ref{fig:shiftexs}.

\begin{figure}[ht]
\textbf{Heavy NP Shift (HNPS)}
\ex. \label{ex:heavy-np-shift}
\a. I met [the tall man selling water to marathon runners]$_\text{NP}$ [at the park]$_\text{PP}$. \label{ex:1a}
\b. I met [at the park]$_\text{PP}$ [the tall man selling water to marathon runners]$_\text{NP}$. \label{ex:1b}

\textbf{Particle Movement (PM)}
\ex. \label{ex:particle-movement}
\a. She looked [up]$_\text{particle}$ [her question]$_\text{NP}$ on her computer. \label{ex:2a}
\b. She looked [her question]$_\text{NP}$ [up]$_\text{particle}$ on her computer. \label{ex:2b}

\textbf{Dative Alternation (DA)}
\ex. \label{ex:dative-alternation}
\a. He sent [her]$_\text{IndirectObj}$ [a gift]$_\text{DirectObj}$ for her birthday. \label{ex:3a}
\b. He sent [a gift]$_\text{DirectObj}$ [to her]$_\text{IndirectObj}$ for her birthday. \label{ex:3b}

\textbf{Multiple PP Shift (MPP)}
\ex. \label{ex:multiple-pp}
\a. I went [to the mall]$_\text{PP}$ [with my sister]$_\text{PP}$ on Sunday. \label{ex:4a}
\b. I went [with my sister]$_\text{PP}$ [to the mall]$_\text{PP}$ on Sunday. \label{ex:4b}

\caption{Examples of constituent movement types: Heavy NP Shift (HNPS), Particle Movement (PM), Dative Alternation (DA) and Multiple PP Shift (MPP).}

\label{fig:shiftexs}
\end{figure}

\begin{figure*}[ht]
    \centering
    \includegraphics[width=\textwidth]{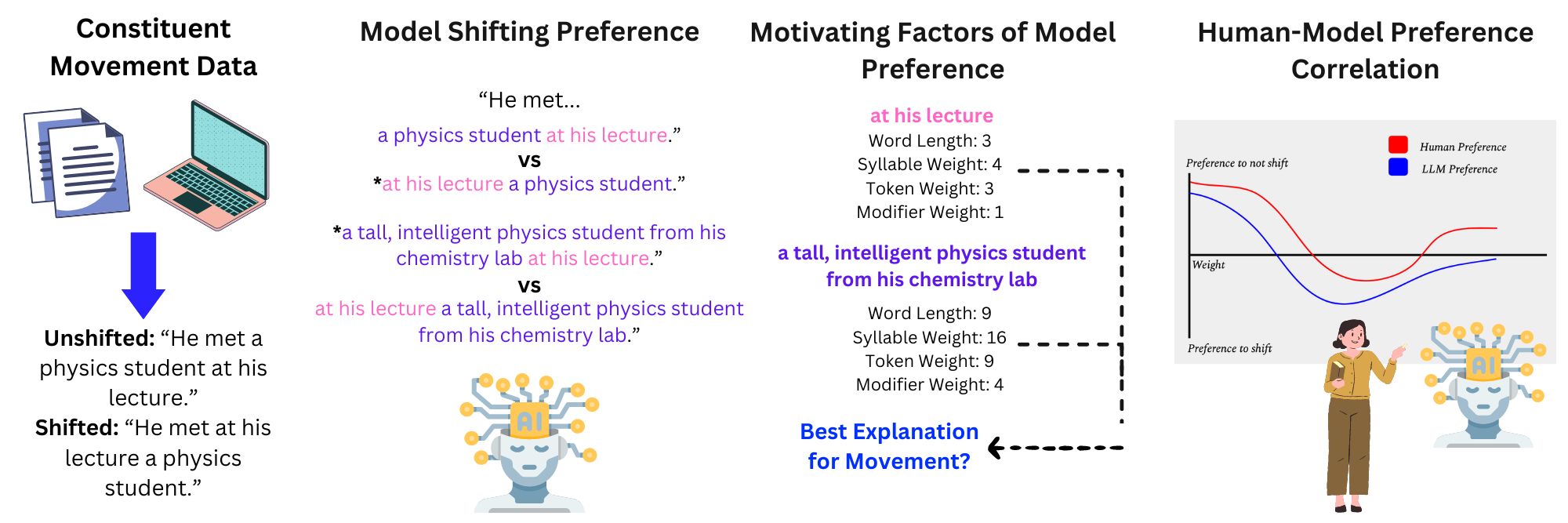}
    \caption{We categorize our work into three main experiments. Our first experiment evaluates model response to constituent movement, our second experiment analyzes what motivates LLM constituent ordering preferences, and our third experiment compares model preferences with human judgements.}
    \label{fig:overallFig}
\end{figure*}

The cause of this movement has been attributed to a variety of factors, including lexical bias \cite{baars1975output, hartsuiker2005lexical, dell1981stages}, semantic connectedness \cite{kayne1983, behaghel1932}, and information structure \cite{knud1994, chafe1976, behaghel1932, gundel1988}. However, one of the most prominent factors, as prior research has suggested, is constituent weight and complexity \cite{quirketal, 1909bezie, Wasow_endweight}. 

As the following examples show, we tend to prefer sentences where longer and more complex constituents are moved to the end of the sentence: 

\ex. \label{ex:weight-ex}
\a. I met [the man]$_\text{NP}$ [at the park]$_\text{PP}$. \label{ex:5a}
\b. * I met [at the park]$_\text{PP}$ [the man]$_\text{NP}$. \label{ex:5b}
\c. I met [at the park]$_\text{PP}$ [the tall man selling water to marathon runners]$_\text{NP}$. \label{ex:5c}
\d. ? I met [the tall man selling water to marathon runners]$_\text{NP}$ [at the park]$_\text{PP}$. \label{ex:5d}

The typical constituent order shown in \ref{ex:5a}, for example, is not readily perturbed to the constituent order shown in \ref{ex:5b}. In sentences \ref{ex:5c} and \ref{ex:5d}, however, where the NP is considerably longer and more complex than the PP, the reverse is true.\footnote{Note that, following common linguistics notation conventions, `*' refers to a sentence judged to be highly unacceptable, while `?' refers to a sentence whose acceptability is questionable.} 

Linguistic theory thus suggests that phrases and constituents are specifically ordered to be presented in increasing complexity, or \textit{weight}; essentially, the larger the constituent, the further to the end of the sentence we expect it to appear \cite{quirketal, 1909bezie, Wasow_endweight, futrell2015}. Consider the example in Figure~\ref{ex:pinker}:

\begin{figure}[ht]
\small{\texttt{
    "In my laboratory we use it as an easily studied instance of mental grammar, allowing us to document {\color{red}[in great detail]} {\color{teal}[the psychology of linguistic rules]} {\color{orange}[from infancy to old age]} {\color{blue}[in both normal and neurologically impaired people]} {\color{purple}[in much the same way that biologists focus on the fruit fly Drosophila to study the machinery of the genes]}."
}}
\caption{Example from \cite{pinker2007language}.}
\label{ex:pinker}
\end{figure}

Other orderings of this sentence, if greatly violating this principle of weight, would likely be considered undesirable. This relationship between the complexity of post-verbal constituents and their ordering raises several questions: 
\begin{itemize}
    \item{\textit{What are the exact effects of weight on constituent ordering, in terms of gradient and ceiling effects, of increasing complexities on sentence acceptability?}}
    
    \item{\textit{Which measures of `weight/complexity' best explain the effects on constituent ordering?}}

    \item{\textit{How exactly do LLM preferences around constituent shifting align with human constituent shifting preferences?}}
\end{itemize}

Psycholinguistic research has provided some insight into these questions for human language processing, but the same cannot yet be said about increasingly powerful non-human language processors \cite{Medeiros_Mains_McGowan_2021, wasow2002}. Prior research, however, supports the abilities of modern language models in assigning relative linguistic plausibility scores aligning similar to human preferences \cite{linzen2016assessingabilitylstmslearn, marvin-linzen-2018-targeted}. Additionally, we hypothesize that the human-feedback mechanism incorporated in instruction-tuned models will present even more similar judgements.

In this work, we study the behavior of LLMs with constituent movement in English. We model Heavy NP Shift (HNPS), Dative Alternation (DA),  Particle Movement (PM), Multiple PP Shift (MPP)---see Figure \ref{fig:shiftexs}---as a function of its weight. Weight corresponds to a number of selected measures: word length, syllable weight, token length, and modifier weight. We analyze these measures to determine which best explains constituent ordering effects. Figure~\ref{fig:overallFig} outlines our contributions, which are as follows:

\begin{itemize}
    \item We evaluate the preferences of models in regards to constituent movement, using a novel constituent shift dataset containing both synthetic and naturally occurring data.
    \item We study the motivating factors for constituent ordering preferences in models, over a variety of candidates.
    \item We compare the behaviors of models with human judgements, analyzing correlative trends between the two.
\end{itemize}

\begin{table*}[ht]
\resizebox{\linewidth}{!}{\begin{tabular}{c|c|c}
     & \textbf{Unshifted Form} & \textbf{Shifted Form}  \\ \Xhline{4\arrayrulewidth}
\textbf{HNPS} & \begin{tabular}[c]{@{}c@{}}S + V + NP + PP\\ \textit{I met {[}the tall man selling water{]}$_\text{NP}$ {[}at the park{]}$_\text{PP}$.}\end{tabular} & \begin{tabular}[c]{@{}c@{}}S + V + PP + NP\\ \textit{I met {[}at the park{]}$_\text{PP}$ {[}the tall man selling water{]}$_\text{NP}$.}\end{tabular} \\ \hline
\textbf{PM}   & \begin{tabular}[c]{@{}c@{}}S + V + NP + PRT\\ \textit{She looked {[}up{]}$_\text{PRT}$ {[}her question{]}$_\text{NP}$ on her computer.}\end{tabular}                & \begin{tabular}[c]{@{}c@{}}S + V + PRT + NP\\ \textit{She looked {[}her question{]}$_\text{NP}$ {[}up{]}$_\text{PRT}$ on her computer.}\end{tabular}                \\ \hline
\textbf{DA}   & \begin{tabular}[c]{@{}c@{}}S + V + NP1 + NP2\\ \textit{He sent {[}her{]}$_\text{NP}$ {[}a gift{]}$_\text{NP}$ for her birthday.}\end{tabular}            & \begin{tabular}[c]{@{}c@{}}S + V + NP1 + PP\\ \textit{He sent {[}a gift{]}$_\text{NP}$ {[}to her{]}$_\text{PP}$ for her birthday.}\end{tabular}          \\ \hline
\textbf{MPP}  & \begin{tabular}[c]{@{}c@{}}S + V + PP1 + PP2\\ \textit{I went {[}to the mall{]}$_\text{PP}$ {[}with my sister{]}$_\text{PP}$ on Sunday.}\end{tabular}                    & \begin{tabular}[c]{@{}c@{}}S + V + PP2 + PP1\\ \textit{I went {[}with my sister{]}$_\text{PP}$ {[}to the mall{]}$_\text{PP}$ on Sunday.}\end{tabular}
\end{tabular}}
\caption{Unshifted versus shifted forms for each shift. Note that the unshifted and shifted form for MPP is ambiguous; the unshifted and shifted forms cannot be derived separately given an example.}

\label{table:formstable}
\end{table*}
\section{Background}
\label{sec:bg}
At a high level, English follows a subject-verb-object (SVO) ordering; beyond this basic structure, other objects, modifiers, constituents, and clauses can be added to form more complex sentences \cite{nonverbpred}. The organization and format of how and when each constituent in a sentence is delivered, or its ordering, can be highly flexible \cite{bakker1998flexibility, namboodiripad2019gradient, namboodiripad2017experimental}.

Constituent shifting is the process of reordering the constituents of a sentence, such that the original meaning of the sentence is maintained, and all semantic truth conditions are unchanged. This work focuses on four specific types of shift, with a prime commonality: each shift involves the movement of constituents from a post-verbal position (i.e. appearing after the verb of a sentence) to another post-verbal position \cite{Wasow_Arnold_2003, wasow2002}.\footnote{Topicalization, for instance, is not considered, as it can involve movement from a post-verbal position to a pre-verbal position, and vice-versa.} Table~\ref{table:formstable} demonstrates how we define shifted/unshifted sentences.

In our experiments, we consider the following measures of weight: \textbf{word length}, which corresponds to the number of words in a constituent; \textbf{syllable weight}, which corresponds to the number of syllables in a constituent; \textbf{token length}, which refers to the number of tokens in a constituent, determined by respective model tokenizers; \textbf{modifier weight}, which refers to the number of adjective phrase (AdjP) and prepositional phrase (PP) modifying the constituent itself (plus 1 to include the weight of the base constituent itself as normalization). 

Absolute weight itself, however, is not the most effective metric for observing motivations for constituent movement; instead, there is reason to believe that it is the \textit{relative} weight of constituents that determines their ordering \cite{Wasow_1997}. For instance, for phrases "with her grandmother" and "around the garden", the ratio of word lengths would be 1 (3:3); the ratio of syllables would be 1 (5:5), etc. For phrases "with her grandmother" and "around the decorated entryway garden with the large fountain", the ratio of word length would be $\frac{1}{3}$ (3:9); the ratio of syllables would be $\frac{5}{17}$ (5:17), etc. As the ratio of any metric increases beyond 1, we know that the weight of the first constituent is larger than the second, and thus, we predict that the motivation to shift will be greater.

\section{Related Work}
Constituent movement has been the subject of considerable linguistic literature; we categorize relevant contributions by our three research questions.

\subsection{What are the exact effects of weight on constituent ordering?}

Significant prior work has focused on investigating the effects of weight on constituent ordering \cite{arnold_et_al, Wasow_Arnold_2003, Wasow_endweight, ARNOLD200455, Hawkins_1995, 1909bezie, hawkins2004}; many contributions find gradient effects by which the shift becomes more frequent in examined language corpora as the relative weight of the relative constituent changes, suggesting that weight is the predominant factor in triggering the shift, even cross-linguistically \cite{Wasow_1997, Wasow_Arnold_2003, faghiri, WangLiu, Hawkins_1999, quirketal, Manetta2012ReconsideringRS}. Furthermore, studies with human participants find similar results, where weight presents a primary role in the shift \cite{Medeiros_Mains_McGowan_2021}. The study, however, also suggests that this movement is constrained by ceiling effects, by which the efficacy of additional weight and complexity plateaus. More closely relevant to this work, \citet{futrell2018rnnslearnhumanlikeabstract} conduct a similar analysis on post-verbal constituent movement using weight as a binary feature (i.e. `long' vs `short'), and find similar trends with LSTMs. 

\subsection{What measure of weight best explains effects of constituent ordering?}

Weight is a measure of a constituent's complexity or size, but how best to measure it is less straightforward \cite{logstruct, haegeman1991, Wasow_endweight}. Related research categorizes and analyzes the effects of three primary measures of weight on constituent movement, particularly with HNPS: the word length of the NP, the number of nodes in the NP's syntactic structure, and the number of modifiers applied to the NP \cite{Wasow_Arnold_2003, Medeiros_Mains_McGowan_2021, Wasow_1997}. The analyses found that, although the word length was statistically the strongest predictor for HNPS, ``no single factor can account for observed constituent order alternation'' \cite[~pg.6]{Medeiros_Mains_McGowan_2021}. Similarly, \citet{Wasow_Arnold_2003} find in a corpus study that for HNPS and DA, constituent movement was best accounted for when considering both word length and modifier weight together, as opposed to either on its own.

\subsection{How exactly do LLM preferences around constituent shifting align with human constituent shifting preferences?}

Research concerning the behaviors of computational models has also shown that models exhibit human-like preferences \cite{Fujihara2022TopicalizationIL, linzen2016assessingabilitylstmslearn, marvin-linzen-2018-targeted, 10.1162/kamath2024}.
Notably, prior work shows models learn syntactic alternations \cite{wilcox2019syntacticstructuresblockdependencies, lau2017}; more directly relevant to us, \citet{futrell2018rnnslearnhumanlikeabstract} finds that behaviors of LSTMs appear to correlate closely with observed judgements of humans on corresponding data, suggesting that constituent movement is motivated similarly in both humans and models. This work, however, predates preference-aligned models \cite{ouyang2022traininglanguagemodelsfollow}; we hypothesize that the behaviors of such models will align even more closely with human preferences around constituent ordering.

\begin{figure*}
    \centering
    \includegraphics[width=\textwidth]{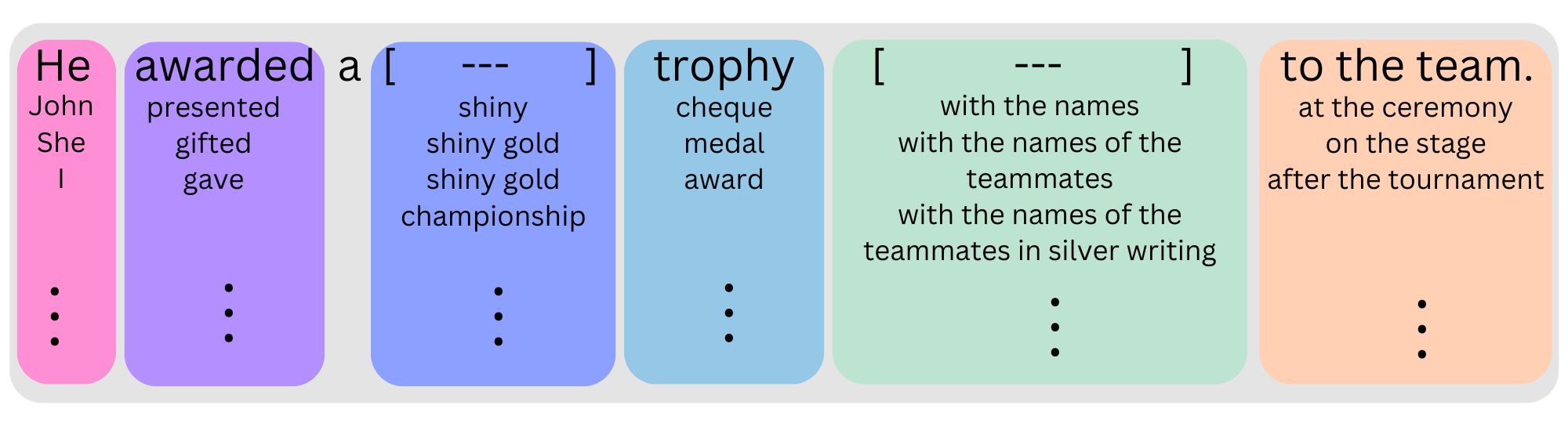}
    \caption{The outline for creating synthetic data, using varying modifier weights.}
    \label{fig:DFig}
\end{figure*}

\section{Models and Data}

\subsection{Data}
We both generate synthetic data using a template modified for the various shifts and structures and mine natural data from the Penn Treebank-2 corpus \cite{treebank_2}. Each shift we consider has a standardized form that we can utilize for both processes, noted in Table~\ref{table:formstable}. We also annotate for the aforementioned weight measures; syllable weight is computed using Syllapy;\footnote{https://github.com/mholtzscher/syllapy} token weight is retrieved simultaneously with the model scoring process; modifier weight is counted when constructing modifier chunks on constituents. 

\subsubsection{Synthetic Data}
We synthetically generate data using the process shown in Figure~\ref{fig:DFig} in order to accumulate large amounts of iteratively more complex data for model evaluations. Using an overall frame for the sentence, we alternate subjects, verbs, moving constituents, and their modifiers; we do this for all variables on a variety of constituents.

\begin{table}[t]
\begin{tabular}{lcc}
\hline
     & \textbf{Synthetic Data} & \textbf{Mined Data}\\ \hline
\textbf{HNPS}            & 3,888 & 314 \\ \hline
\textbf{PM}              & 4,136 & 131 \\ \hline
\textbf{DA}              & 210,304 & 123 \\ \hline
\textbf{MPP}             & 180,224 & 130 \\ \hline
\textbf{Total Size}     & \textbf{398,552} & \textbf{698} \\ \hline
\end{tabular}
\caption{Dataset sizes by sentence count. Synthetic datasets for DA and MPP are larger due to modification of both constituents, rather than one, leading to more constituent weight ratios.}
\vspace*{-2ex}
\end{table}

\subsubsection{Mined Data}
Our synthetic data, however, contains limited syntactic variation, and may not represent naturally occurring data; thus, we mine from existing data to ground our results. We use the Penn Treebank-2 to retrieve sentences following the structure of each shift \cite{treebank_2}. We collect over 1,000 examples for HNPS and DA, and approximately 400 for PM and MPP, selecting a random sample of 500 and 400 sentences, respectively. We manually inspect and exclude low-quality datapoints.\footnote{Low-quality data includes nonsensical or malformed sentences, unnecessary repetitions, or incorrect modifier structures.} As weight measures, we include constituent word length, syllable weight, and token length, but exclude modifier weight due to the complexity of accurately extracting this from such data.

\subsection{Models}
We select a range of open autoregressive models with diverse attributes: the entire GPT-2 model family \cite{radford2019language} was analyzed to study behaviors over scaled model sizes; Llama-3 8B, Llama-3 8B Instruct \cite{llama3modelcard}, Mistral v0.3 7B, Mistral v0.3 7B Instruct \cite{jiang2023mistral7b}, OLMo 7B, and OLMo 7B Instruct \cite{groeneveld2024olmoacceleratingsciencelanguage} were used to compare standard and instruction-tuned models; BabyFlamingo and BabyOPT \cite{warstadt-etal-2023-findings} were used to study LLMs trained on BabyLM data, a child-directed dataset to simulate language stimulus during the early human language acquisition period. We do not inspect closed-source models such as GPT-4 \cite{openai2024gpt4technicalreport}, due to inaccessibility of their underlying logits.

\section{Shifting Preference of Models}

\subsection{Approach}
\label{sec:appr}
To observe the effects of constituent weight on ordering preferences, we compute the difference in log probabilities assigned by models to shifted and unshifted sentences at a range of constituent weight ratios (see Section~\ref{sec:bg}). We begin by extracting log probabilities assigned to each sentence by the aforementioned models, using the \texttt{minicons} library \cite{misra2022minicons}. This score corresponds to the model's judgement of the sentence's linguistic plausibility, computed with the following equation:
\[
\resizebox{\columnwidth}{!}{$
M_{score}(\mathbf{w}) = \sum_{t=1}^{T} \log P_{\text{M}}(w_t | w_1, w_2, \ldots, w_{t-1}; \boldsymbol{\theta}).
$}
\]

M$_{score}$ is the log probability score of the sequence \( \mathbf{w} = [w_1, w_2, \ldots, w_t, \ldots, w_T]^T \), where \( w_t \) is the token at position \( t \). The term \( P_{\text{M}} \) is the conditional probability from the model $M$ of token \( w_t \) given the preceding tokens, while \( \boldsymbol{\theta} \) are the model parameters. The output of this formula is the sum of the probabilities of all tokens in the sequence, given previous tokens and model parameters, which equates to the overall sequence probability. The closer M$_{score}$ is to 0, the more strongly the model judges the sequence to likely occur in human language.

Upon computing this score for each minimal pair of shifted-unshifted sentences (denoted as $U$ and $S$), we calculate the difference of the M$_{score}$ for each sentence:
\[
\text{M}_{preference} = \text{M}_{score}(\textbf{$U$}) - \text{M}_{score}(\textbf{$S$}).
\]

This metric aligns closely with the surprisal-based measure used by \citet{futrell2018rnnslearnhumanlikeabstract}, and is in line with other surprisal-based metrics commonly used in work in psycholinguistics and computational linguistics \cite{linzen2016assessingabilitylstmslearn, futrell2018rnnspsycholinguisticsubjectssyntactic, wilcox-etal-2018-rnn, schuster2022sentence, baroni2022properrolelinguisticallyorienteddeep}.

Intuitively, the value of M$_{preference}$ captures the model's relative preference for the unshifted version of the sentence.
If this value is >0, the model has a stronger preference for the unshifted sentence, and if it is <0, the model has a stronger preference for the shifted version of the sentence; values approaching 0 suggest no clear preference between the two.

Though the usage of instruction-tuned models attracts the idea of directly prompting such LMs for their linguistic preferences, \citet{hu2023promptingsubstituteprobabilitymeasurements} and \citet{10.1162/kamath2024} show results that suggests meta-linguistic prompting underestimates linguistic capacities; thus, we only consider raw probability scores in our experimentation.

\begin{figure}[t]
    \centering
    \includegraphics[width=\columnwidth]{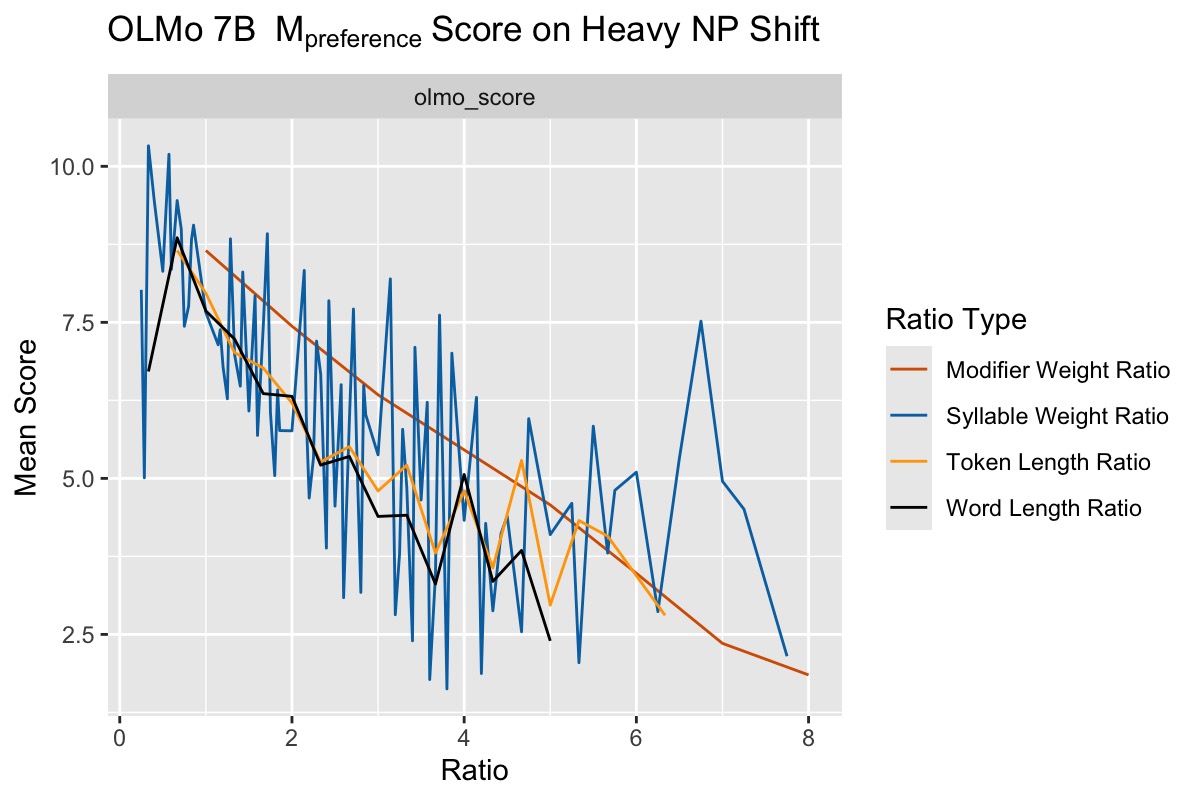}
    \caption{OLMo 7B M$_{preference}$ scores with respect to different measures of weight.}
    \label{fig:olmo-hnps}
\end{figure}

\subsection{Results: Are models motivated by weight to shift?}
\label{sec:res1}
We analyze model M$_{preference}$ scores by weight ratios (see Section~\ref{sec:bg}) considering various metric types in the analysis. Figure \ref{fig:olmo-hnps} shows OLMo 7B performance on HNPS; see Section~\ref{sec:sdplots} for remaining models and variables. We observe similar trends overall across models. 

For HNPS, we find that model M$_{preference}$ scores initially start positive, i.e., models prefer unshifted sentence, but converge above 0 as constituent weight ratios increase, indicate some continued preference for unshifted constituent orderings beyond a given point.
In the case of DA and MPP, we see a similar effect of weight M$_{preference}$ scores, but with a plateau \textit{below} zero, indicating that models eventually converge upon a relative preference for the shifted version of a sentence.
In the case of PM, however, we see scores initially drop sharply from around 0, but later somewhat rise.

\section{Motivating Factors of Model Preference}
Literature regarding human language finds varying levels of importance correlated with different measures of constituent weight, particularly in corpus studies and human judgement tasks \cite{Wasow_endweight, Wasow_Arnold_2003, Medeiros_Mains_McGowan_2021}. Whether or not this same trend is seen with language models is unknown. We conduct a regression analysis on the data collected for the first experiment, using a generalized additive mixed model \cite{wood2017generalized, sóskuthy2017generalisedadditivemixedmodels}, with the goal of measuring how significantly each weight measure both impacts the models' judgements and serves to fit regression lines on the data.

\subsection{Approach}
To compare how well different measures of weight explain shifting preferences, we fit Generalized Additive Mixed Models, or GAMMs \cite{wood2017generalized, sóskuthy2017generalisedadditivemixedmodels}, on our M$_{preference}$ scores (see Section~\ref{sec:appr}) as a function of various weight measures: word length, token length, syllable weight, and modifier weight. GAMMs allow for the fitting of highly non-linear relationships as the sum of multiple predictor-wise smooth functions: basis functions that allow for an arbitrary degree of smoothness. Crucially, aside from providing interpretable measures of goodness of fit, GAMMs also allow for grouping structures in the data to be captured as random effects \cite[~ch.6]{wood2017generalized}.

Bearing in mind that multiple measures of weight may jointly determine the accessibility of a shift \cite{Wasow_1997, Wasow_Arnold_2003}, we analyze the relative importance of each weight measure in the following manner. For each model, first, we fit a GAMM on the model’s M$_{preference}$ scores as a function of all weight predictors, with verb-wise random intercepts and slopes. We then iteratively ablate each weight predictor while retaining all others, and compare the quality of fit yielded by the full model with that of the ablated model. Intuitively, this provides an indication of how important the dropped predictor is for the LLM: it captures how much less of the LLM’s behaviour is explained when information about that given measure of weight is ignored.

\subsection{Results: Which measure of weight best explains shifting?}
Table~\ref{table:semireg} presents a subset of the results of our regression analysis (full results in Appendix Table~\ref{table:reg}). 
Crucially, we find that \textbf{syllable weight} is often the most important predictor of LLM behavior around constituent ordering preference, since the drop in R-squared scores is highest when syllable weight is not used.
For DA, word length seems to be the best predictor.

\begin{table}[t]
\centering\resizebox{\columnwidth}{!}{\begin{tabular}{ccccccc}
\Xhline{4\arrayrulewidth}
\textbf{Model} & \textbf{Var} & \textbf{Full} & \textbf{Token$_{ab}$} & \textbf{Word$_{ab}$} & \textbf{Syll$_{ab}$} & \textbf{Mods$_{ab}$} \\ \Xhline{4\arrayrulewidth}
GPT-2 Medium & & \textbf{.654} & .629 & .616 & .540 & .635 \\ 
Llama-3 &  HNPS & .542 & .524 & .519 & .485 & .538 \\ 
Llama-3 Instruct & & .452 & .438 & .408 & \underline{.359} & .451 \\ 
BabyLlama & & .527 & .514 & .466 & .415 & .509 \\ \Xhline{4\arrayrulewidth}
GPT-2 Medium & & .605 & .602 & .581 & \underline{.534} & .580 \\ 
Llama-3 &  PM & .608 & .603 & .586 & .564 & .590 \\ 
Llama-3 Instruct &  & .627 & .619 & .599 & .555 & .602 \\ 
BabyLlama & & \textbf{.719} & .702 & .663 & .651 & .701 \\ \Xhline{4\arrayrulewidth}
GPT-2 Medium &  & .571 & .561 & .552 & .568 & .568 \\ 
Llama-3 & DA & .554 & .544 & .532 & .538 & .549 \\ 
Llama-3 Instruct & & .503 & .493 & .490 & \underline{.486} & .496 \\ 
BabyLlama & & \textbf{.630} & .616 & .603 & .622 & .623 \\ \Xhline{4\arrayrulewidth}
GPT-2 Medium & & \textbf{.368} & .309 & .302 & .300 & .356 \\ 
Llama-3 &  MPP & .358 & .316 & .321 & .271 & .351 \\ 
Llama-3 Instruct & & .320 & .284 & .291 & \underline{.208} & .313 \\ 
BabyLlama &  & .310 & .297 & .298 & .281 & .306 \\ \Xhline{4\arrayrulewidth}
\end{tabular}}
\caption{R-squared scores for a subset of models; see full table in Appendix~\ref{sec:reg}. $Full$ denotes R-squared values from GAMMs with all weight measure predictors included, while $[Metric]_{ab}$ denotes the R-squared score when excluding the metric; a larger difference compared to the original R-squared score denotes higher significance in contributing to the overall R-squared. Numbers bolded denote high R-sq.; numbers underlined denote low R-sq.}
\label{table:semireg}
\end{table}

We find that GPT-2 Medium achieves the highest R-squared overall on both HNPS and MPP. Further, contrary to our hypotheses, across almost all shift types, instruction-tuned models consistently achieve lower R-squared scores than their base model counterparts. Table~\ref{table:semireg}, for example, shows these results for Llama-3 and Llama-3 Instruct; see Appendix~\ref{sec:sdplots} for remaining plots on synthetic data. The BabyLM models also present high R-squared scores on both PM and DA, with BabyLlama yielding the highest R-squared values (see Table~\ref{table:semireg} for BabyLlama results and Appendix~\ref{sec:sdplots} for BabyOPT results). Finally, despite the high performance of GPT-2 Medium, we do not observe consistent improvements in R-squared values as model sizes scale.

\section{Human-Model Preference Correlation}

\subsection{Approach}
To adequately compare the behaviors of LLMs with those of humans, a direct study of preferences on identical data points is necessary. We collect human judgements on a subset of data presented to models; though human judgements and model scores are not identical metrics, they can act as proxies when comparing relative trends. 

We conduct a crowdsourced study through Prolific, collecting judgments from 126 native English speakers residing in Anglophone countries, on 500 sentence pairs. Each participant is presented 25 sentence pairs and asked to judge how natural they sound in relation to each other, assigning a score between 1 and 7; 1 corresponds to the first sentence presented appearing far more natural than the second, and 7 the reverse. We exclude datapoints with low inter-annotator agreement to minimize noise.\footnote{Datapoints with standard deviation of participation responses >1.5, as well as those from participants who consistently failed attention checks, were excluded.}

\subsection{Results: Do LLM and human preferences correlate?}
\begin{figure}[t]
    \centering
    \includegraphics[width=\columnwidth]{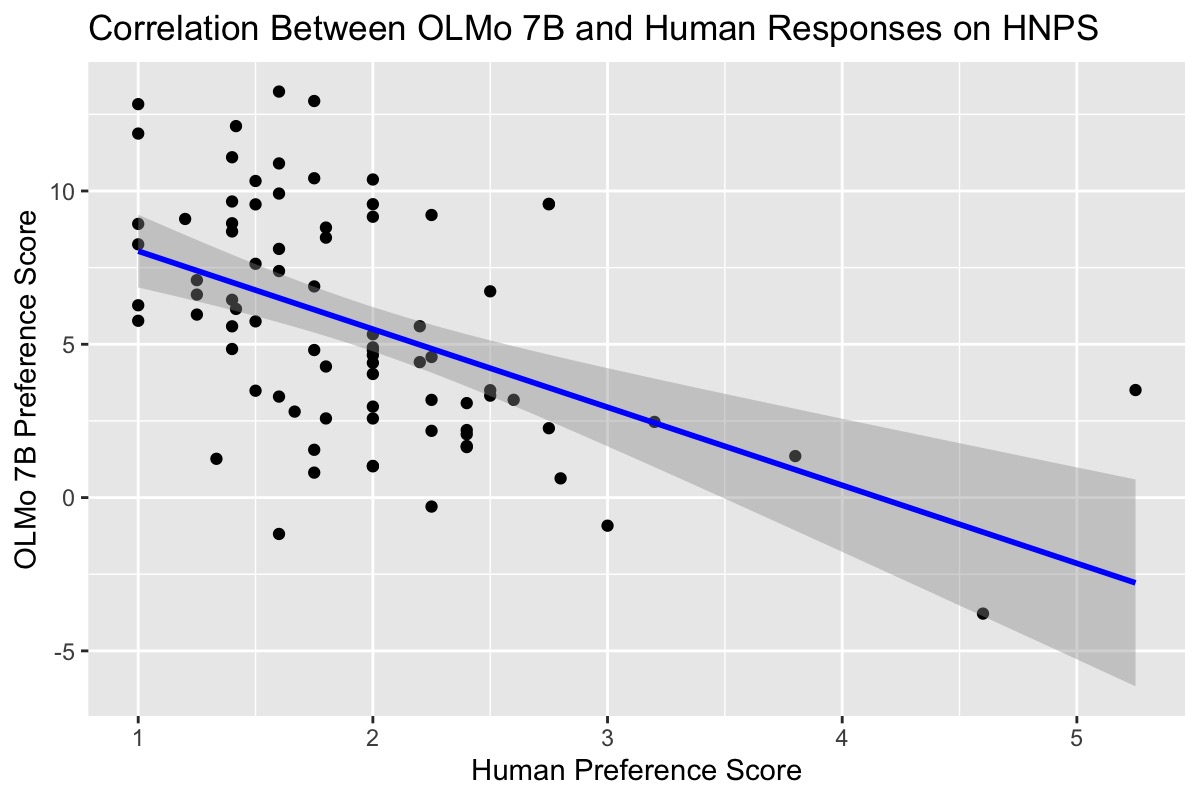}
    \caption{Scatterplot of human judgment scores and OLMo 7B M$_{preference}$ scores on Heavy NP Shift data.}
    \label{fig:hnps-olmo-human}
\end{figure}

To analyze how this human judgment data compares with model results, we compute the Spearman correlation between the average human score for each data point and the model's M$_{preference}$ score. We present these scores in Table~\ref{table:hmcorr}. We also plot model scores against human judgment data to observe the correlation visually; these plots are presented in Figure~\ref{fig:hnps-olmo-human} and Section~\ref{sec:hmagree}.

\begin{table*}[ht]
\centering\resizebox{\textwidth}{!}{\begin{tabular}{ccccccccccccc}
\toprule
& \textbf{GPT-2} & \textbf{GPT-2 Med} & \textbf{GPT-2 Large} & \textbf{GPT-2 XL} & \textbf{BabyOPT} & \textbf{BabyLlama} & \textbf{Llama-3} & \textbf{Llama-3 I} & \textbf{Mistral v0.3} & \textbf{Mistral v0.3 I} & \textbf{OLMo} & \textbf{OLMo I} \\ \midrule
\textbf{\begin{tabular}[c]{@{}c@{}}HNPS\end{tabular}} & \large{.410}     & \large{.440}            & \large{.342}           & \large{.390}        & \large{.261}       & \large{.420}         & \large{.428}       & \large{.386}           & \large{.414}             & \large{.344} & \textbf{\large{.509}}    & \large{.431}             \\ \midrule
\textbf{\begin{tabular}[c]{@{}c@{}}PM\end{tabular}}   & \large{.212}     & \large{.125}            & \large{.213}           & \large{.295}        & \large{.196}        & \large{.261}         & \large{.305}       & \large{.293}            & \large{.315}             & \large{.347} & \large{.430}    & \textbf{\large{.431}}             \\ \midrule
\textbf{\begin{tabular}[c]{@{}c@{}}DA\end{tabular}}   & \textbf{\large{.651}}     & \large{.565}            & \large{.505}           & \large{.541}        & \large{.256}       & \large{.391}         & \large{.600}       & \large{.511}            & \large{.524}            & \large{.449} & \large{.494}    & \large{.478}             \\ \hline
\textbf{\begin{tabular}[c]{@{}c@{}}MPP\end{tabular}} & \large{.371}     & \large{.395}            & \large{.361}           & \textbf{\large{.579}}        & \large{.233}       & \large{.357}         & \large{.402}       & \large{.222}            & \large{.487}            & \large{.412} & \large{.513}    & \large{.263}             \\ \bottomrule
\end{tabular}}
\caption{Spearman correlation values (absolute) between LLMs and humans on sentence judgement tasks.
}
\vspace*{-0.2cm}
\label{table:hmcorr}
\end{table*}

Comparing preferences of humans and models on a statistical level introduces interesting findings. 
Notably, the GPT-2 and OLMo 7B classes of models appear to align most closely with human behaviors, achieving the highest correlation scores across all shift types.

Remarkably, most of the more highly correlated models are the base versions of their respective model category---the instruction-tuned models quite often performed in manners less correlated with human preferences than their base-model counterparts. This contrast is most stark in the case of MPP. Interestingly, no clear trend in alignment can be seen in the GPT-2 family as model sizes scale on any of the judgement tasks. We present remaining correlation figures in Section~\ref{sec:hmagree}.
Crucially, correlations between model scores and human judgments are particularly low on the PM data.

\section{Discussion}

\subsection{What are the exact effects of weight on constituent ordering?}
In Figure~\ref{fig:olmo-hnps} and Appendix~\ref{sec:sdplots}, we observe converging effects as weight increases, suggesting that prior theories defining weight as a prime factor in motivating the shift hold with computational models as well. Further, we observe, specifically with PM, that weight, beyond a certain threshold, begins to detriment motivation for shifting.

We observe similar effects on the scraped corpus data, in Appendix~\ref{sec:mdplots}, though with more noise. Given the specificity of the data itself, being rooted in financial reports and news articles, some noise and outliers were expected, and in some cases, observable trends remain. Similar to the synthetic data plots, we see convergence on HNPS, and some on MPP, as well as an initial drop followed by a rise in M$_{preference}$ scores for PM.

\subsection{What measure of weight best explains effects of constituent ordering?}
The syllable weight was the best measure of weight for explaining motivations to shift in LLMs. This raises an obvious question--why is syllable weight a more effective predictor of model behavior than token weight, which would intuitively be most aligned with a model's processing of weight and complexity? This finding acts as initial evidence that models may induce linguistic information not just at the token level, but also implicitly at the level of syllables.\footnote{Admittedly, however, this evidence is not conclusive proof of such `knowledge'; it is also possible that syllabic information fits the data best simply because of how much it varies--syllabic information is more likely to be a signal for specific words and phrases, due to the fact that while any two tokens could have the same ratio of tokens between them, they may vary in their syllable weights.}

\subsection{How exactly do LLM preferences around constituent shifting align with human constituent shifting preferences?}
Broadly speaking, a clear trend is maintained across humans and models, following what was presented by \citet{futrell2018rnnslearnhumanlikeabstract} in their analysis. Where human language sees motivation for movement with increasing weight, model behavior follows closely. Our experiment, which includes graded data beyond binary weight categorizations, and a wider range of models, yields relatively high correlation effects between preferences of models and humans, as presented in Table~\ref{table:hmcorr}---suggesting noticeably similar behaviors between the two on particular linguistic tendencies, with the notable exception of particle movement.

Interestingly, we observe an unexpected trend where instruction-tuned models, which consistently correlate less with human data than their corresponding base model, as well as, quite often, yield lower R-squared scores.
This runs against our initial hypothesis around instruction-tuned models, and suggests inadequacy in providing consistent and explainable trends compared to base models.

\subsection{Future Work}
Our findings suggest that even though newer models are equipped with more parameters, training data, and the human-feedback mechanism, they fail to align better with human linguistic preferences than their earlier counterparts, raising questions for future study. Equally, it invites further research into how models generate such sentences in standard conversational usage. We also wish to study further the motivations of LLMs in constituent movement, primarily regarding analysis of theories suggesting disparities between ordering preference motivations of listeners and speakers \cite{wasow2002}.

\section{Conclusions}
In this work, we present a thorough analysis of the behaviors of LLMs in response to constituent movement, using both a novel set of nearly 400K minimal pairs of variably ordered sentences, as well as naturally occurring data. We collect human judgements and model preference scores and observe comparable trends between the behaviors of humans and models. Such comparisons indicate that humans and LLMs largely hold similar linguistic preferences around constituent ordering, with the exception of particle movement. Our findings---and in particular, the surprising gap we find between instruction-tuned models and their vanilla counterparts---invite further research into when and how linguistic preferences of models and humans align.

\section{Limitations}
We only focus on constituent movement in English, even though this phenomena is known to manifest cross-linguistically \cite{faghiri, WangLiu, Hawkins_1999, quirketal, Manetta2012ReconsideringRS, Fujihara2022TopicalizationIL}.

\section{Ethics Statement}
Our experimentation poses no risks or harms for any participants involved. Human participants from our data study through Prolific were compensated on average US\$12 per hour. Anonymous participants were informed of the purpose of the study and how their responses would be used, as well as their rights regarding submitted data.

\section{Acknowledgements}
This work was partly funded by IVADO R3AI NLP Regroupement and a Doctoral Training Award from the \textit{Fonds de Recherche du Québec—Société et Culture}. We also thank Morgan Sonderegger for his invaluable feedback and guidance.

\bibliography{custom}
\onecolumn

\appendix

\section{Appendix}
\label{sec:appendix}

\subsection{Regression Table}

\label{sec:reg}
\begin{table}[ht]
\centering\resizebox{0.54\columnwidth}{!}{\begin{tabular}{ccccccc}
\Xhline{4\arrayrulewidth}
\textbf{Model} & \textbf{Var} & \textbf{Full} & \textbf{Token$_{ab}$} & \textbf{Word$_{ab}$} & \textbf{Syll$_{ab}$} & \textbf{Mods$_{ab}$} \\ \Xhline{4\arrayrulewidth}
GPT-2 & PM & 0.543 & 0.533 & 0.515 & 0.457 & 0.517 \\ \hline
GPT-2 Med & PM & 0.605 & 0.602 & 0.581 & 0.534 & 0.580 \\ \hline
GPT-2 Large & PM & 0.611 & 0.605 & 0.579 & 0.556 & 0.585 \\ \hline
GPT-2 XL & PM & 0.613 & 0.610 & 0.583 & 0.561 & 0.590 \\ \hline
Llama-3 & PM & 0.608 & 0.603 & 0.586 & 0.564 & 0.590 \\ \hline
Llama-3 I & PM & 0.627 & 0.619 & 0.599 & 0.555 & 0.602 \\ \hline
BabyOPT & PM & 0.562 & 0.550 & 0.534 & 0.507 & 0.542 \\ \hline
BabyLlama & PM & \textbf{0.719} & 0.702 & 0.663 & 0.651 & 0.701 \\ \hline
Mistral v0.3 & PM & 0.606 & 0.588 & 0.563 & 0.565 & 0.588 \\ \hline
Mistral v0.3 I & PM & 0.630 & 0.618 & 0.585 & 0.601 & 0.618 \\ \hline
OLMo & PM & 0.655 & 0.639 & 0.623 & 0.612 & 0.637 \\ \hline
OLMo I & PM & 0.511 & 0.505 & 0.494 & 0.469 & 0.481 \\ \Xhline{4\arrayrulewidth}
GPT-2 & MPP & 0.379 & 0.325 & 0.337 & 0.264 & 0.370 \\ \hline
GPT-2 Med & MPP & 0.368 & 0.309 & 0.302 & 0.300 & 0.356 \\ \hline
GPT-2 Large & MPP & 0.311 & 0.265 & 0.273 & 0.213 & 0.302 \\ \hline
GPT-2 XL & MPP & 0.323 & 0.249 & 0.260 & 0.244 & 0.310 \\ \hline
Llama-3 & MPP & 0.358 & 0.316 & 0.321 & 0.271 & 0.351 \\ \hline
Llama-3 I & MPP & 0.320 & 0.284 & 0.291 & 0.208 & 0.313 \\ \hline
BabyOPT & MPP & \textbf{0.441} & 0.433 & 0.435 & 0.368 & 0.432 \\ \hline
BabyLlama & MPP & 0.310 & 0.297 & 0.298 & 0.281 & 0.306 \\ \hline
Mistral v0.3 & MPP & 0.325 & 0.302 & 0.315 & 0.239 & 0.314 \\ \hline
Mistral v0.3 I & MPP & 0.269 & 0.238 & 0.258 & 0.186 & 0.256 \\ \hline
OLMo & MPP & 0.306 & 0.303 & 0.296 & 0.222 & 0.294 \\ \hline
OLMo I & MPP & 0.263 & 0.257 & 0.253 & 0.237 & 0.250 \\ \Xhline{4\arrayrulewidth}
GPT-2  & HNPS & 0.356 & 0.343 & 0.332 & 0.278 & 0.349 \\ \hline
GPT-2 Med & HNPS & \textbf{0.654} & 0.629 & 0.616 & 0.540 & 0.635 \\ \hline
GPT-2 Large & HNPS & 0.589 & 0.570 & 0.554 & 0.430 & 0.585 \\ \hline
GPT-2 XL & HNPS & 0.592 & 0.587 & 0.557 & 0.527 & 0.589  \\ \hline
Llama-3 & HNPS & 0.542 & 0.524 & 0.519 & 0.485 & 0.538 \\ \hline
Llama-3 I & HNPS & 0.452 & 0.438 & 0.408 & 0.359 & 0.451 \\ \hline
BabyOPT  & HNPS & 0.491 & 0.457 & 0.472 & 0.421 & 0.450 \\ \hline
BabyLlama & HNPS & 0.527 & 0.514 & 0.466 & 0.415 & 0.509 \\ \hline
Mistral v0.3   & HNPS & 0.488 & 0.476 & 0.461 & 0.451 & 0.458 \\ \hline
Mistral v0.3 I & HNPS & 0.467 & 0.441 & 0.436 & 0.438 & 0.434 \\ \hline
OLMo & HNPS & 0.518 & 0.500 & 0.468 & 0.443 & 0.513 \\ \hline
OLMo I & HNPS & 0.411 & 0.391 & 0.377 & 0.290 & 0.395 \\ \Xhline{4\arrayrulewidth}
GPT-2 & DA & 0.562 & 0.554 & 0.542 & 0.545 & 0.549  \\ \hline
GPT-2 Med & DA & 0.571 & 0.561 & 0.552 & 0.568 & 0.568 \\ \hline
GPT-2 Large & DA & 0.541 & 0.528 & 0.506 & 0.533 & 0.533 \\ \hline
GPT-2 XL & DA & 0.561 & 0.553 & 0.540 & 0.552 & 0.555 \\ \hline
Llama-3 & DA & 0.554 & 0.544 & 0.532 & 0.538 & 0.549 \\ \hline
Llama-3 I & DA & 0.503 & 0.493 & 0.490 & 0.486 & 0.496  \\ \hline
BabyOPT & DA & 0.252 & 0.242 & 0.238 & 0.241 & 0.237 \\ \hline
BabyLlama & DA & \textbf{0.630} & 0.616 & 0.603 & 0.622 & 0.623 \\ \hline
Mistral v0.3 & DA & 0.540 & 0.529 & 0.526 & 0.528 & 0.534 \\ \hline
Mistral v0.3 I & DA & 0.478 & 0.469 & 0.461 & 0.467 & 0.467 \\ \hline
OLMo  & DA & 0.467 & 0.460 & 0.458 & 0.449 & 0.459 \\ \hline
OLMo I & DA & 0.392 & 0.384 & 0.388 & 0.374 & 0.385  \\ \Xhline{4\arrayrulewidth}
\end{tabular}}
\caption{R-sq. scores for each model on each variable. $[Metric]_{ab}$ denotes the R-sq. score when excluding the metric; a larger difference from the original score denotes higher significance in contributing to the overall score.}
\label{table:reg}
\end{table}

\newpage

\subsection{Model Behavior Plots}

\subsubsection{Synthetic Data}
\label{sec:sdplots}
\begin{figure}[ht]
    \centering
    \includegraphics[width=\textwidth]{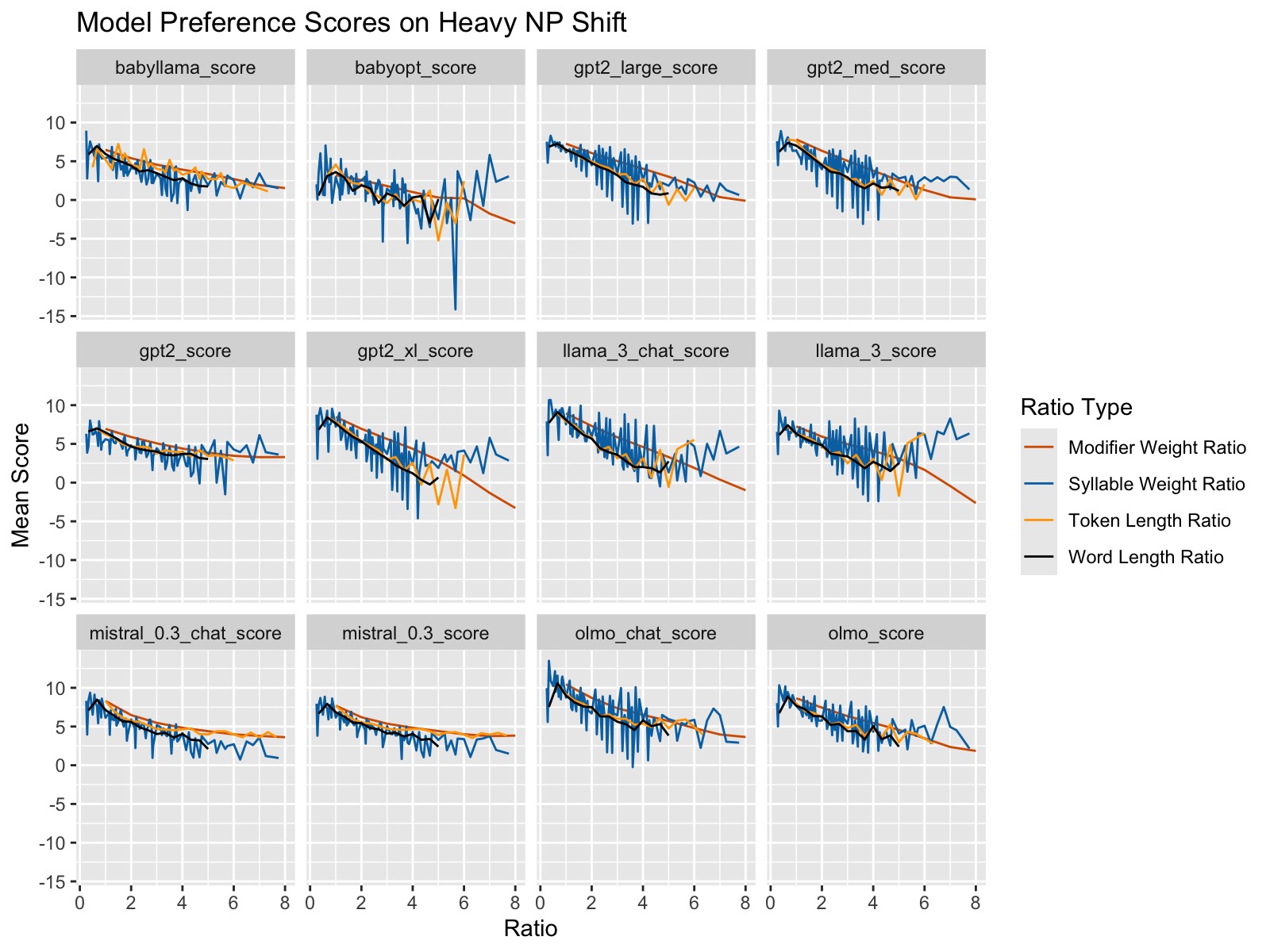}
\end{figure}
\begin{figure}[ht]
    \centering
    \includegraphics[width=\textwidth]{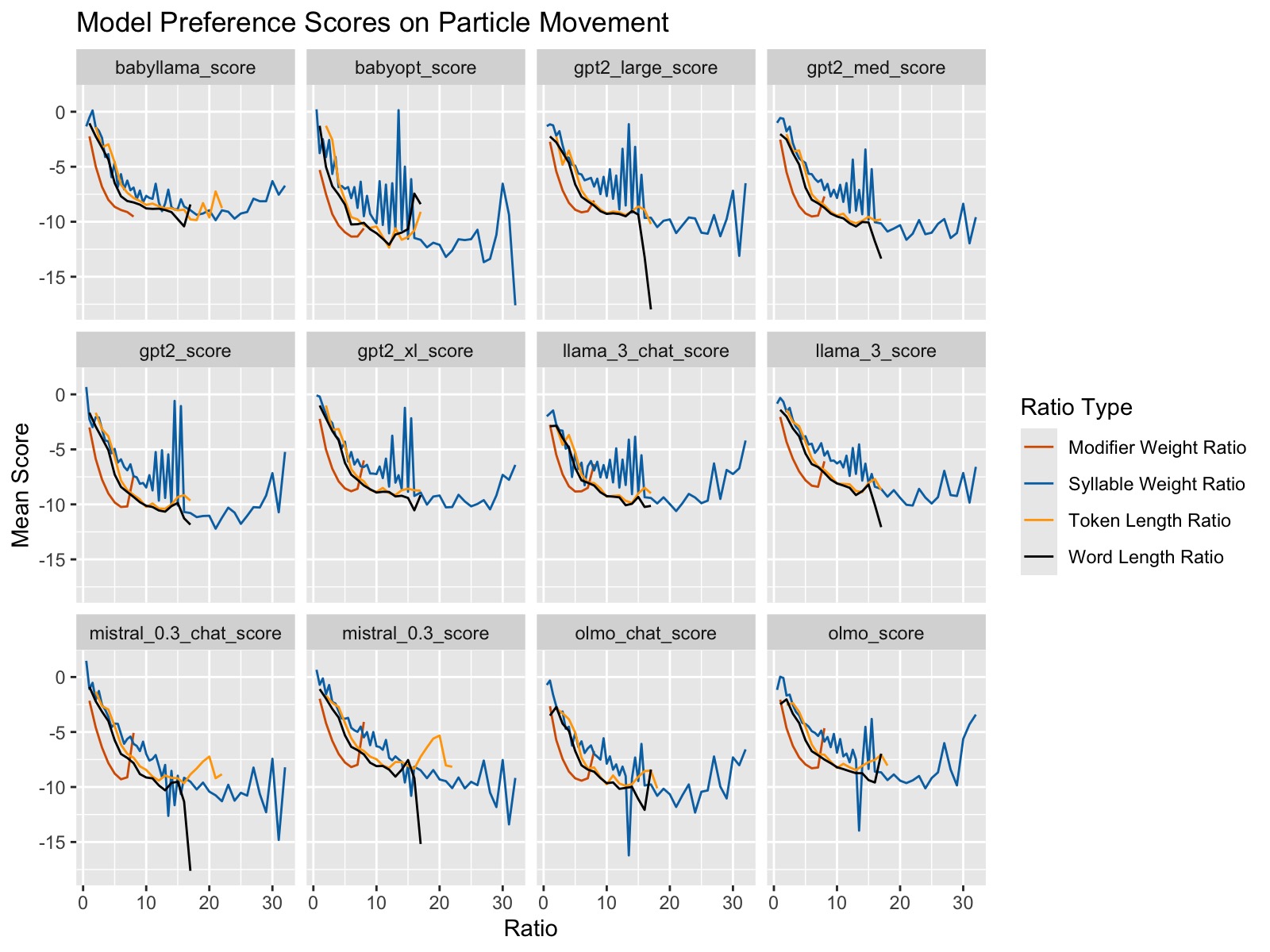}
\end{figure}
\begin{figure}[ht]
    \centering
    \includegraphics[width=\textwidth]{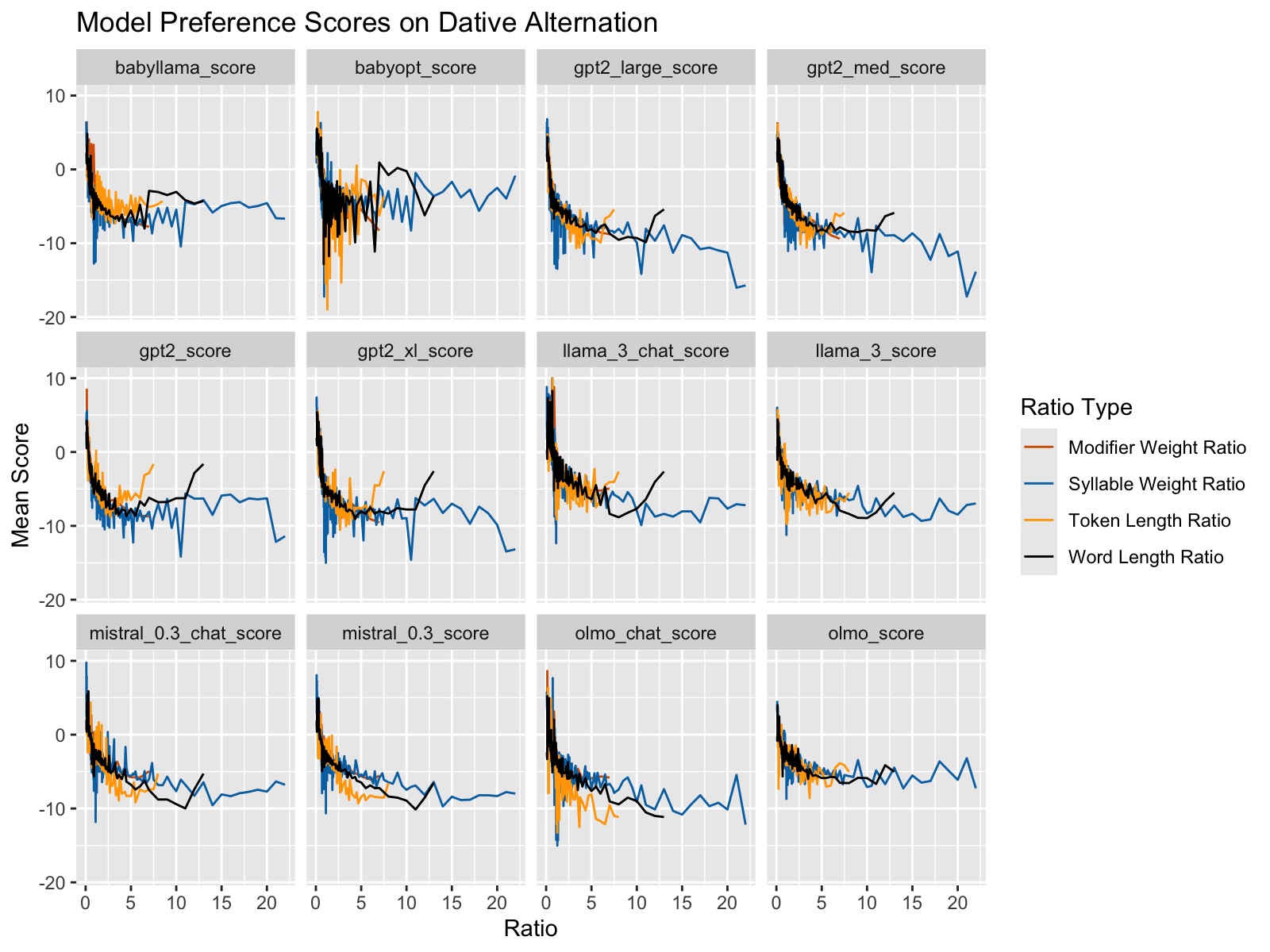}
\end{figure}
\begin{figure}[ht]
    \centering
    \includegraphics[width=\textwidth]{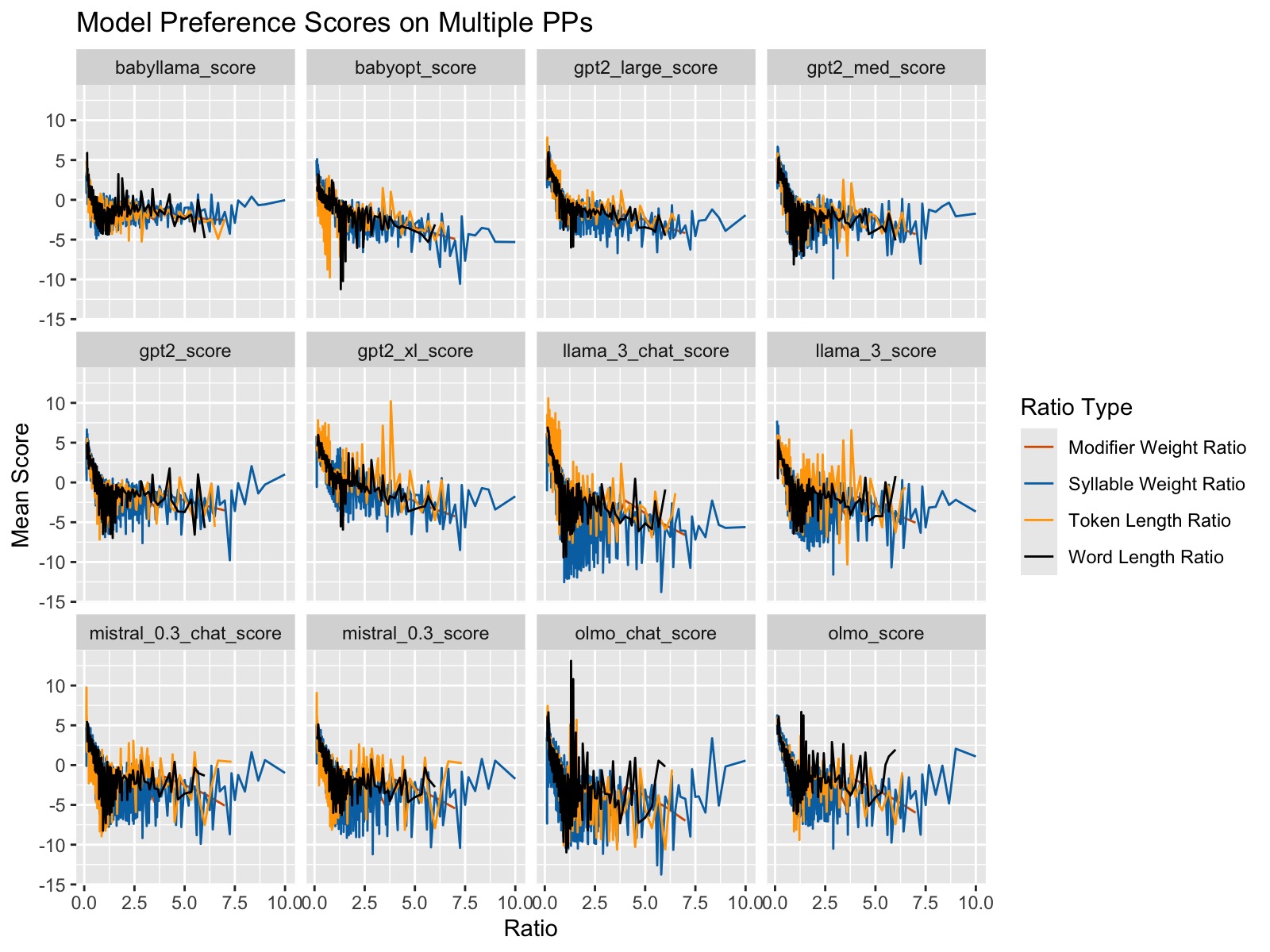}
\end{figure}

\newpage\newpage

\subsubsection{Mined Data}
\label{sec:mdplots}
\begin{figure}[ht]
    \centering
    \includegraphics[width=\textwidth]{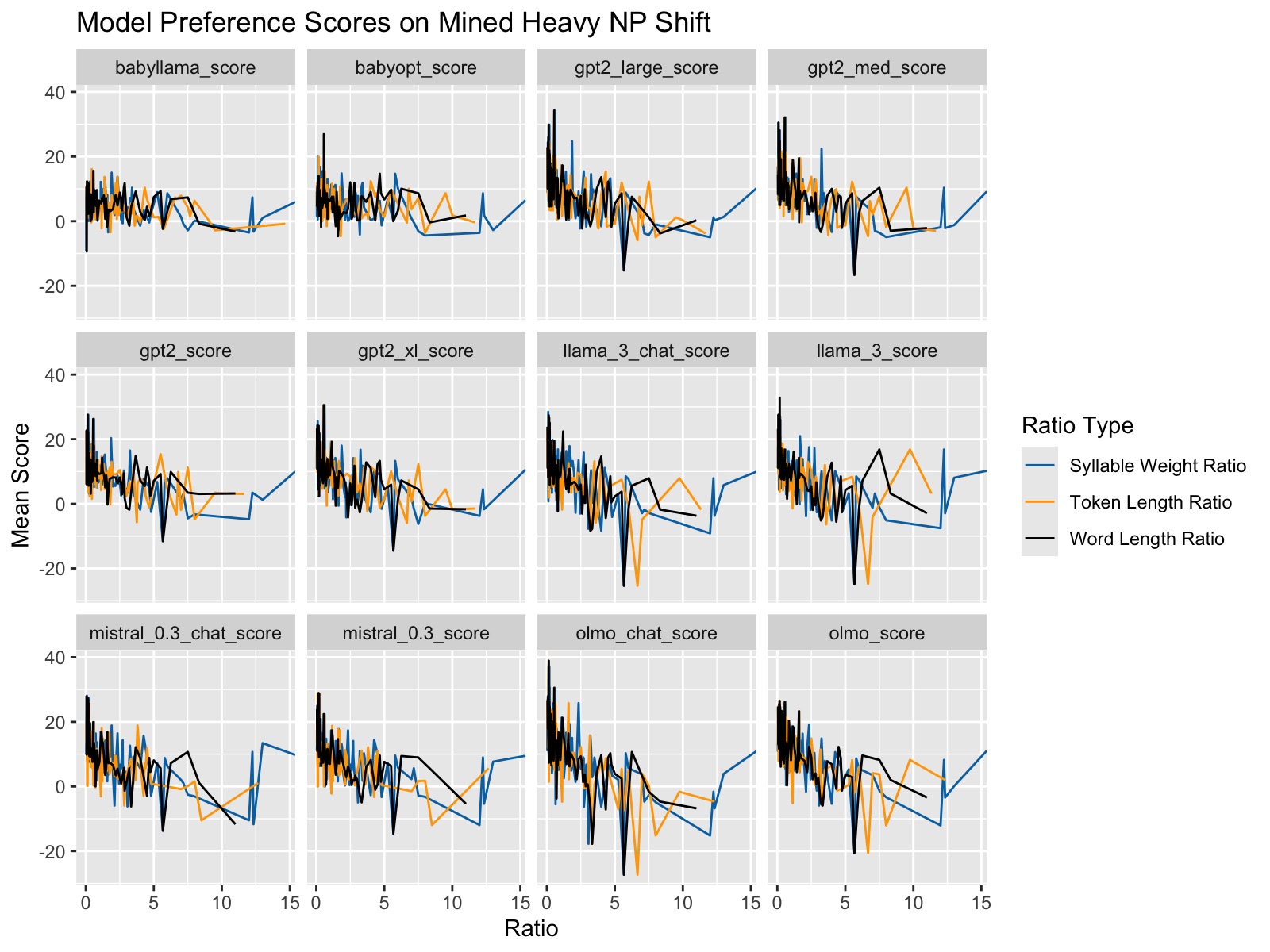}
\end{figure}
\begin{figure}[ht]
    \centering
    \includegraphics[width=\textwidth]{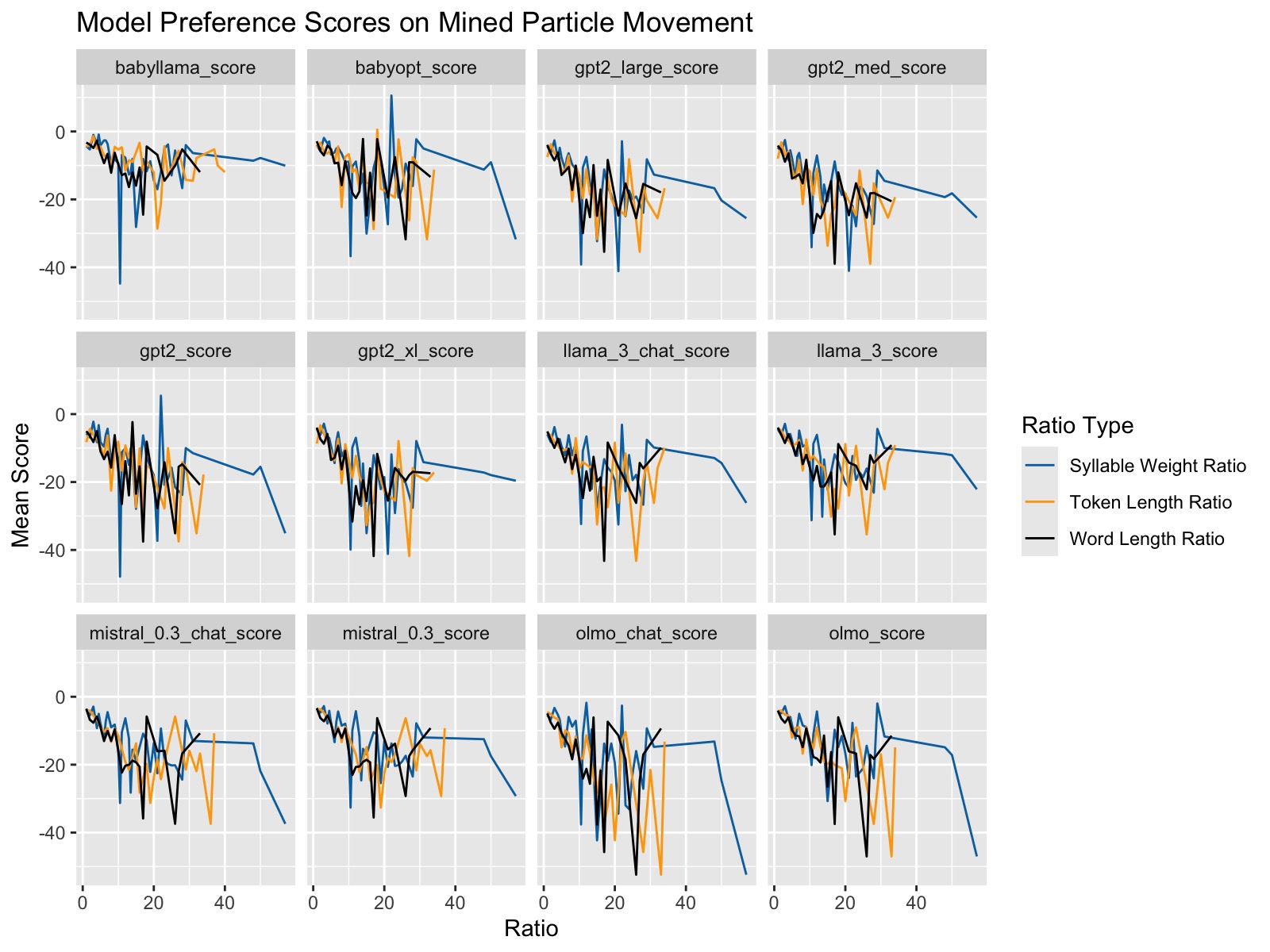}
\end{figure}
\begin{figure}[ht]
    \centering
    \includegraphics[width=\textwidth]{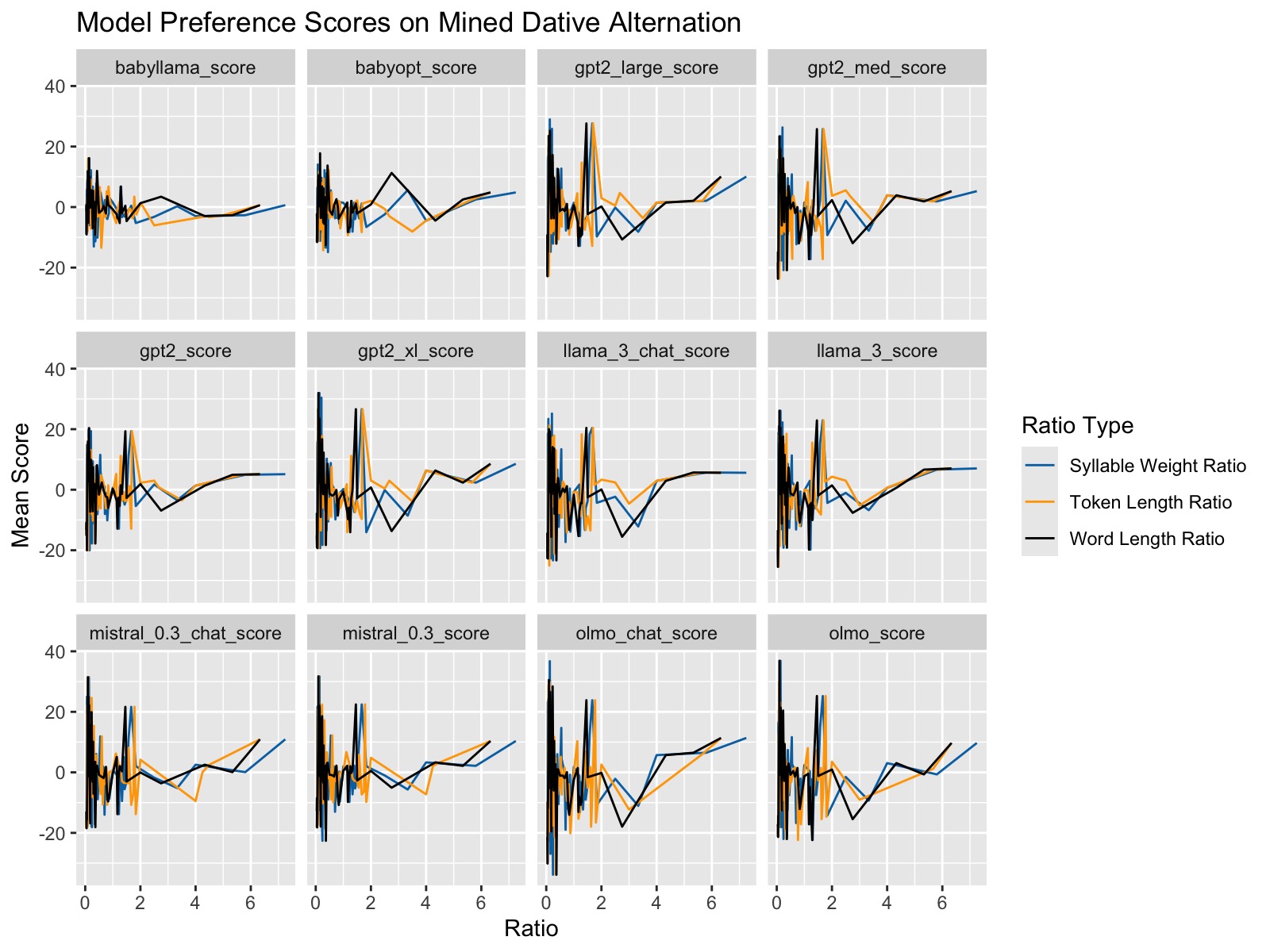}
\end{figure}
\begin{figure}[ht]
    \centering
    \includegraphics[width=\textwidth]{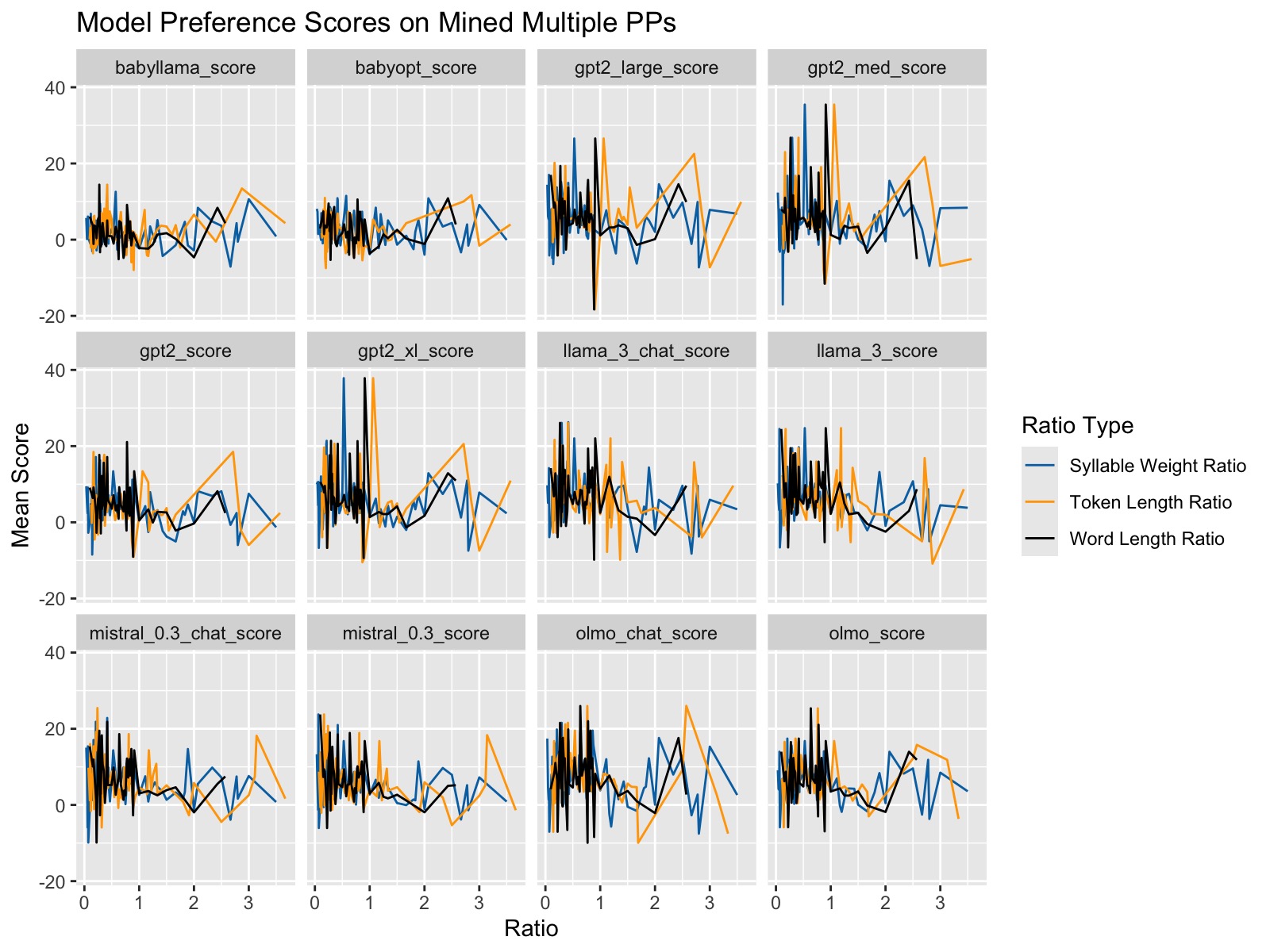}
\end{figure}

\newpage

\twocolumn
\subsection{Human-Model Agreement Plots}
\label{sec:hmagree}
\begin{figure}[ht]
    \centering
    \includegraphics[width=\columnwidth]{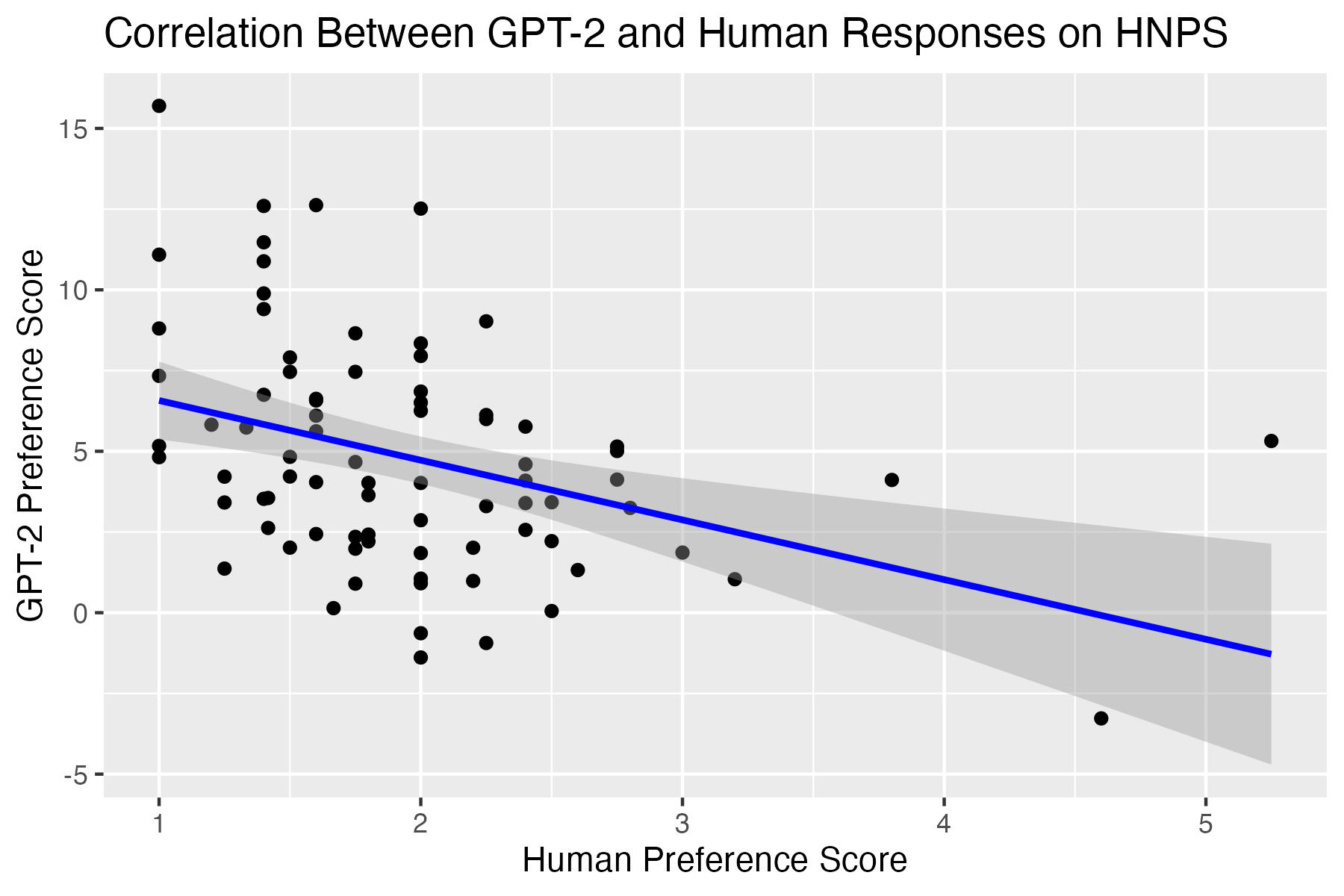}
\end{figure}
\begin{figure}[ht]
    \centering
    \includegraphics[width=\columnwidth]{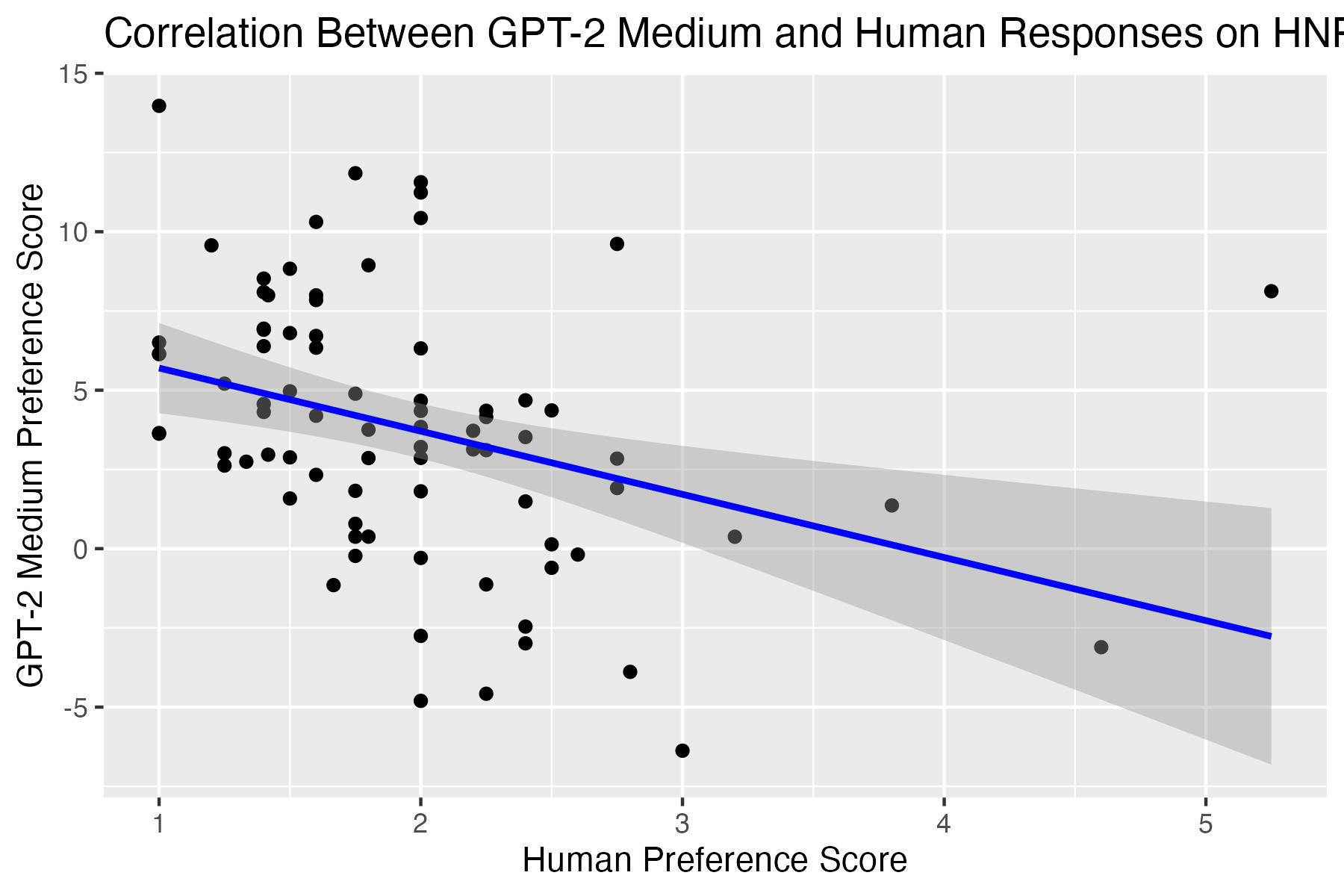}
\end{figure}
\begin{figure}[ht]
    \centering
    \includegraphics[width=\columnwidth]{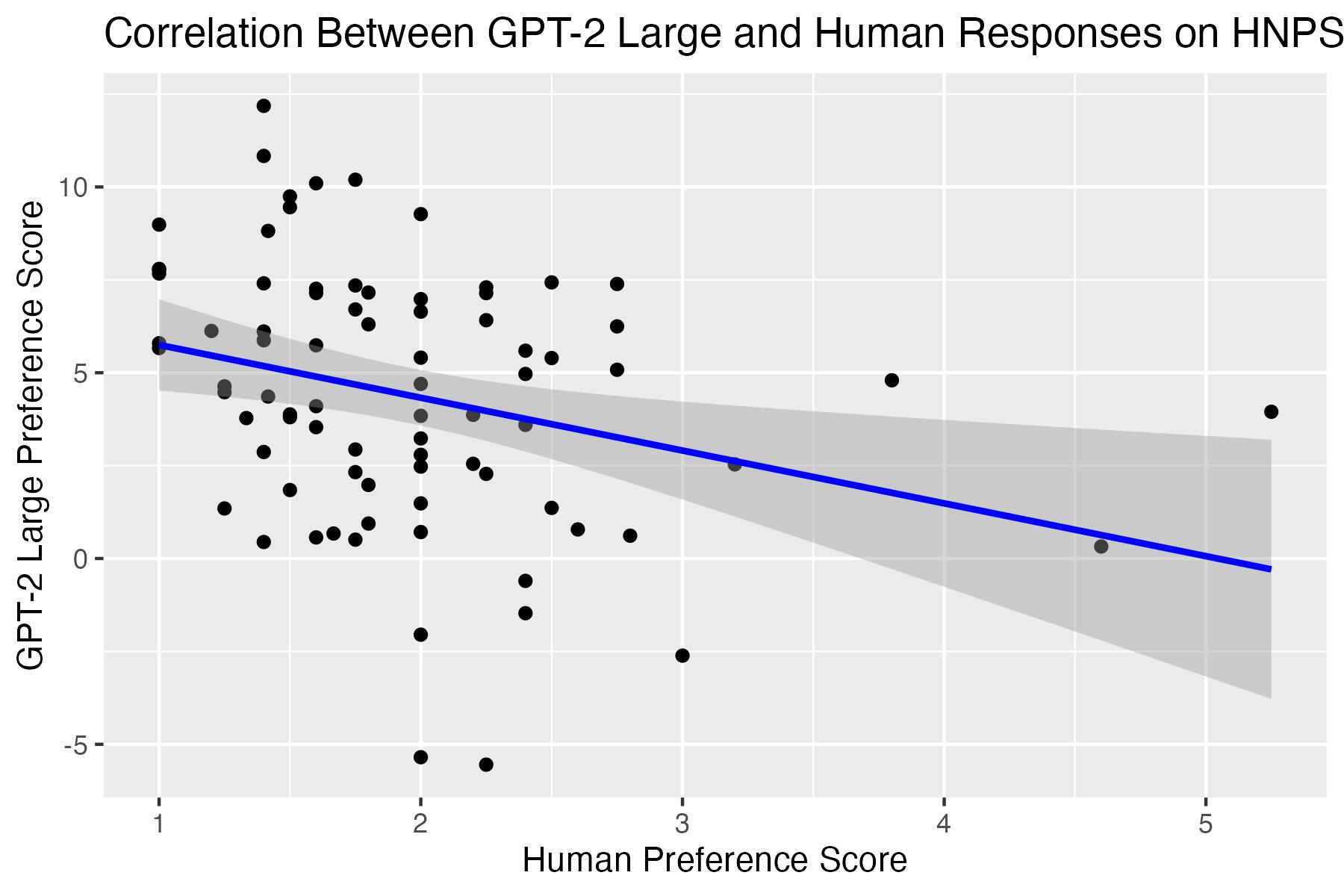}
\end{figure}
\begin{figure}[ht]
    \centering
    \includegraphics[width=\columnwidth]{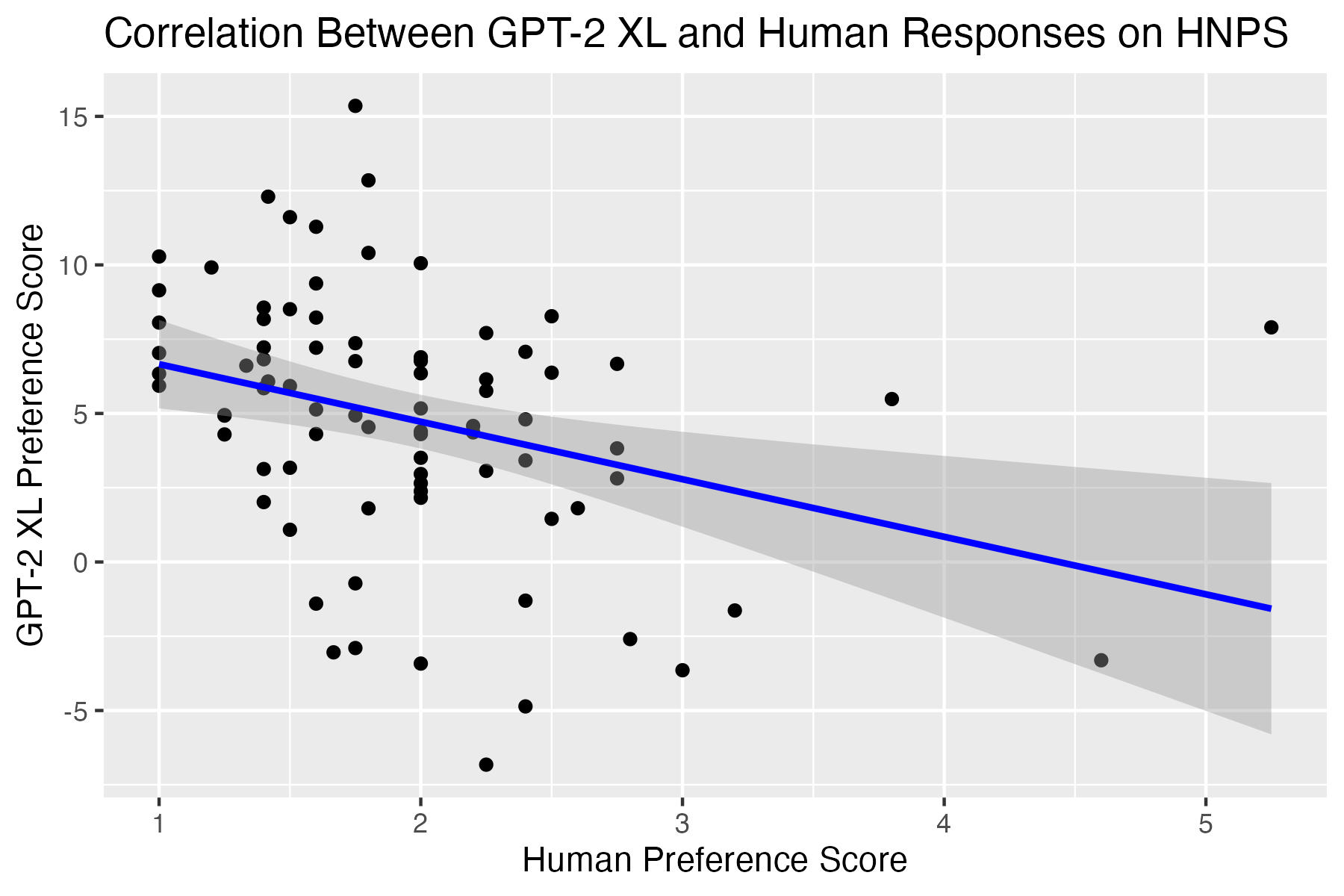}
\end{figure}
\begin{figure}[ht]
    \centering
    \includegraphics[width=\columnwidth]{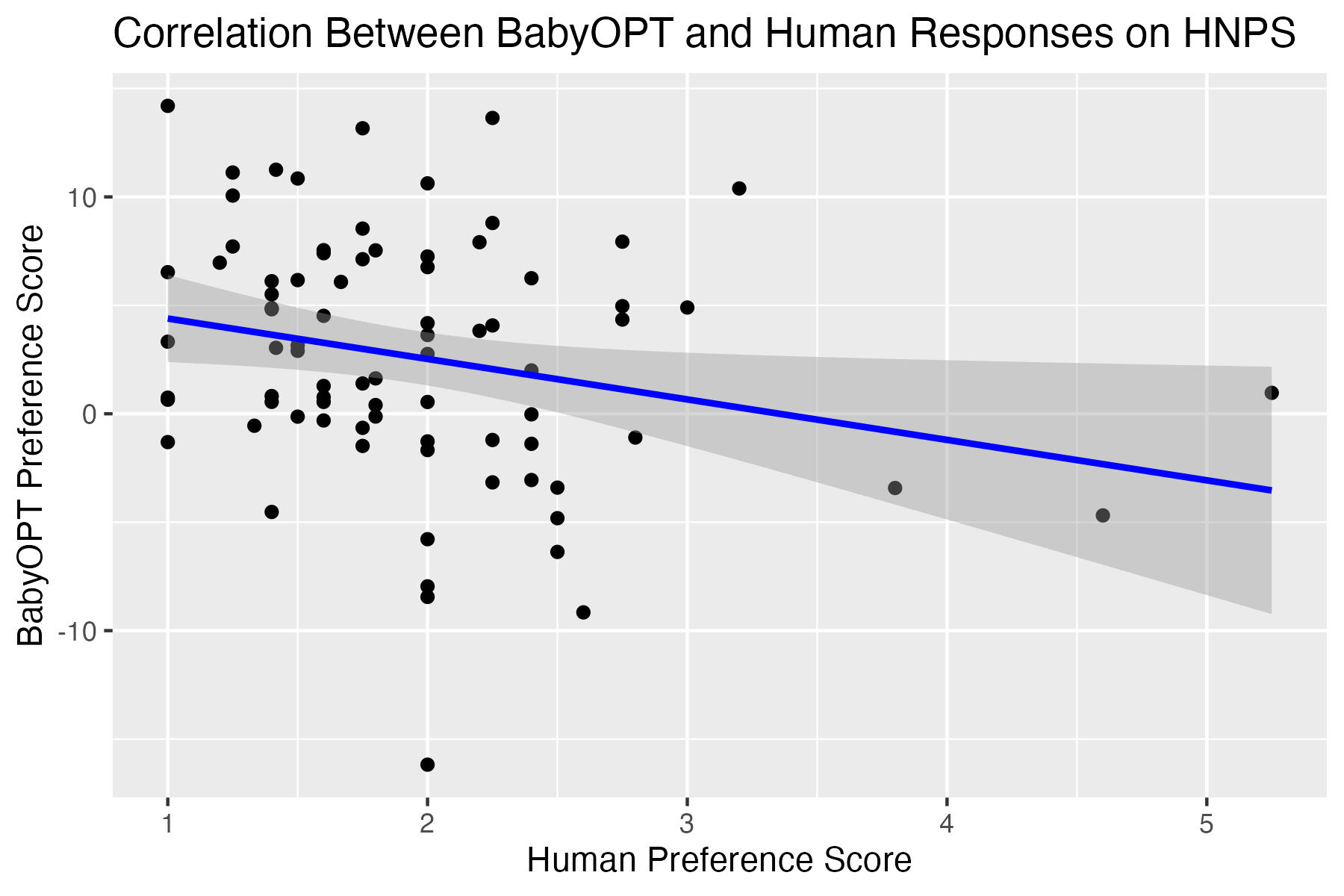}
\end{figure}
\begin{figure}[ht]
    \centering
    \includegraphics[width=\columnwidth]{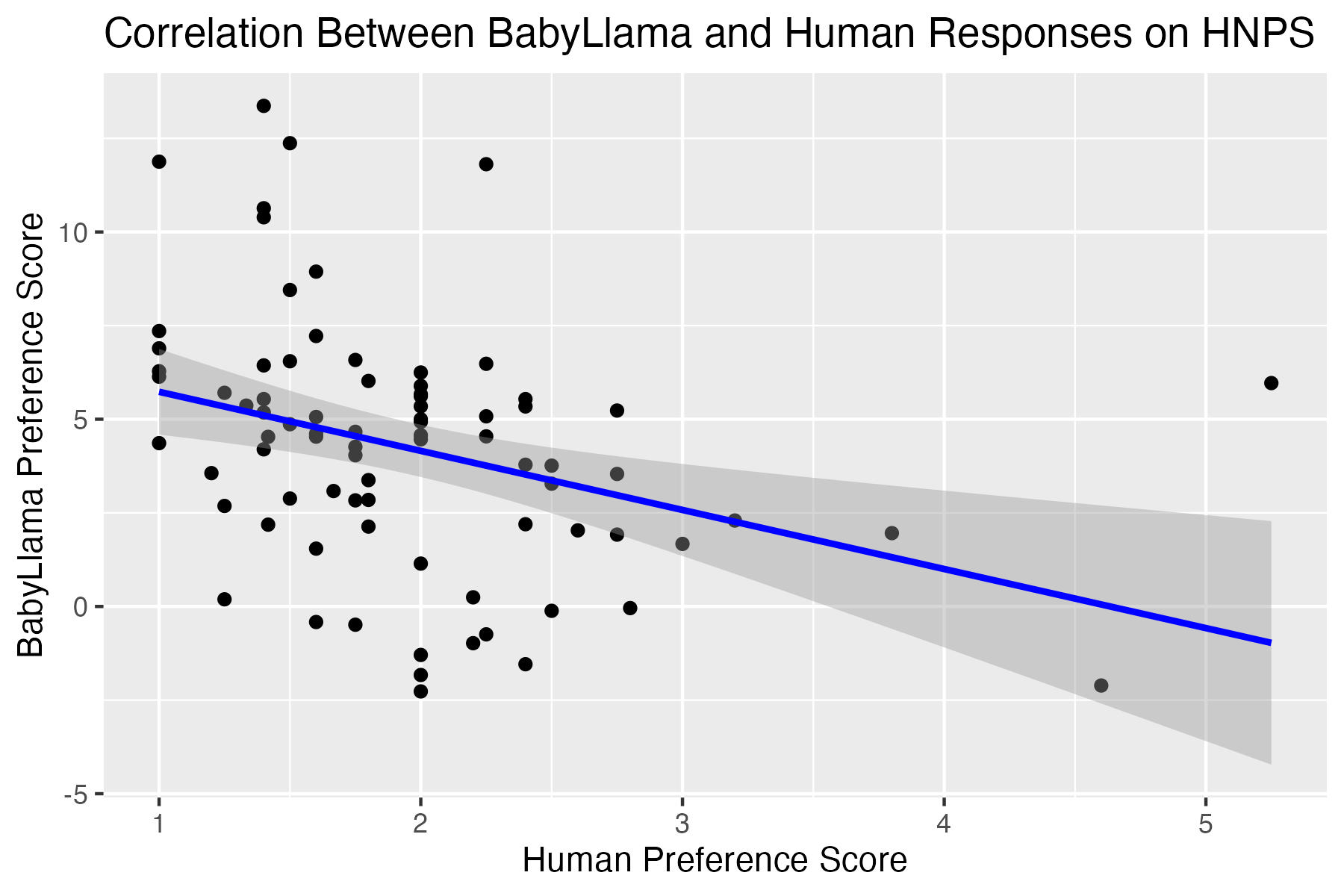}
\end{figure}
\begin{figure}[ht]
    \centering
    \includegraphics[width=\columnwidth]{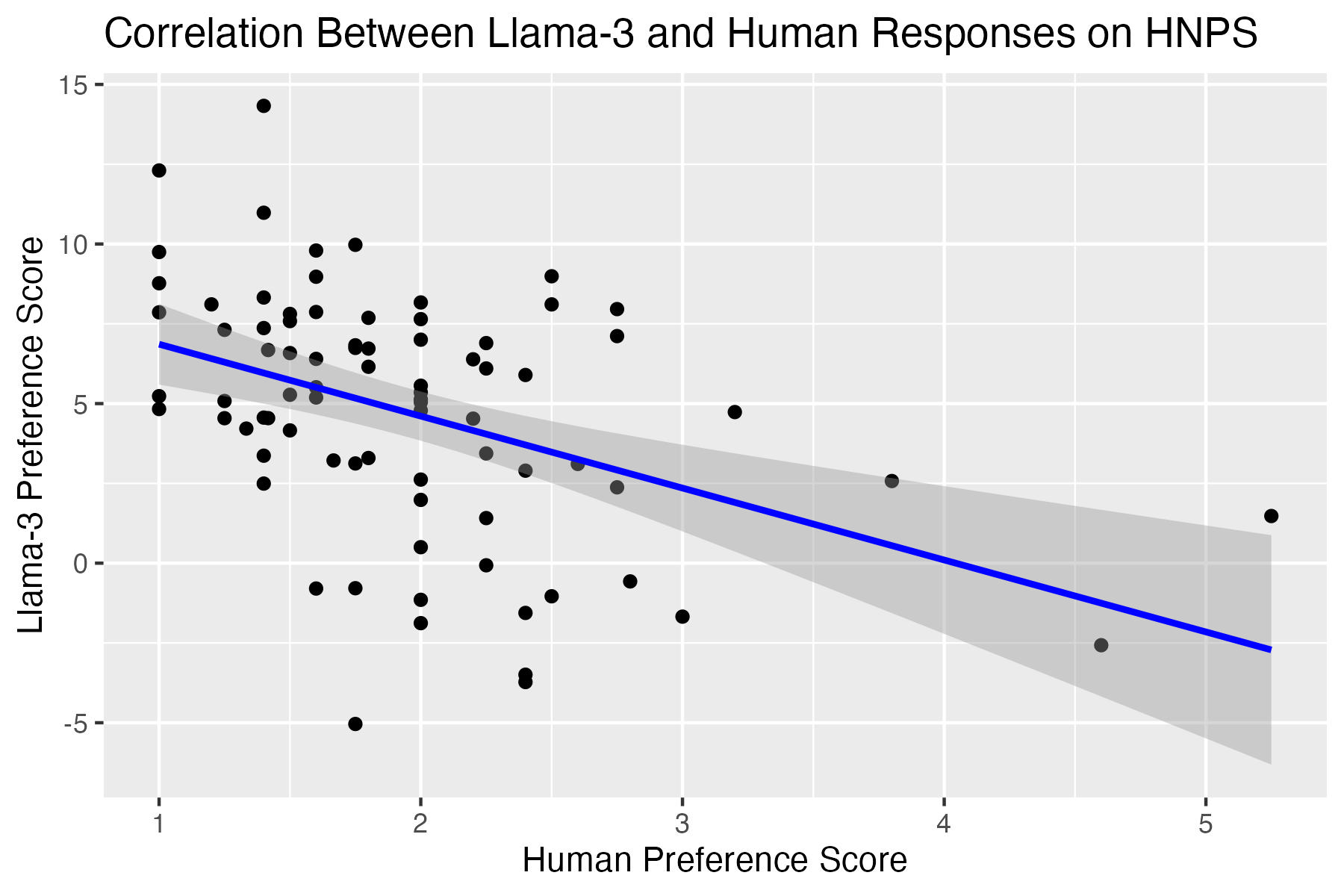}
\end{figure}
\begin{figure}[ht]
    \centering
    \includegraphics[width=\columnwidth]{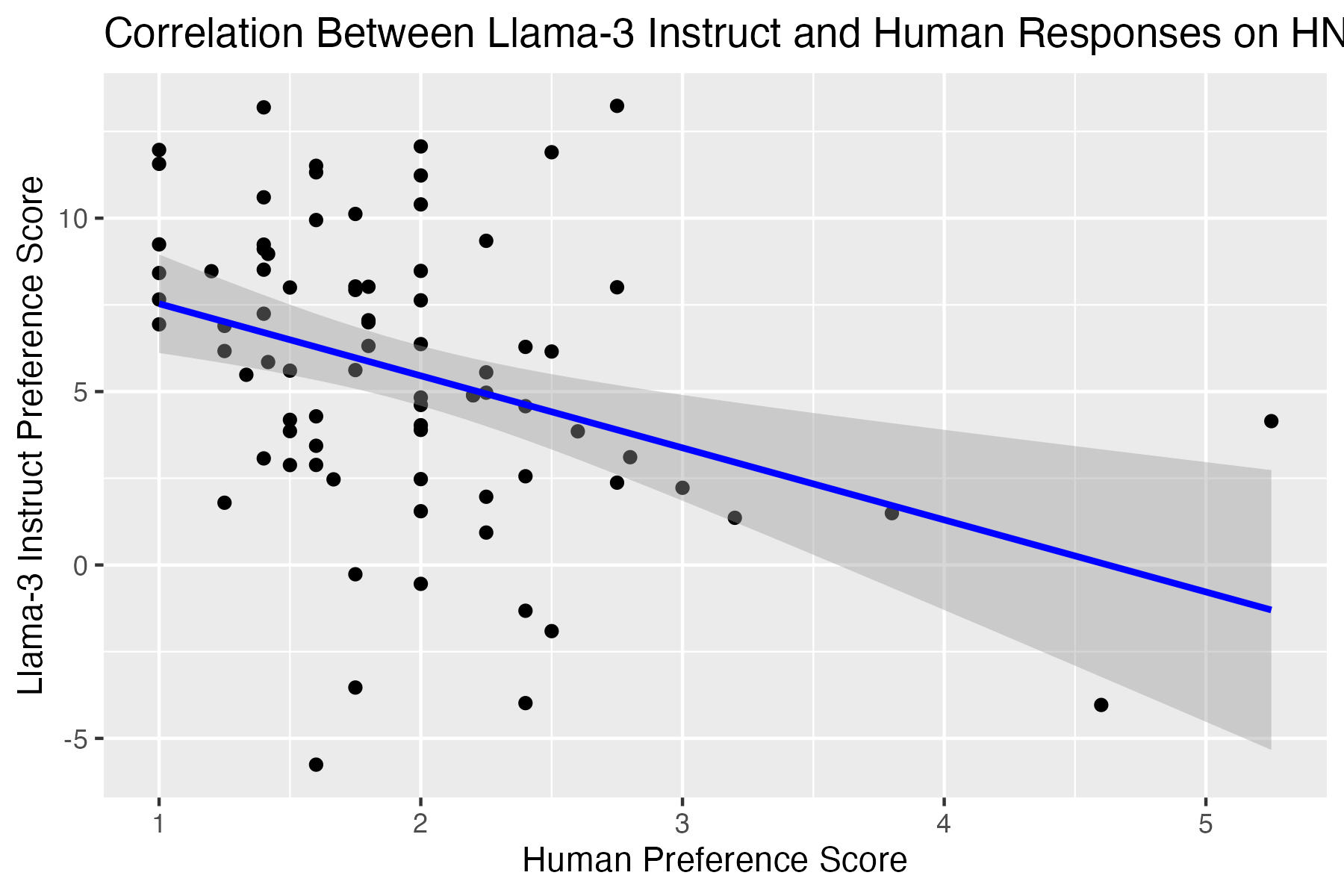}
\end{figure}
\begin{figure}[ht]
    \centering
    \includegraphics[width=\columnwidth]{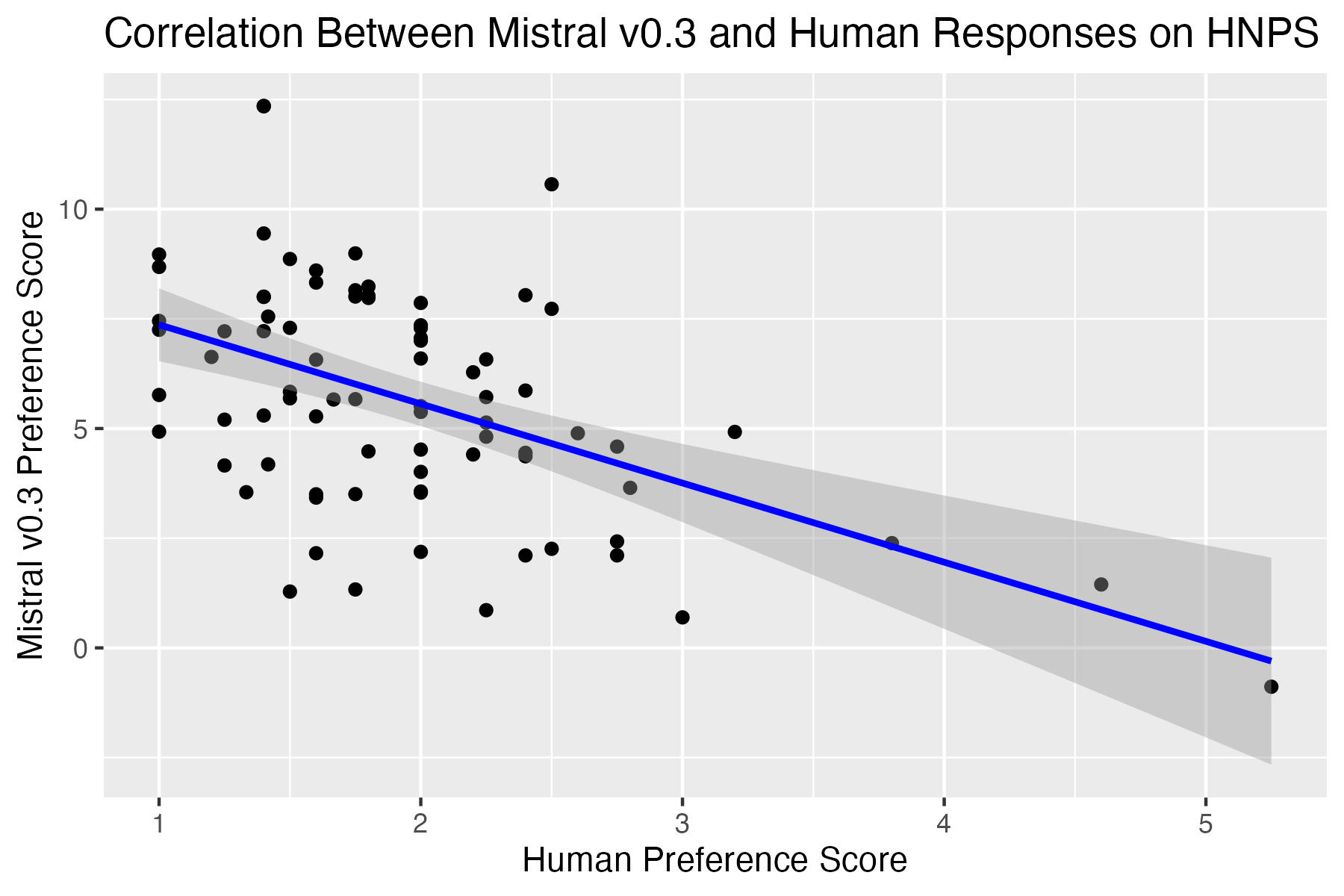}
\end{figure}
\begin{figure}[ht]
    \centering
    \includegraphics[width=\columnwidth]{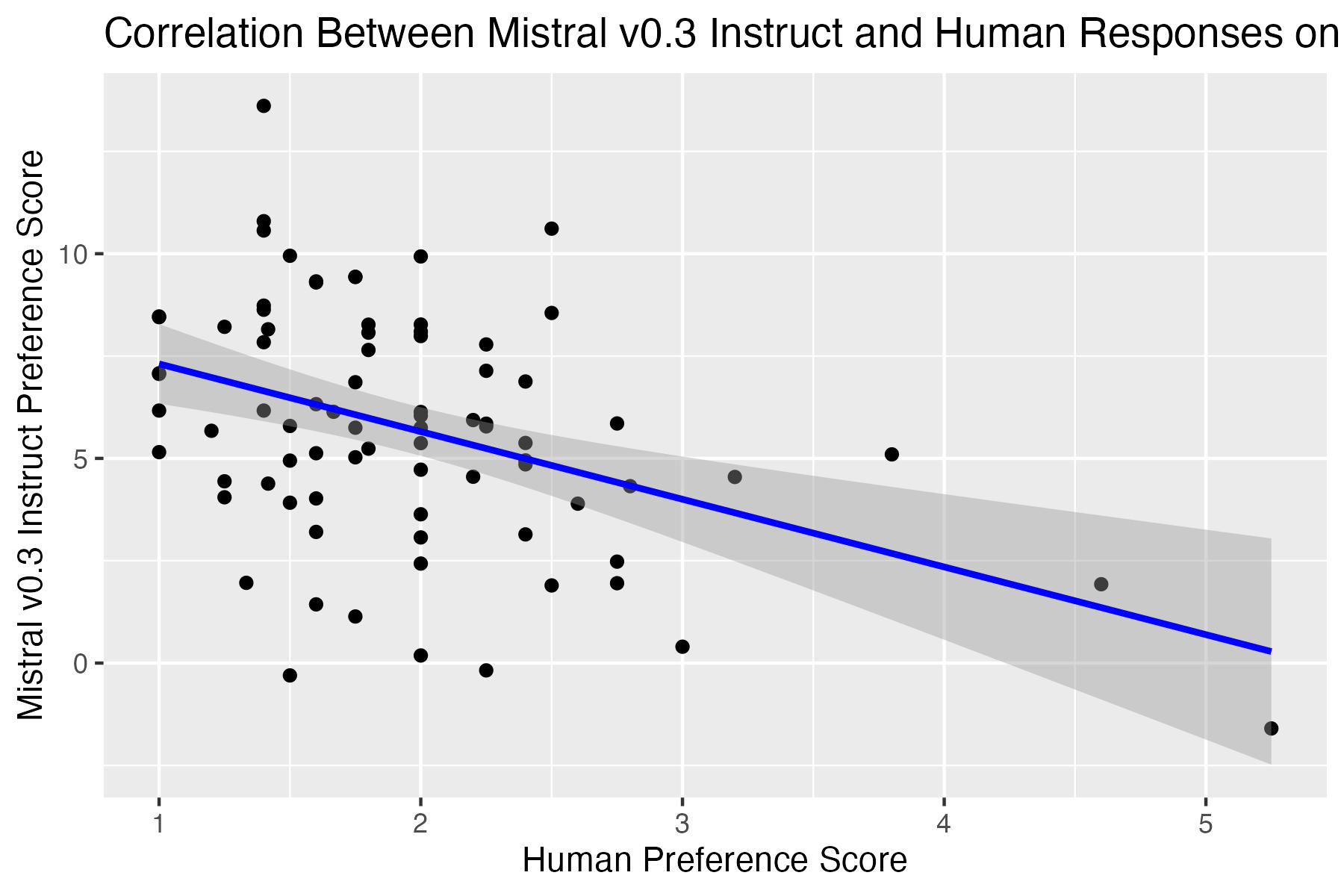}
\end{figure}
\begin{figure}[ht]
    \centering
    \includegraphics[width=\columnwidth]{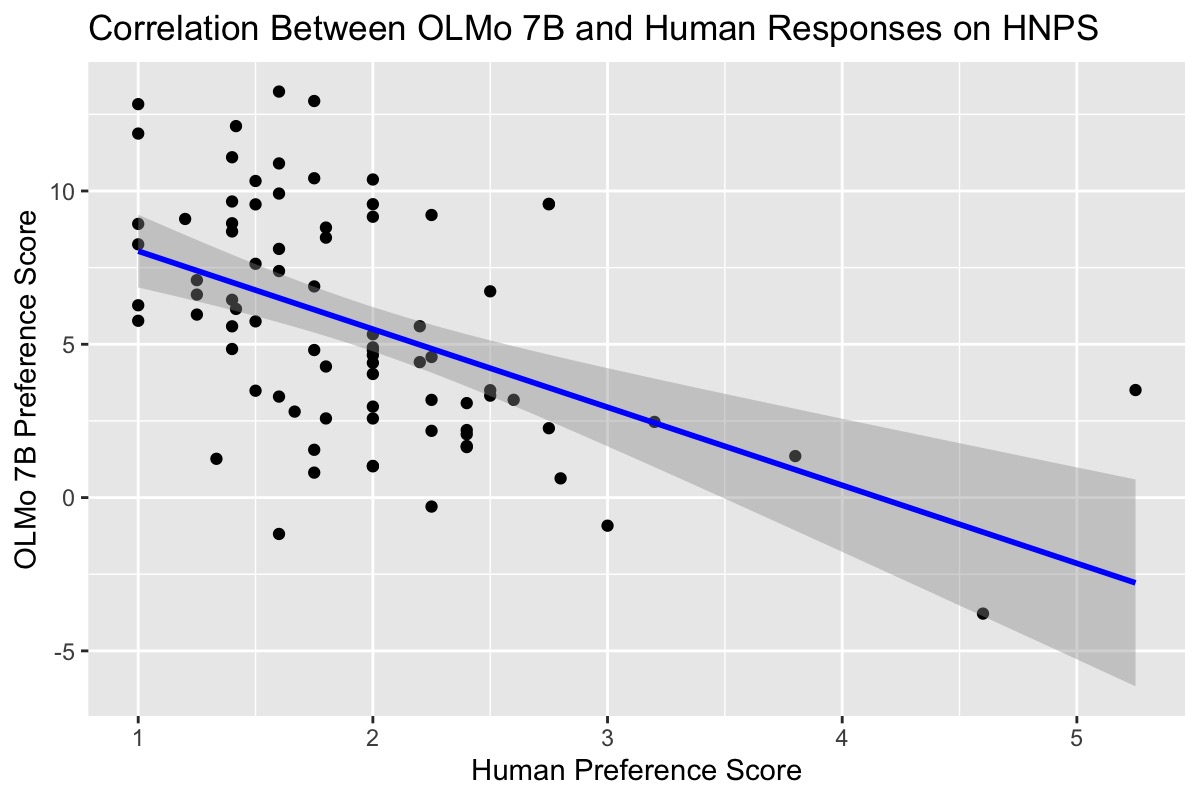}
\end{figure}
\begin{figure}[ht]
    \centering
    \includegraphics[width=\columnwidth]{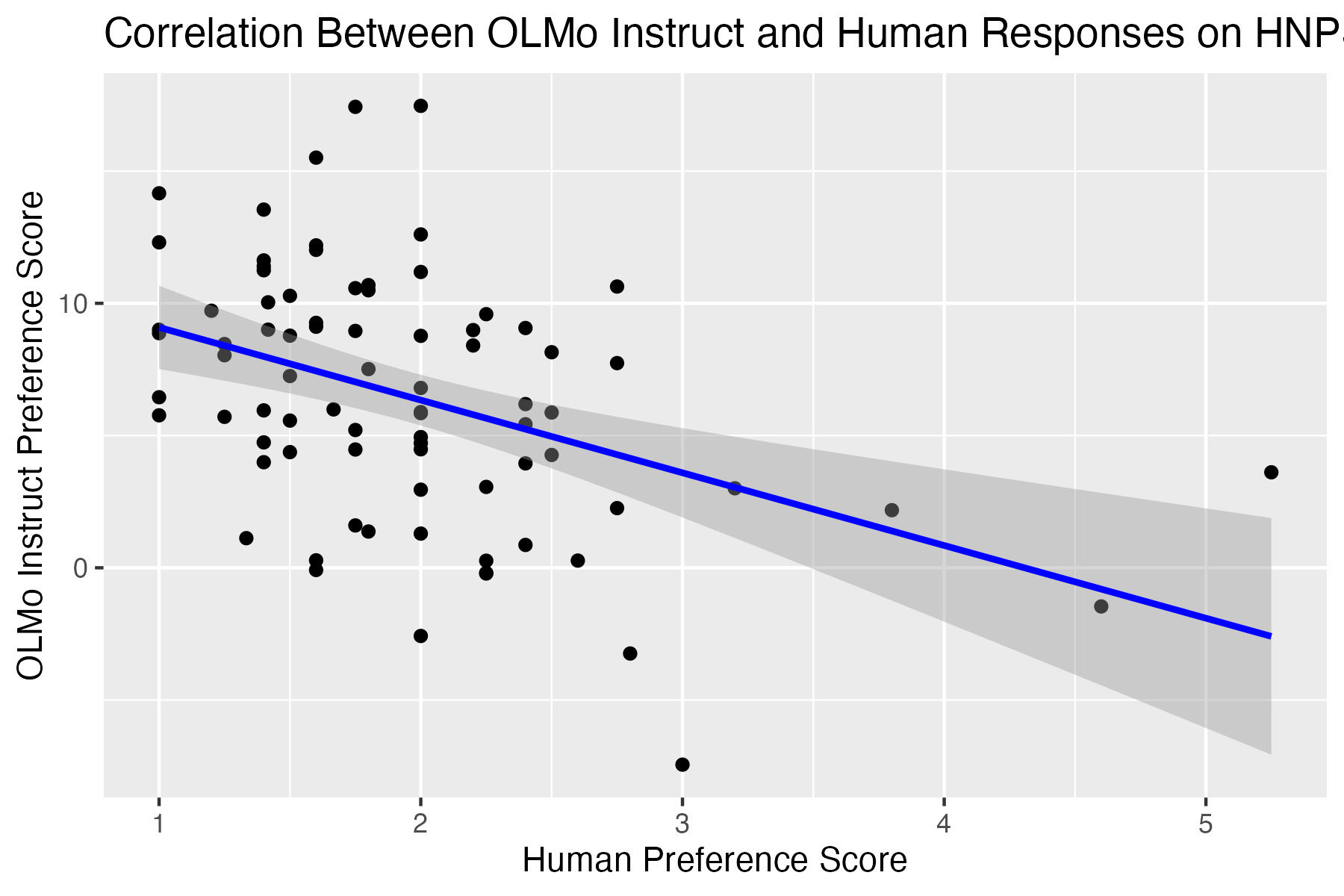}
\end{figure}
\begin{figure}[ht]
    \centering
    \includegraphics[width=\columnwidth]{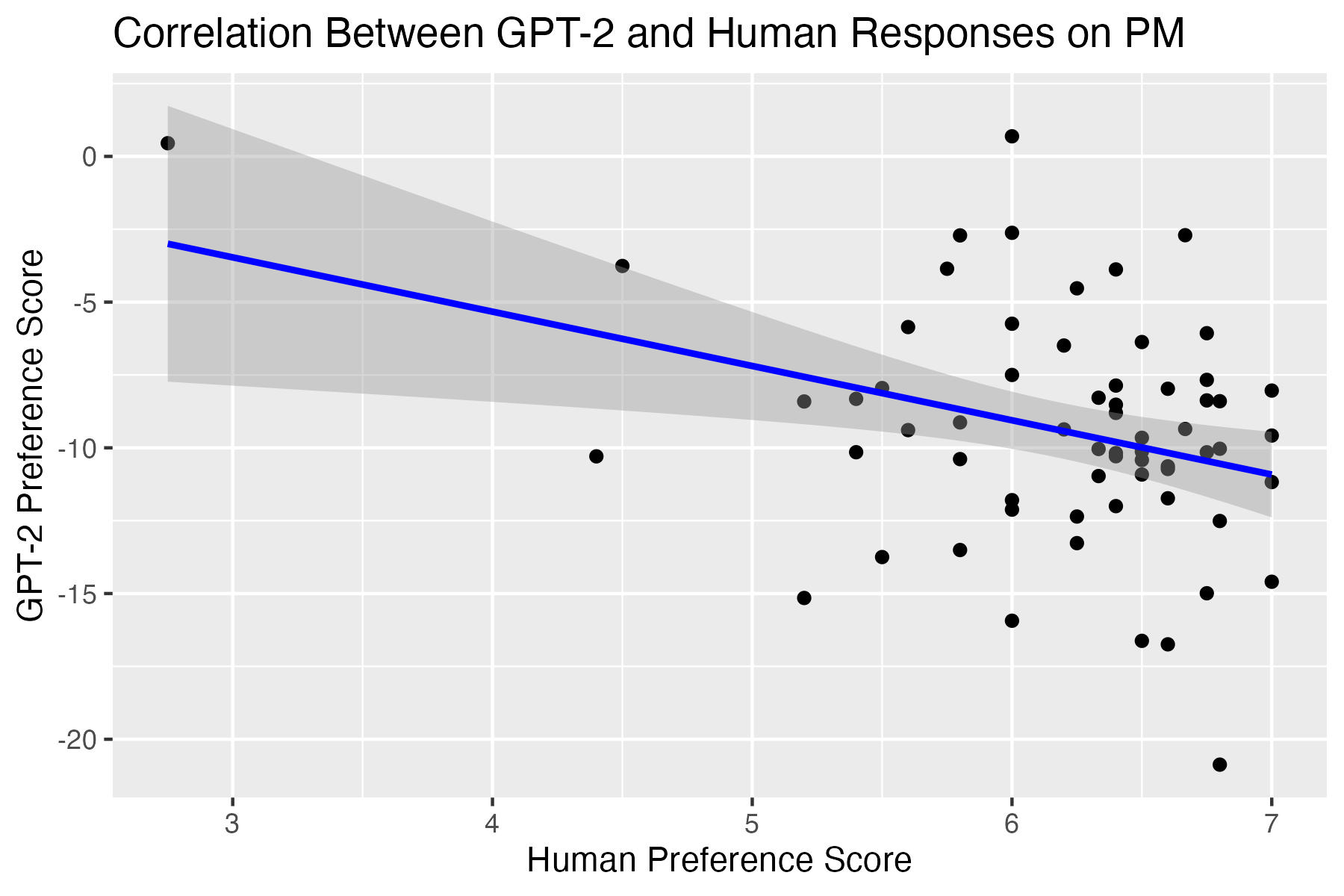}
\end{figure}
\begin{figure}[ht]
    \centering
    \includegraphics[width=\columnwidth]{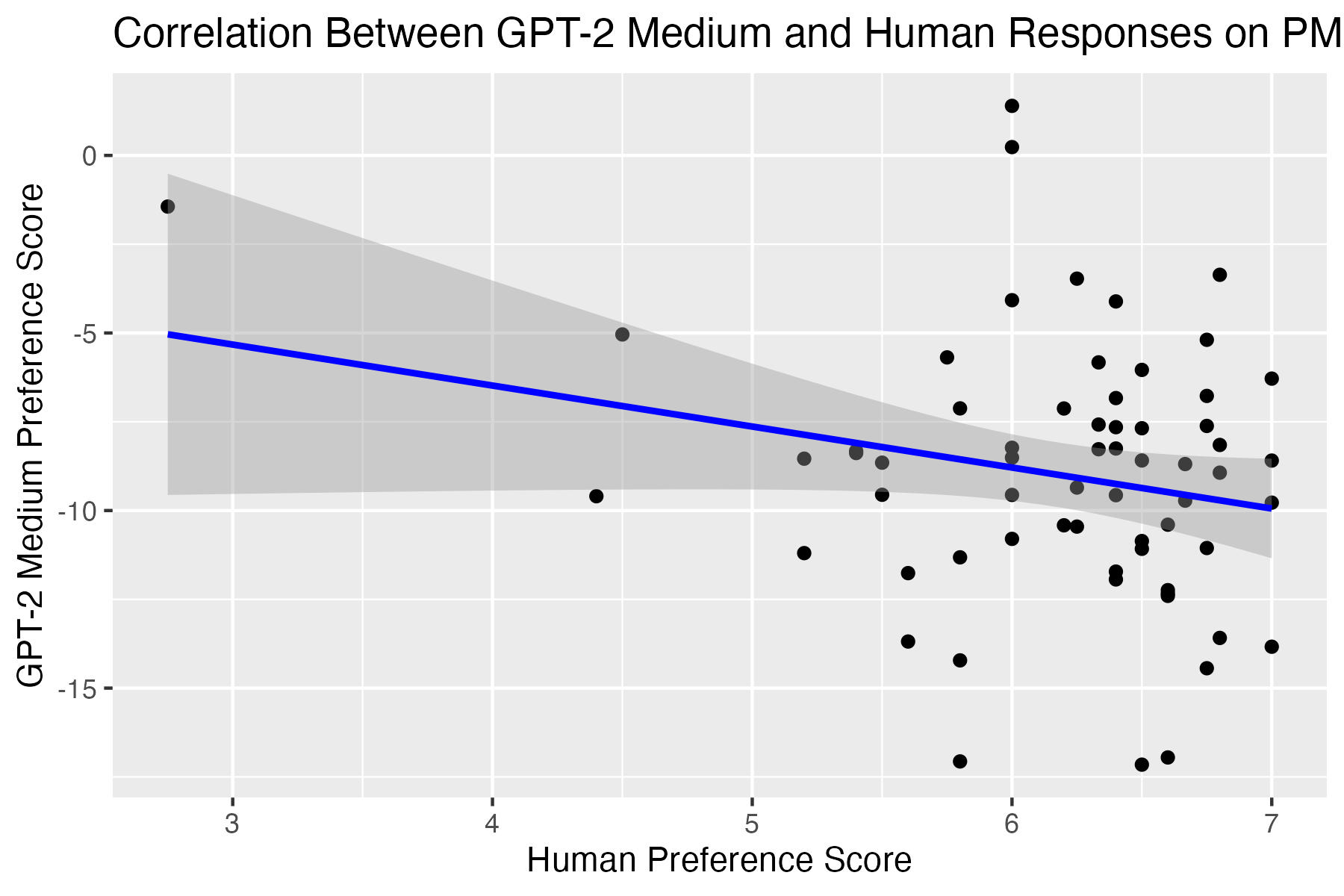}
\end{figure}
\begin{figure}[ht]
    \centering
    \includegraphics[width=\columnwidth]{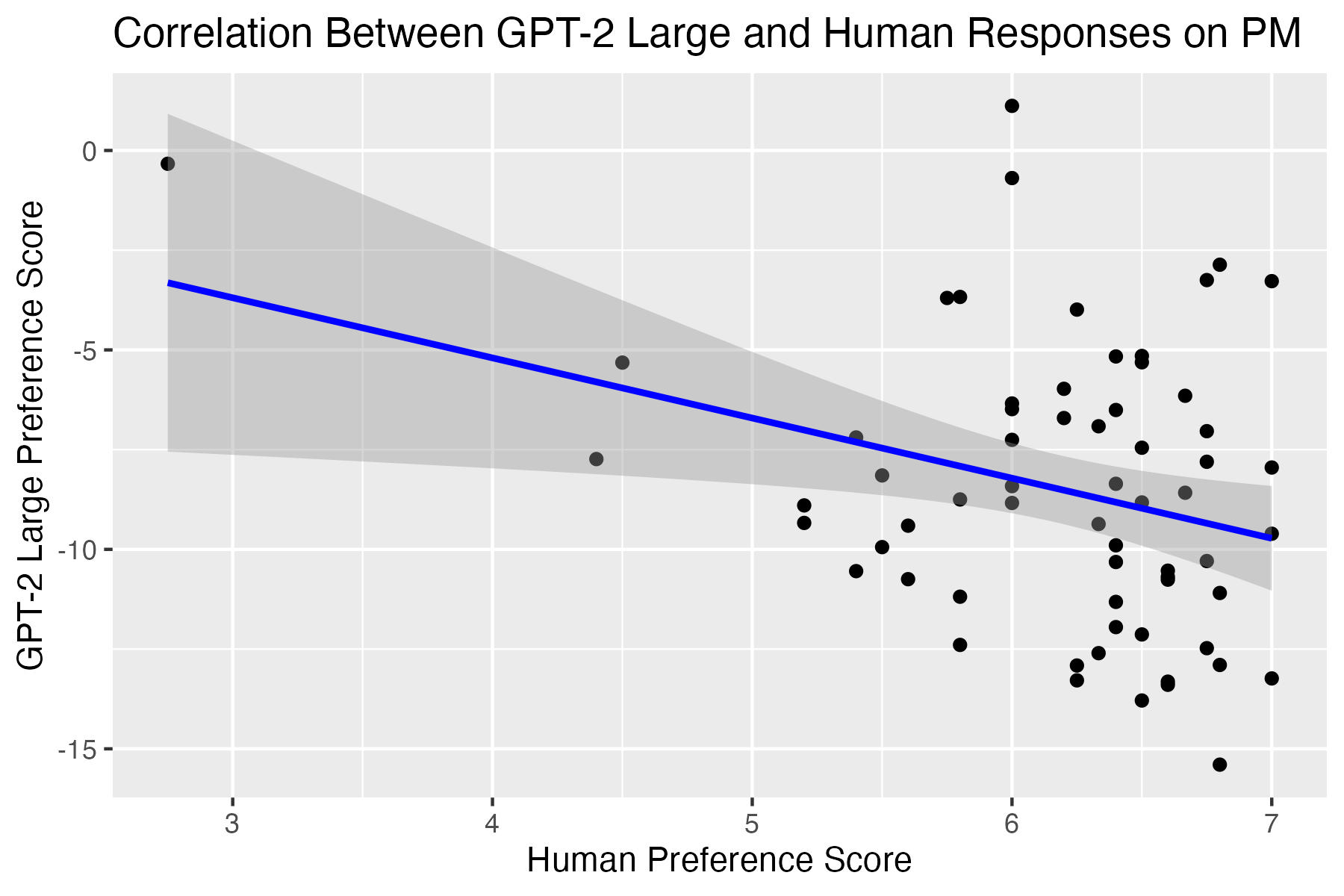}
\end{figure}
\begin{figure}[ht]
    \centering
    \includegraphics[width=\columnwidth]{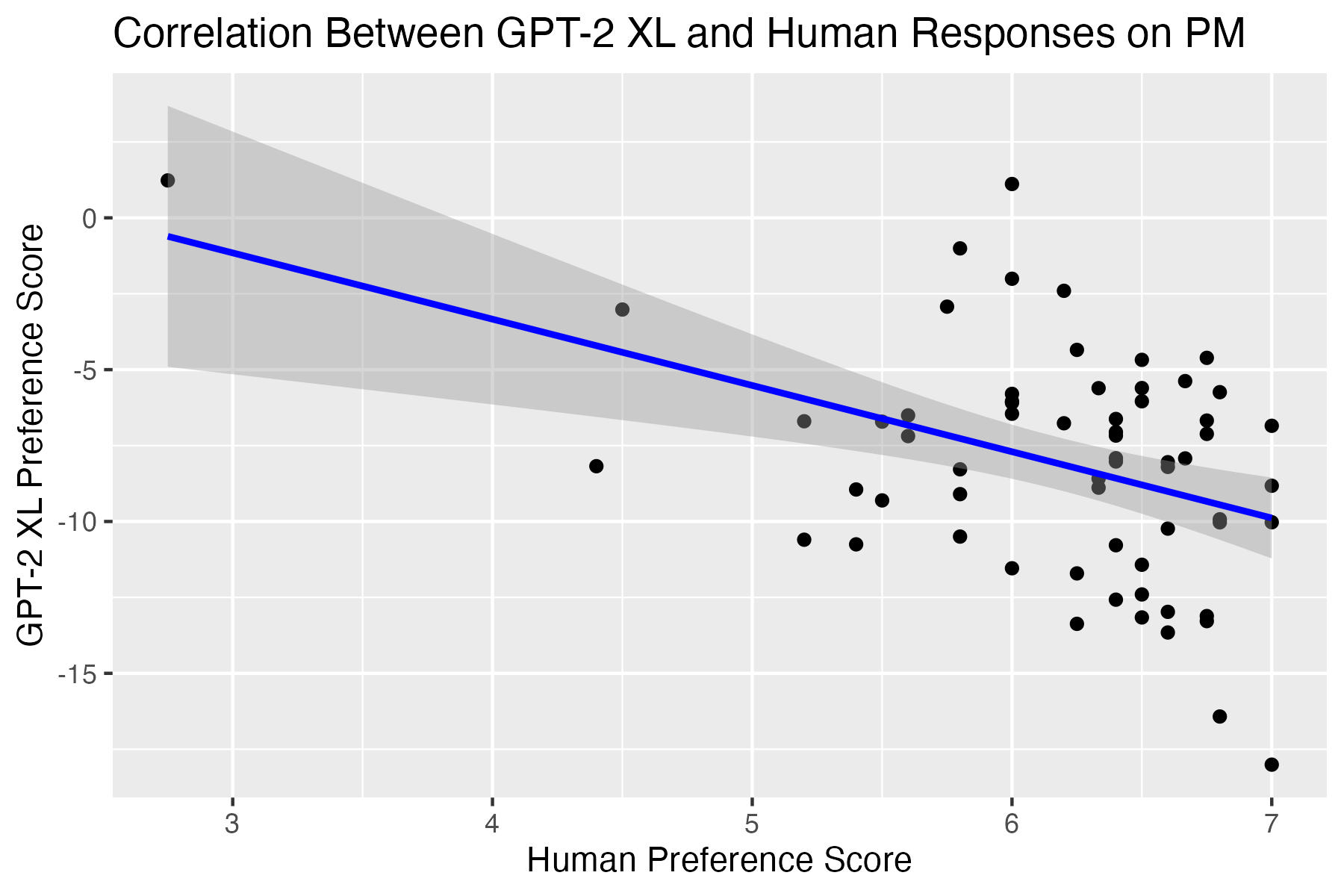}
\end{figure}
\begin{figure}[ht]
    \centering
    \includegraphics[width=\columnwidth]{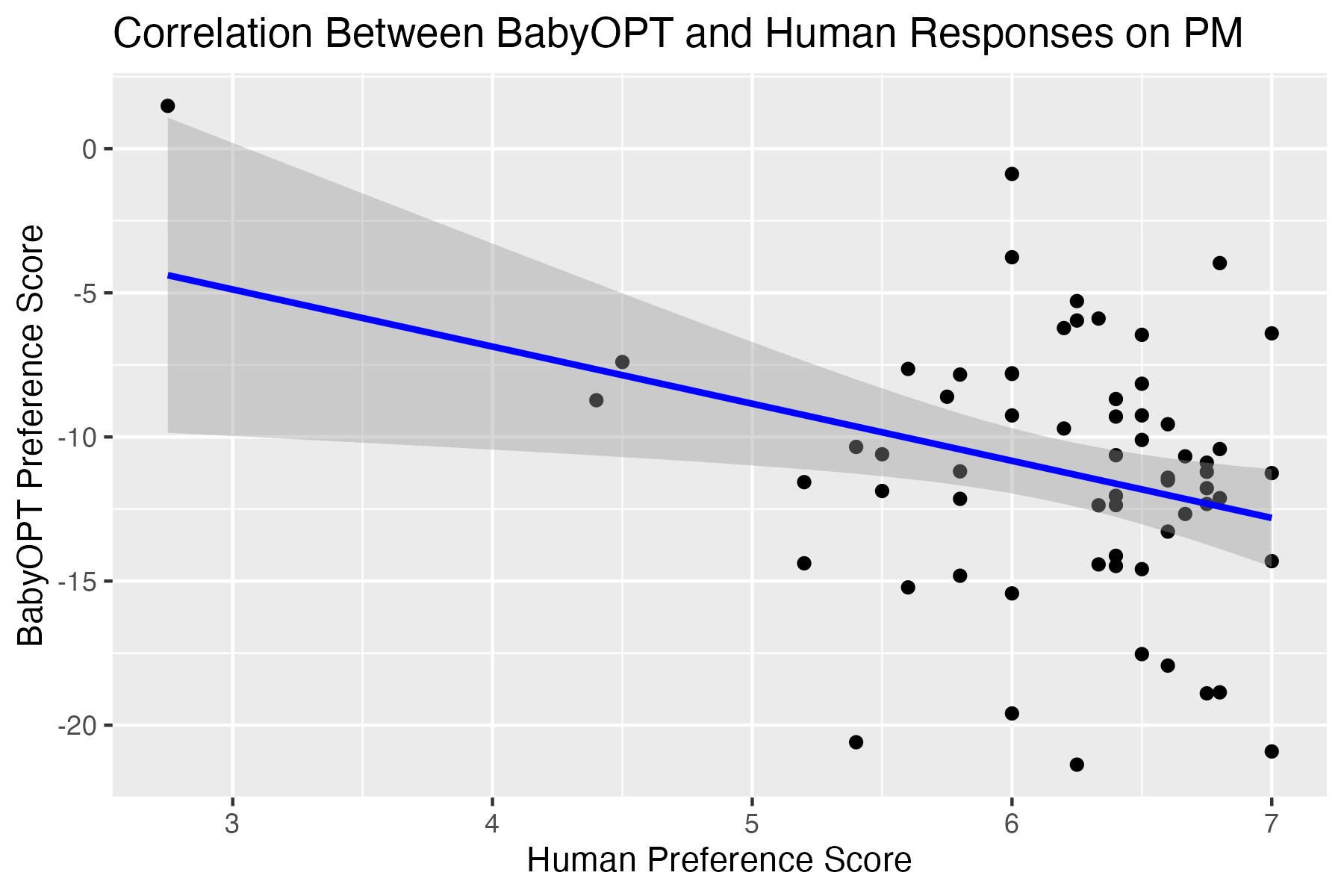}
\end{figure}
\begin{figure}[ht]
    \centering
    \includegraphics[width=\columnwidth]{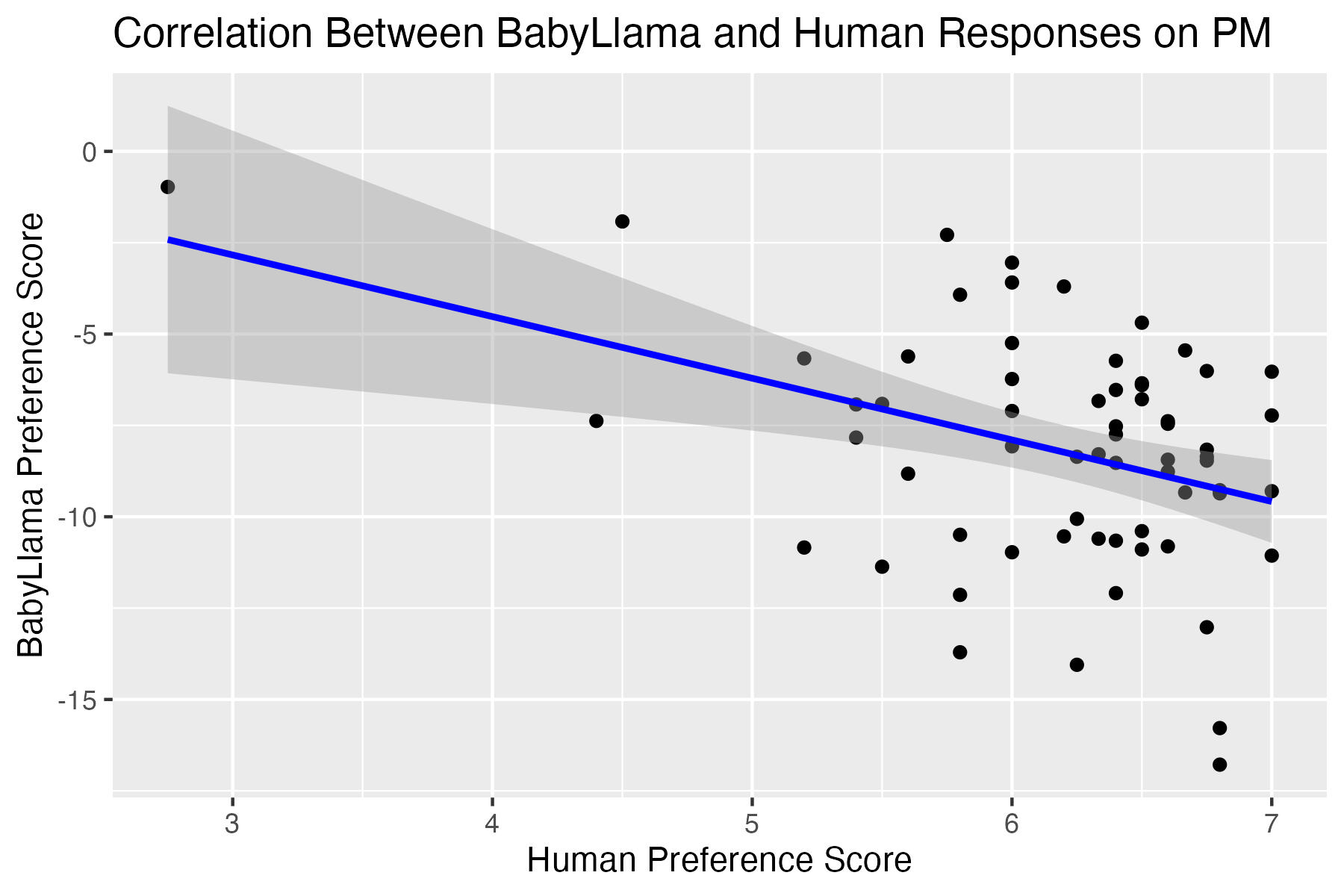}
\end{figure}
\begin{figure}[ht]
    \centering
    \includegraphics[width=\columnwidth]{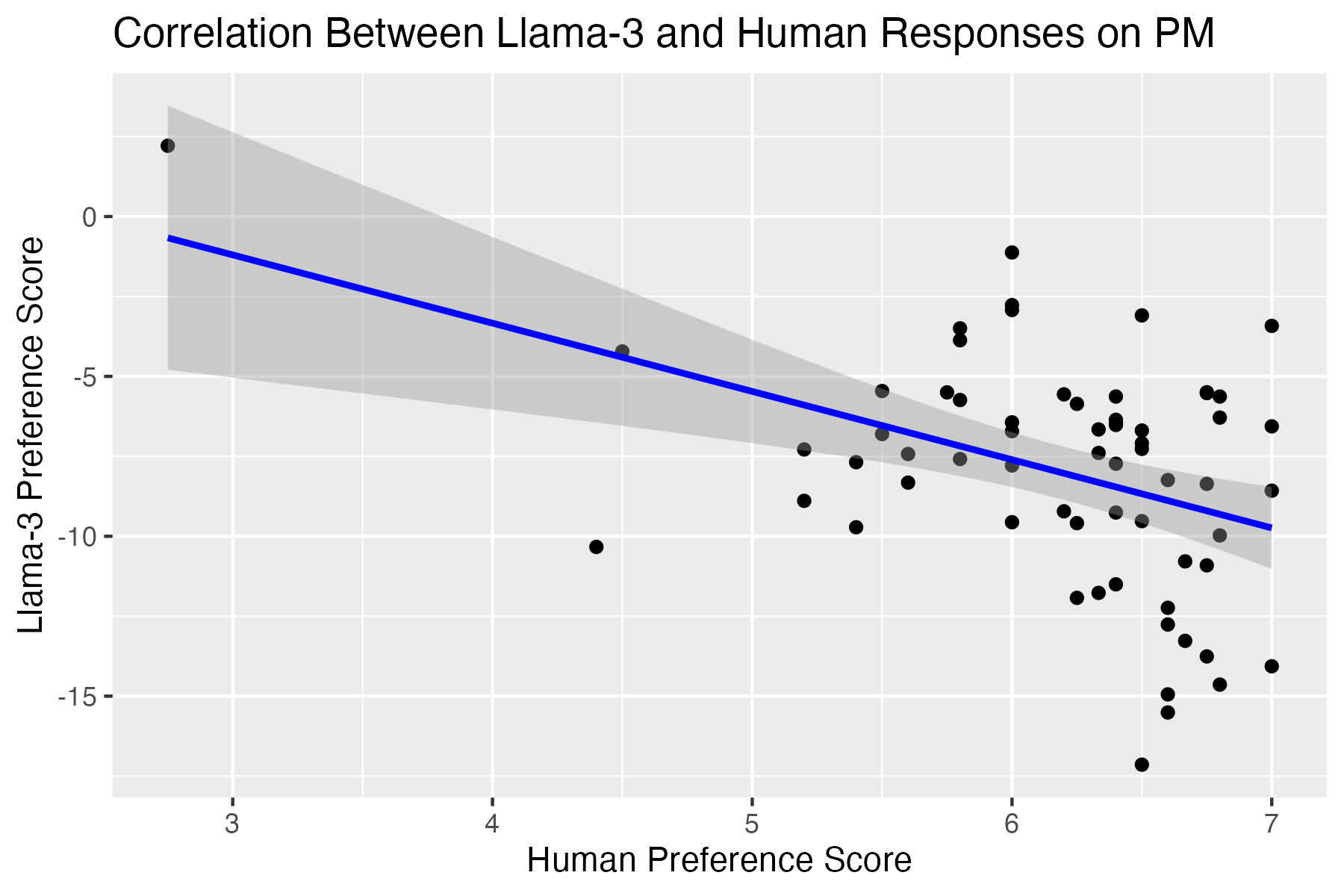}
\end{figure}
\begin{figure}[ht]
    \centering
    \includegraphics[width=\columnwidth]{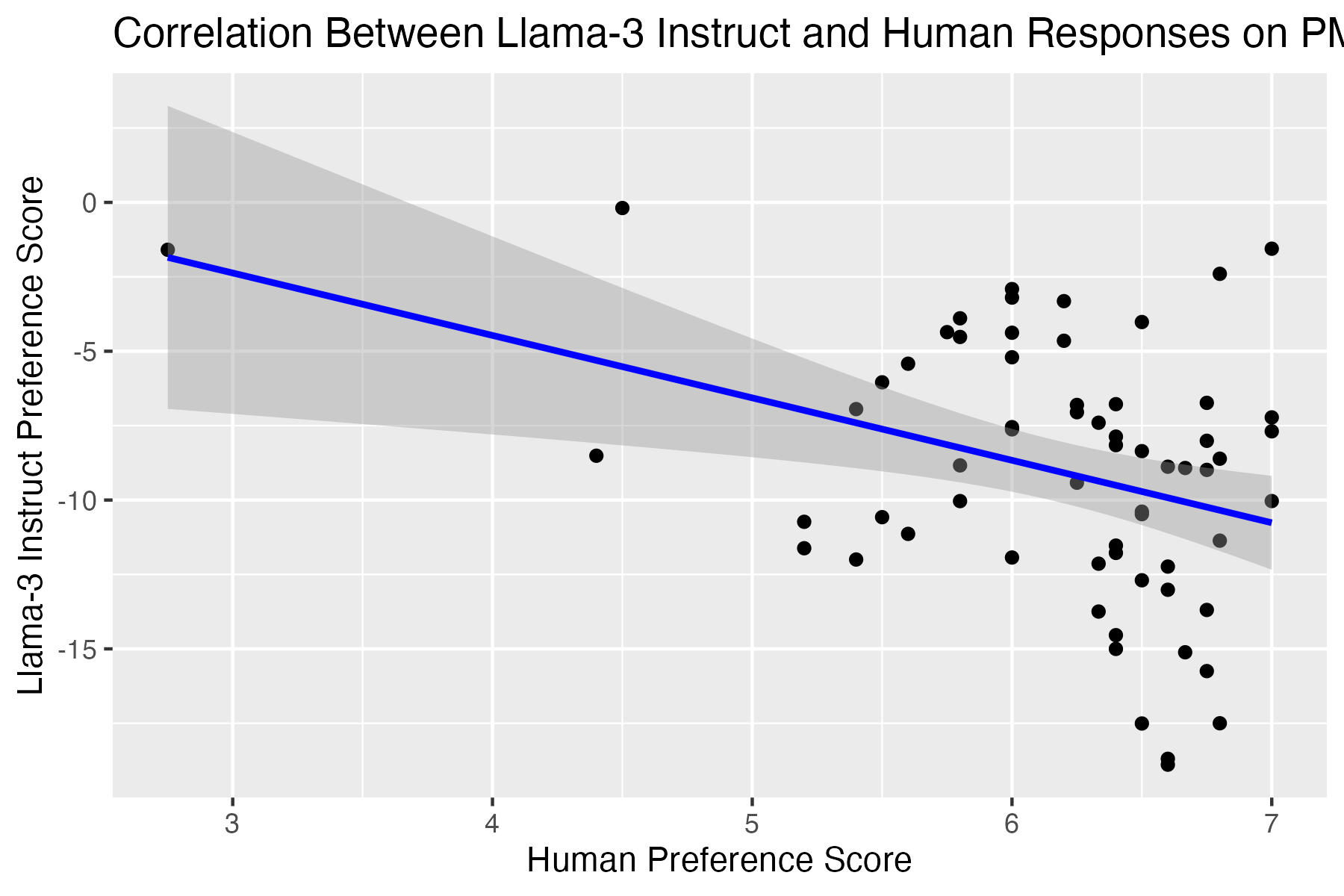}
\end{figure}
\begin{figure}[ht]
    \centering
    \includegraphics[width=\columnwidth]{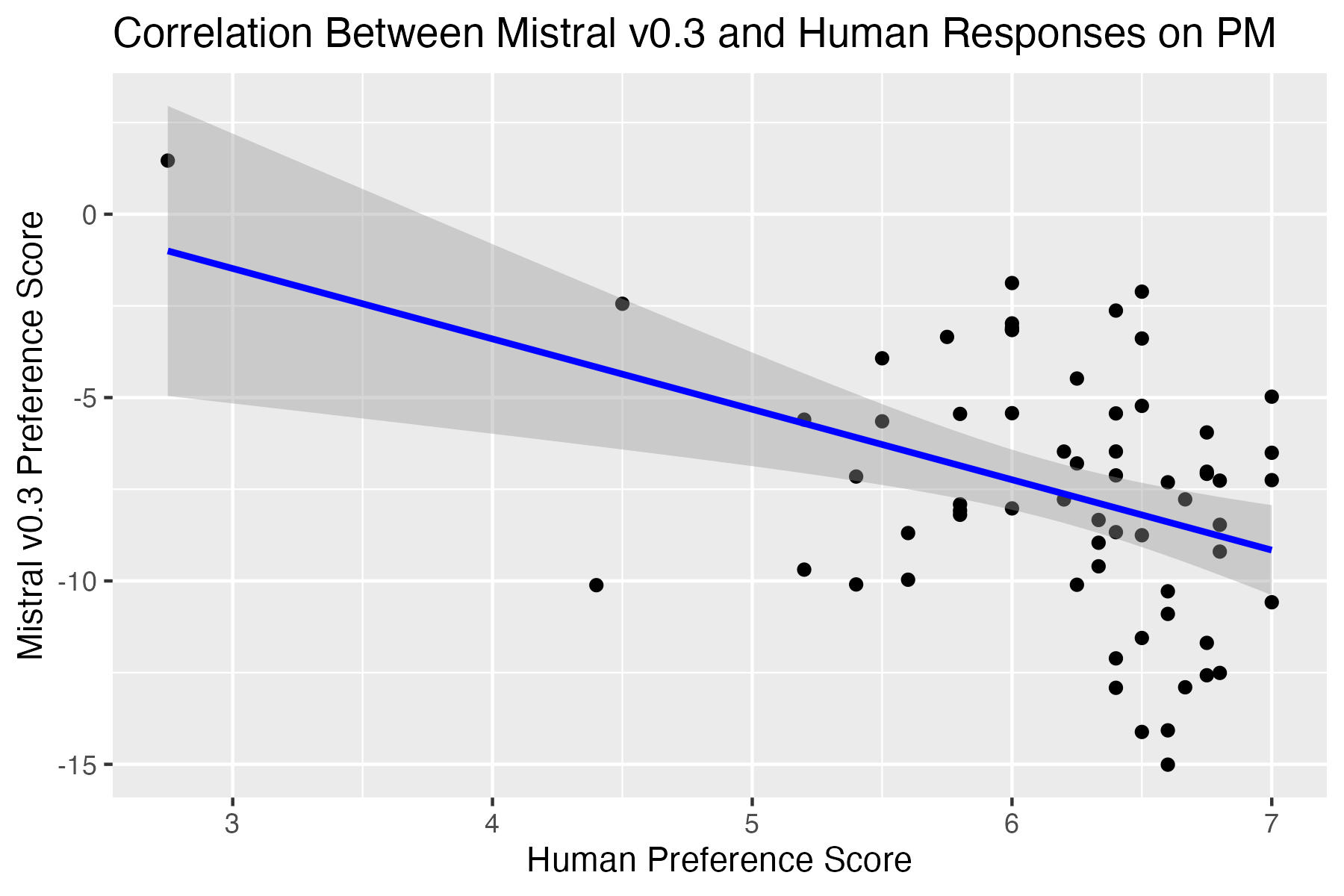}
\end{figure}
\begin{figure}[ht]
    \centering
    \includegraphics[width=\columnwidth]{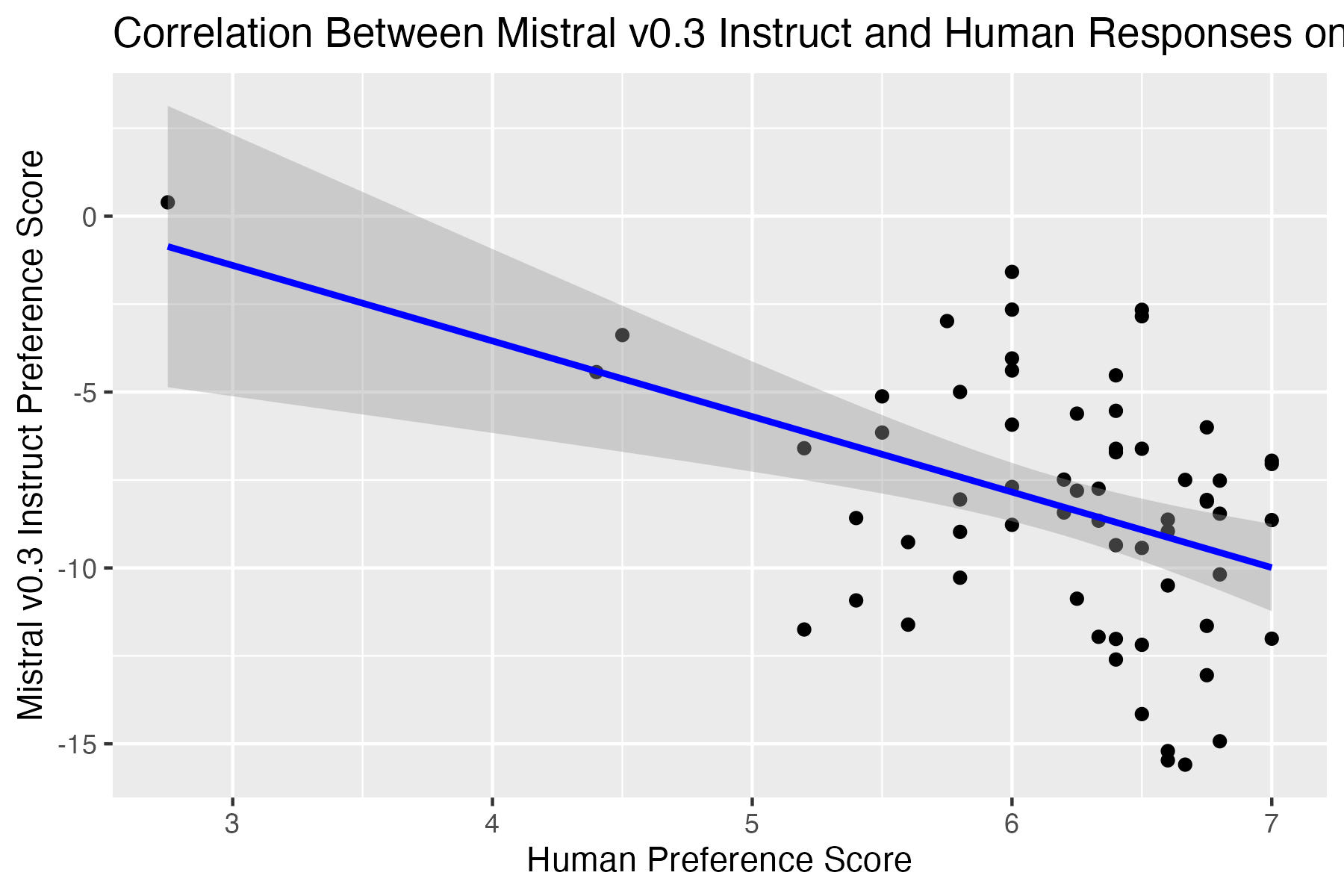}
\end{figure}
\begin{figure}[ht]
    \centering
    \includegraphics[width=\columnwidth]{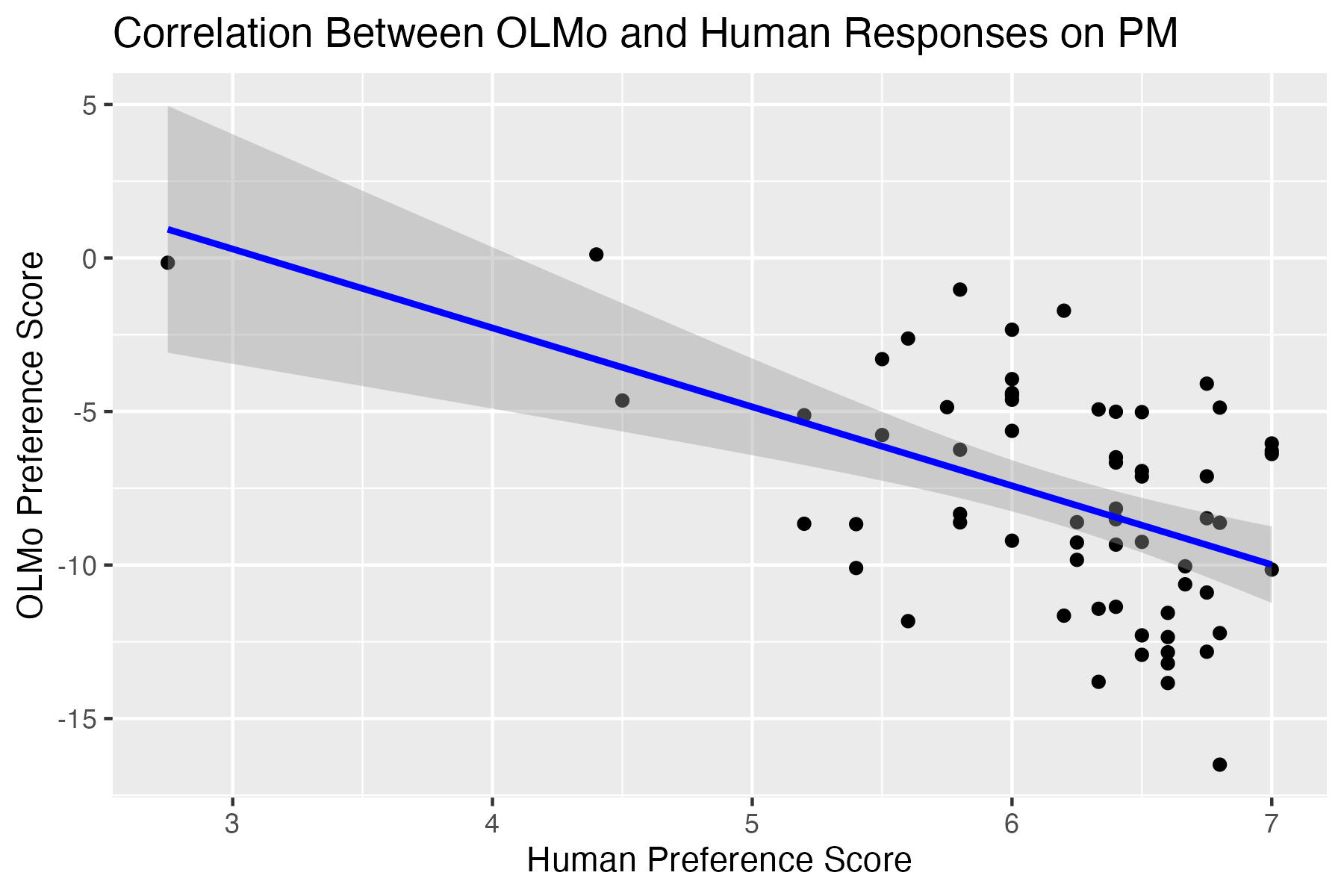}
\end{figure}
\begin{figure}[ht]
    \centering
    \includegraphics[width=\columnwidth]{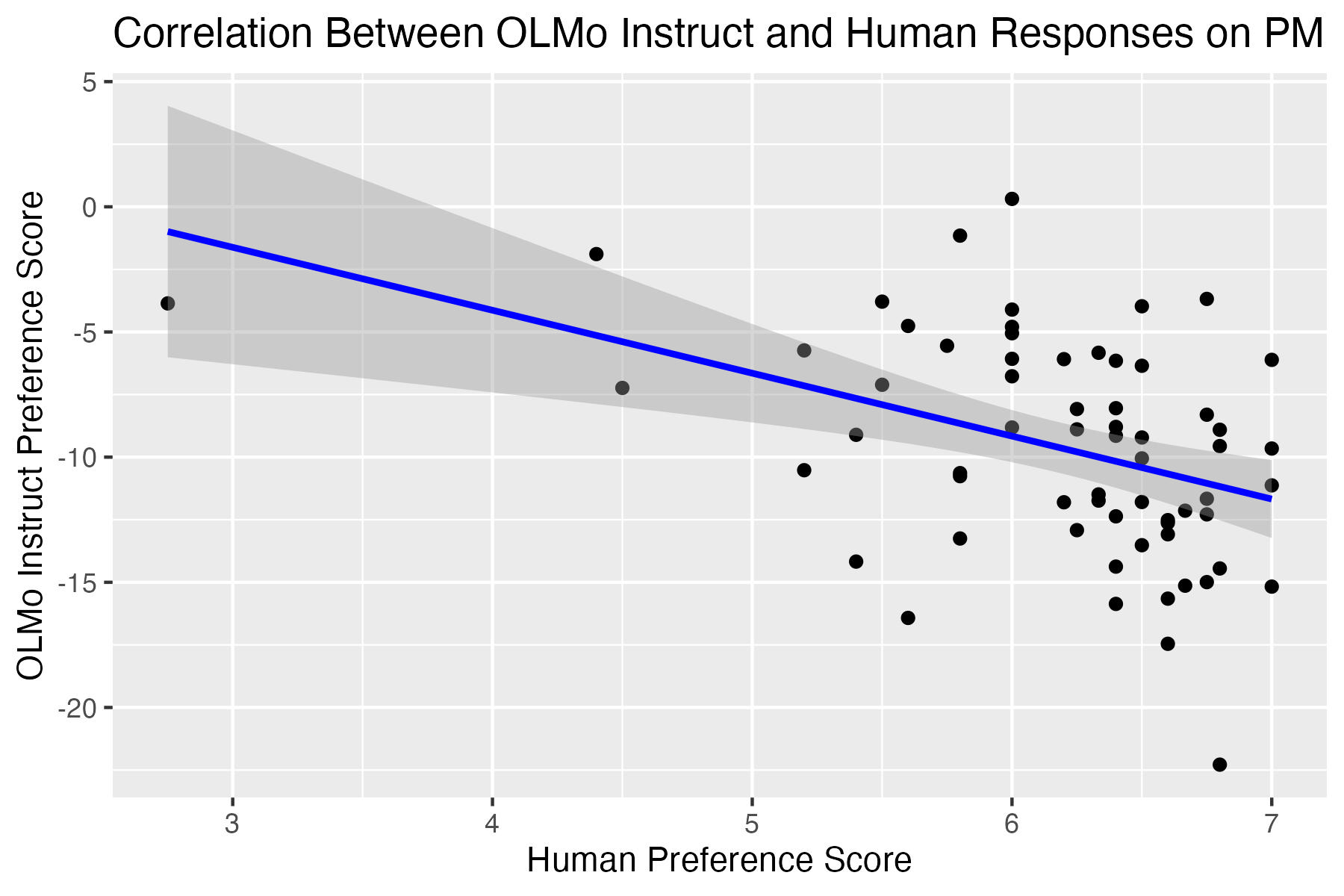}
\end{figure}
\begin{figure}[ht]
    \centering
    \includegraphics[width=\columnwidth]{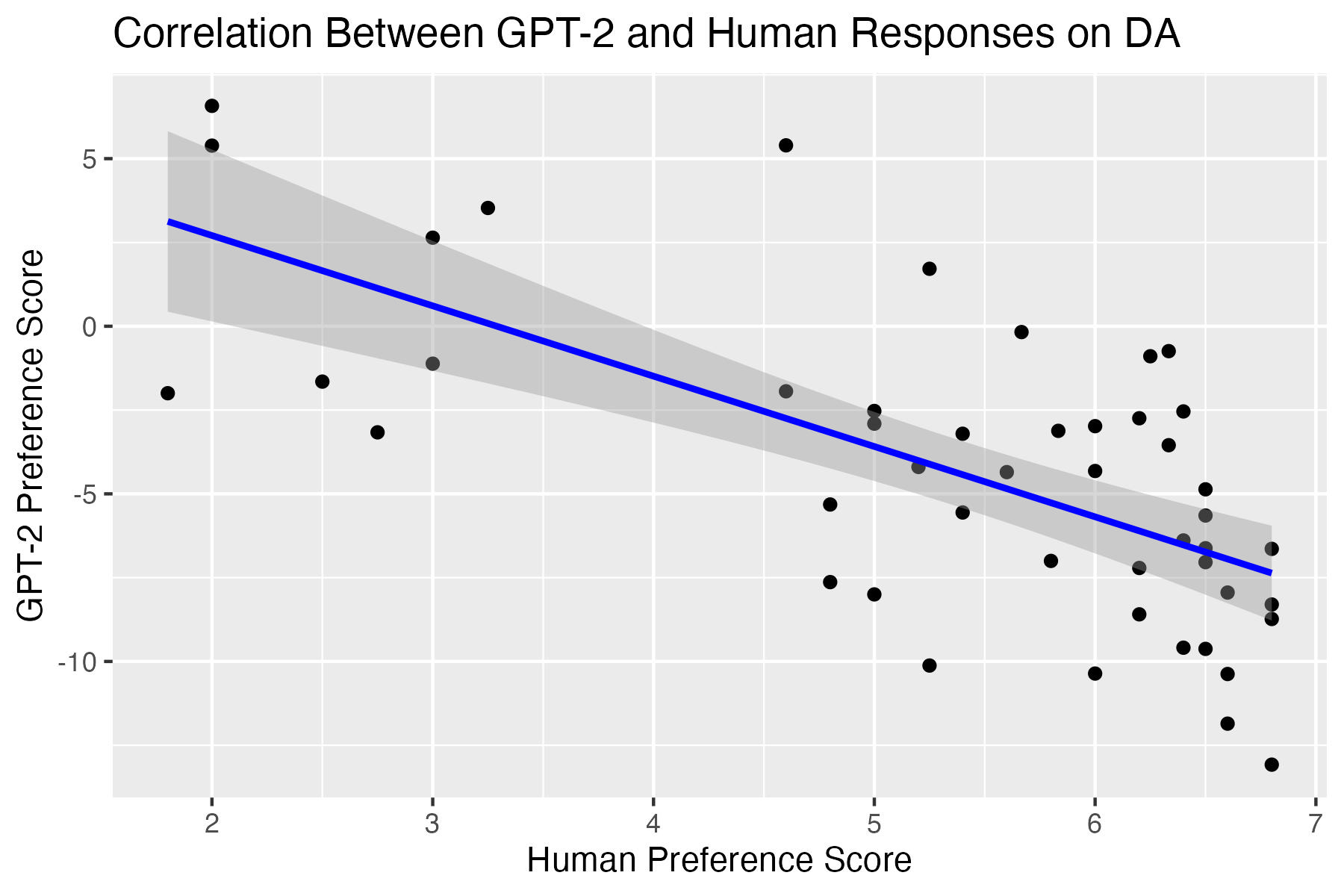}
\end{figure}
\begin{figure}[ht]
    \centering
    \includegraphics[width=\columnwidth]{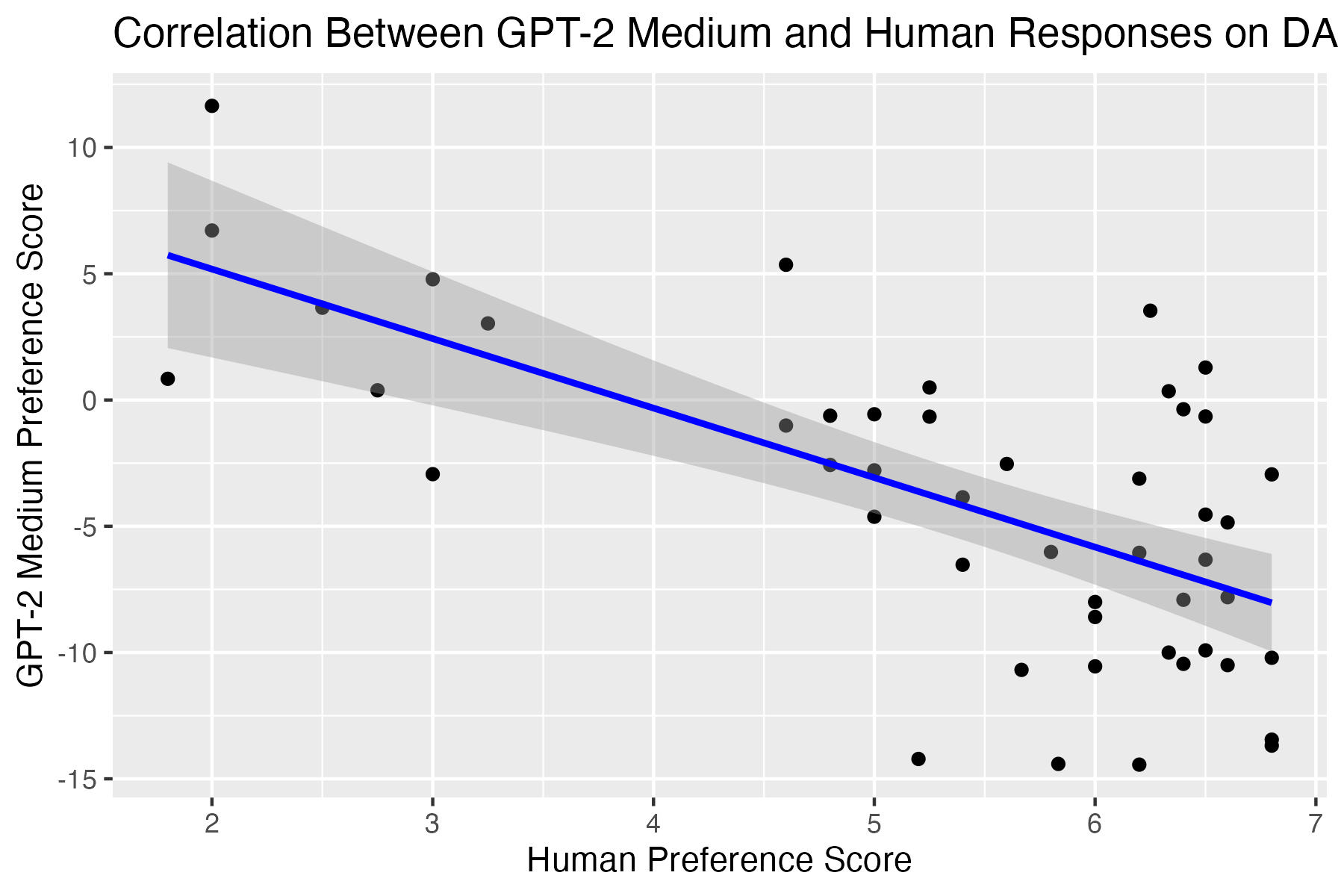}
\end{figure}
\begin{figure}[ht]
    \centering
    \includegraphics[width=\columnwidth]{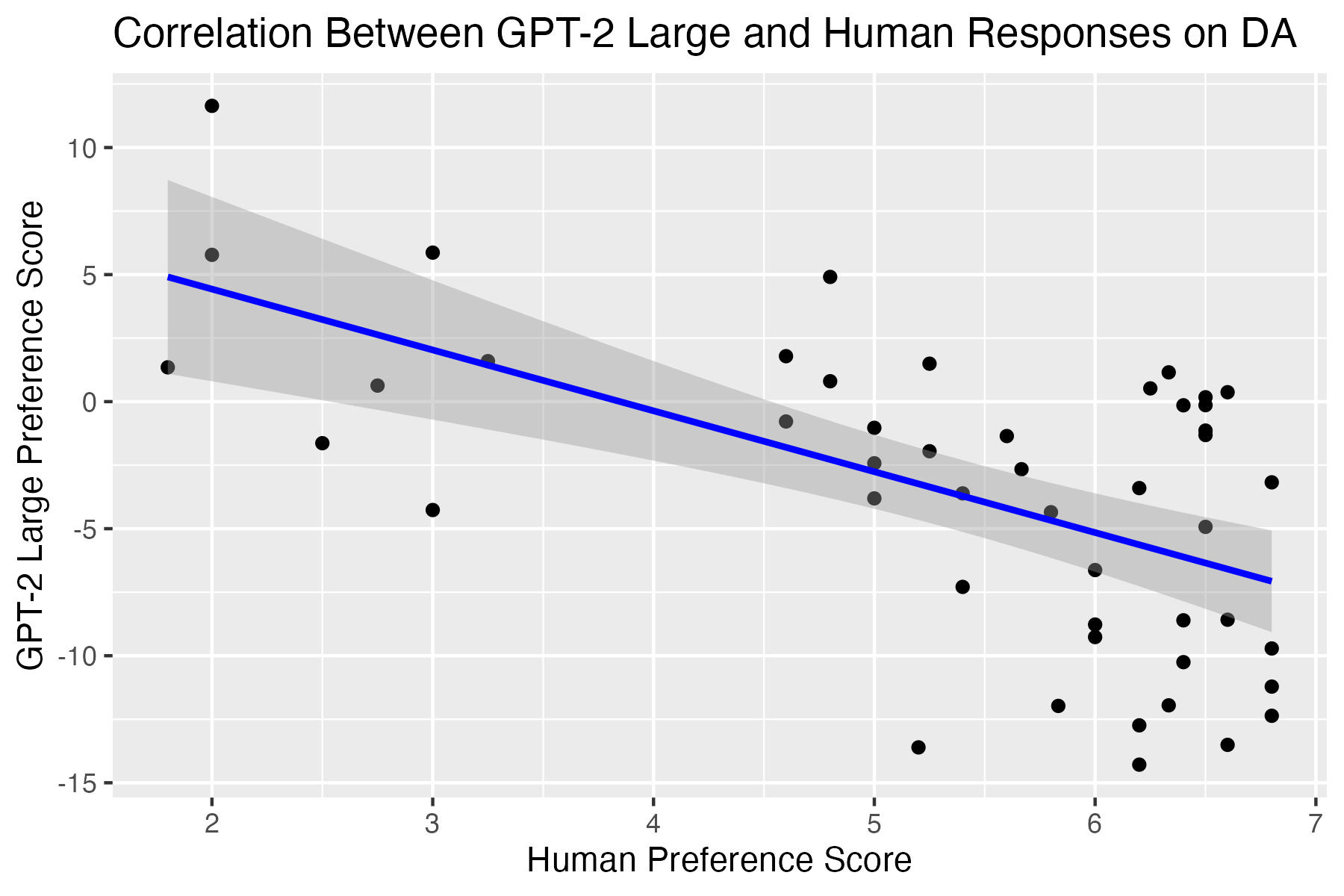}
\end{figure}
\begin{figure}[ht]
    \centering
    \includegraphics[width=\columnwidth]{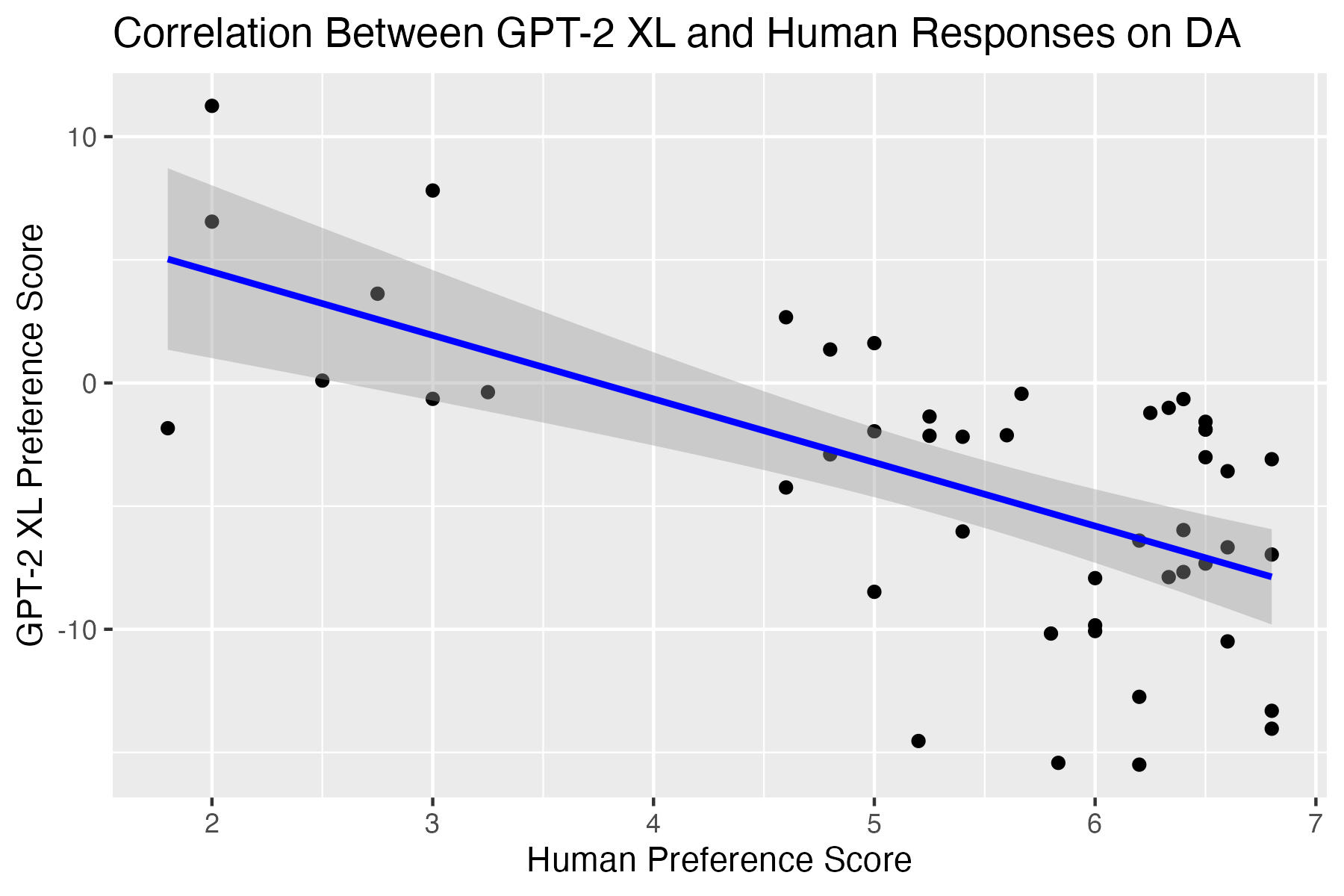}
\end{figure}
\begin{figure}[ht]
    \centering
    \includegraphics[width=\columnwidth]{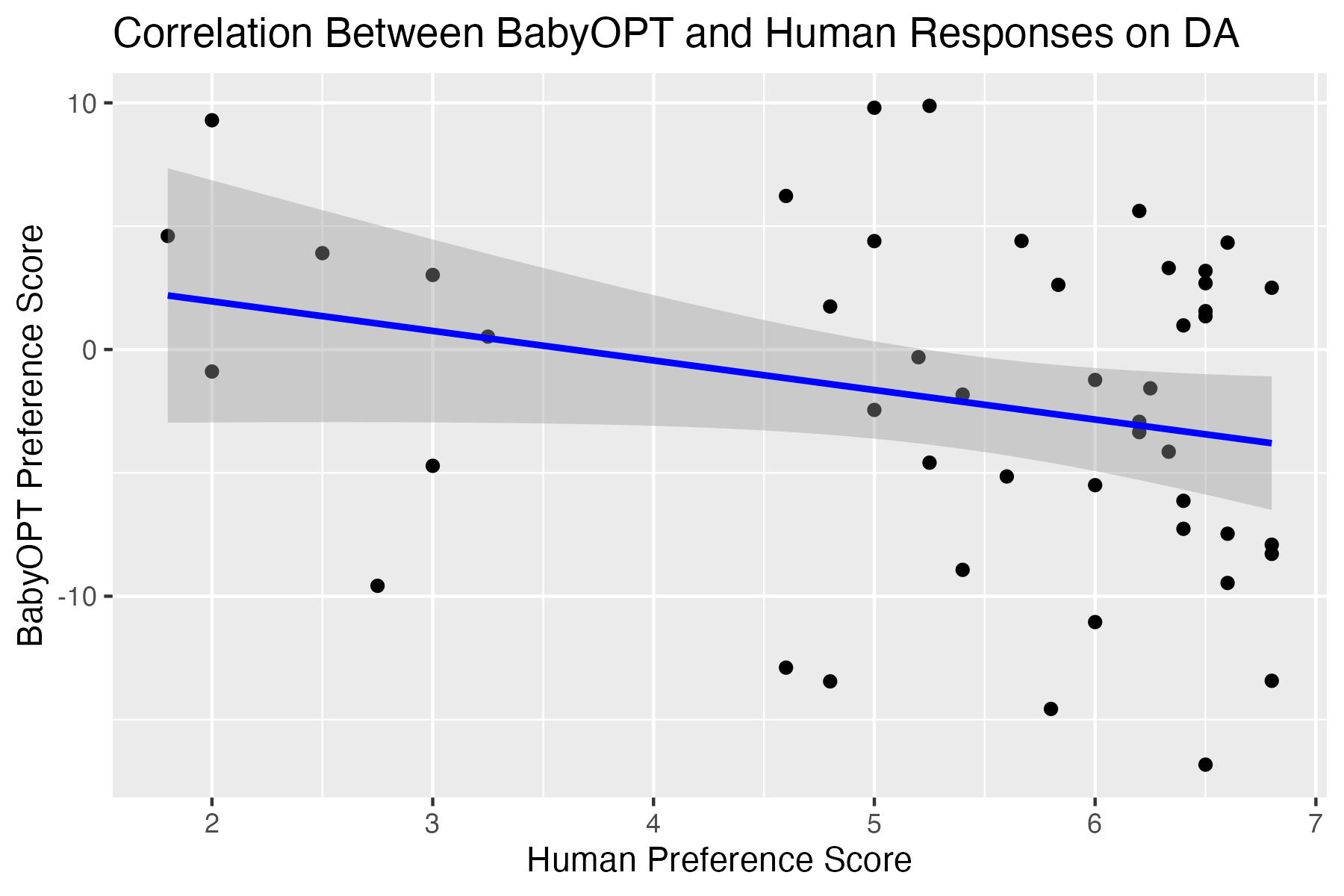}
\end{figure}
\begin{figure}[ht]
    \centering
    \includegraphics[width=\columnwidth]{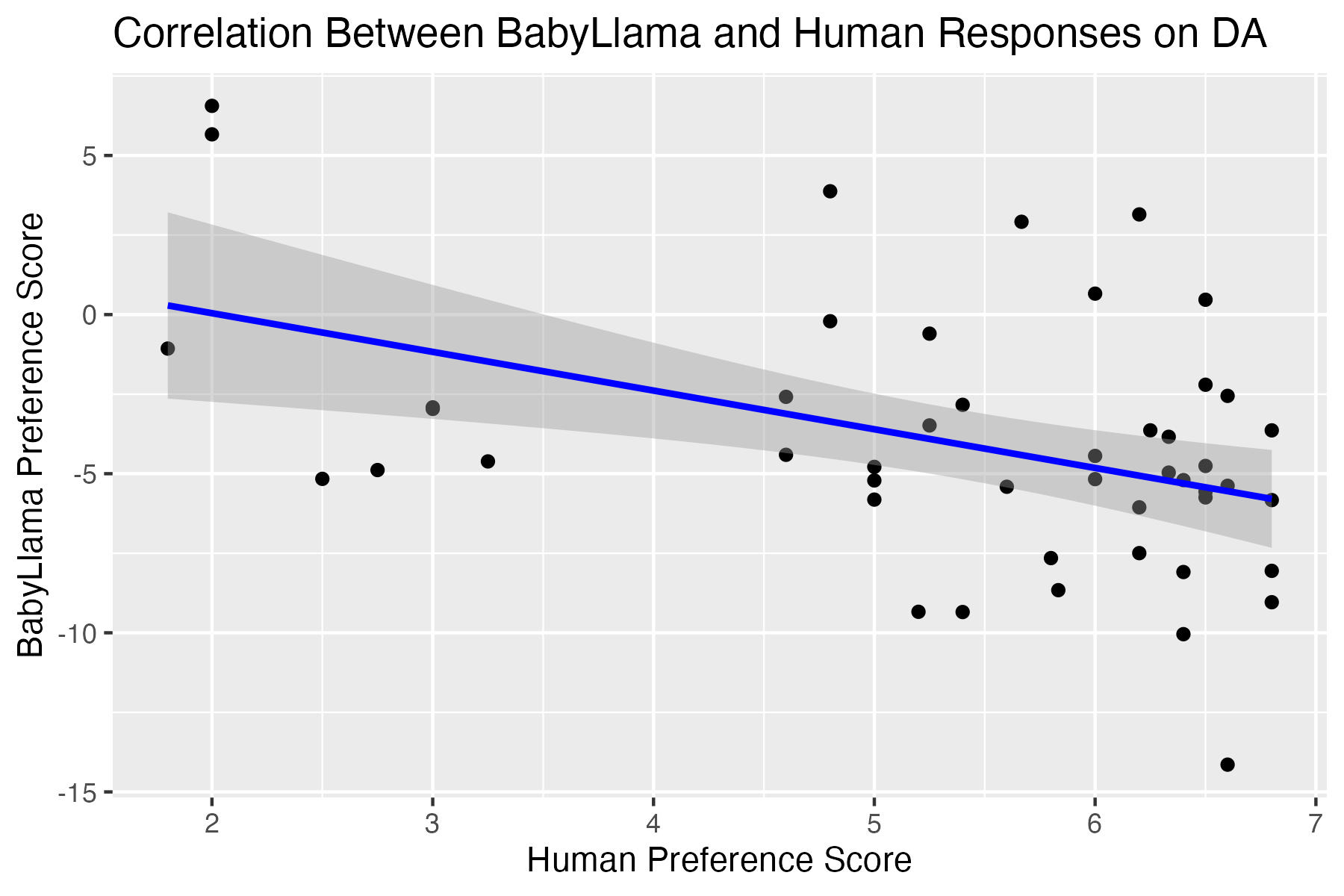}
\end{figure}
\begin{figure}[ht]
    \centering
    \includegraphics[width=\columnwidth]{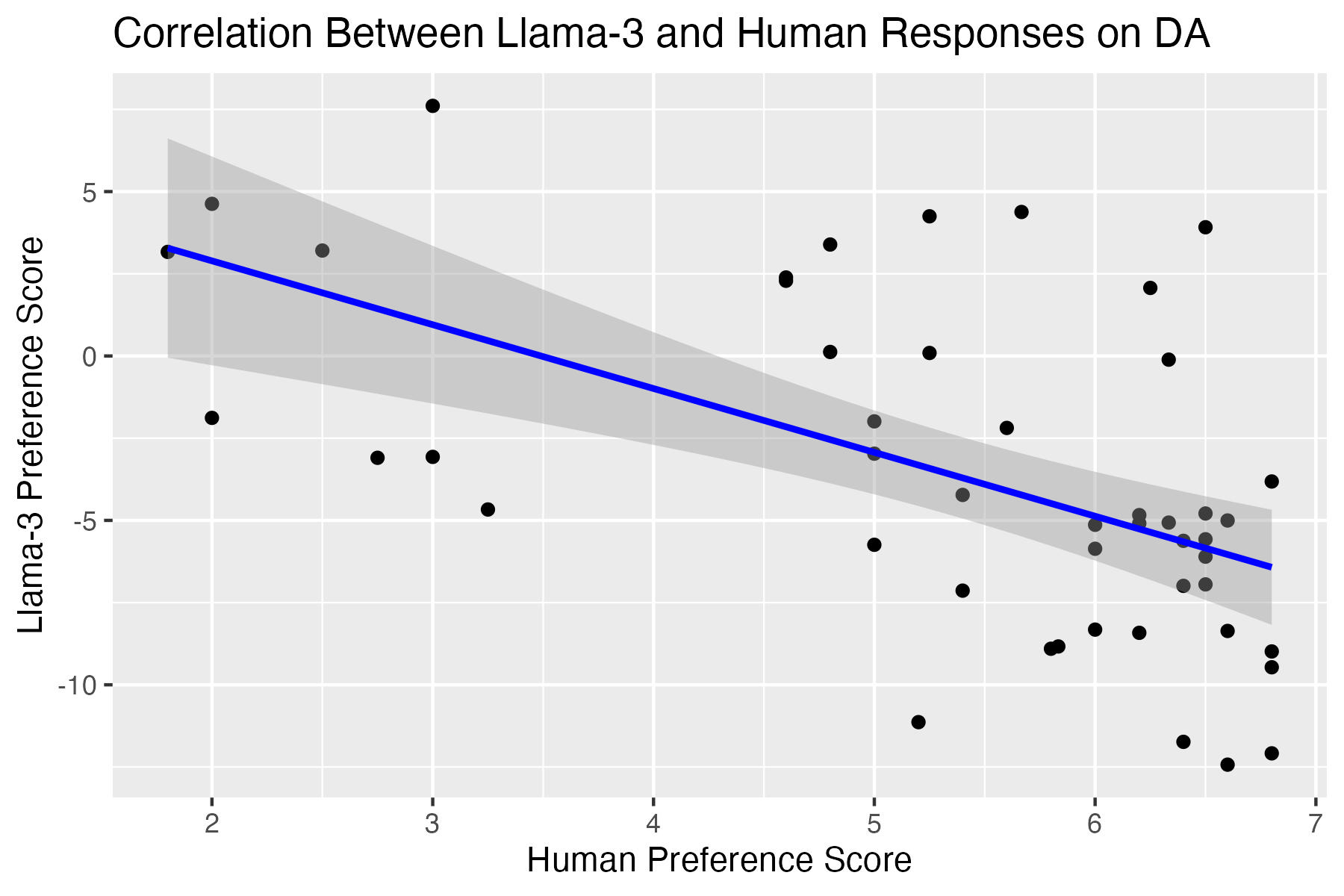}
\end{figure}
\begin{figure}[ht]
    \centering
    \includegraphics[width=\columnwidth]{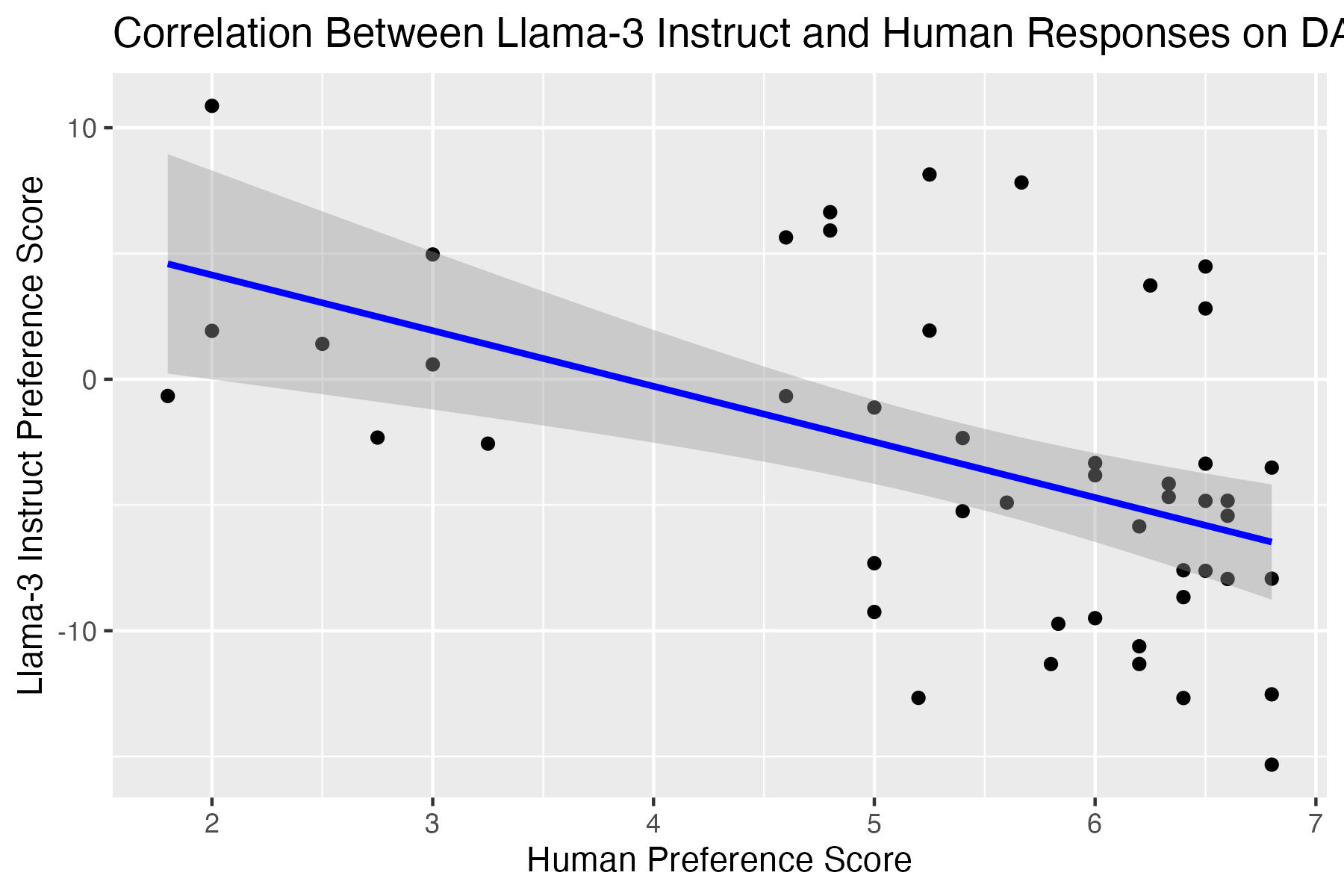}
\end{figure}
\begin{figure}[ht]
    \centering
    \includegraphics[width=\columnwidth]{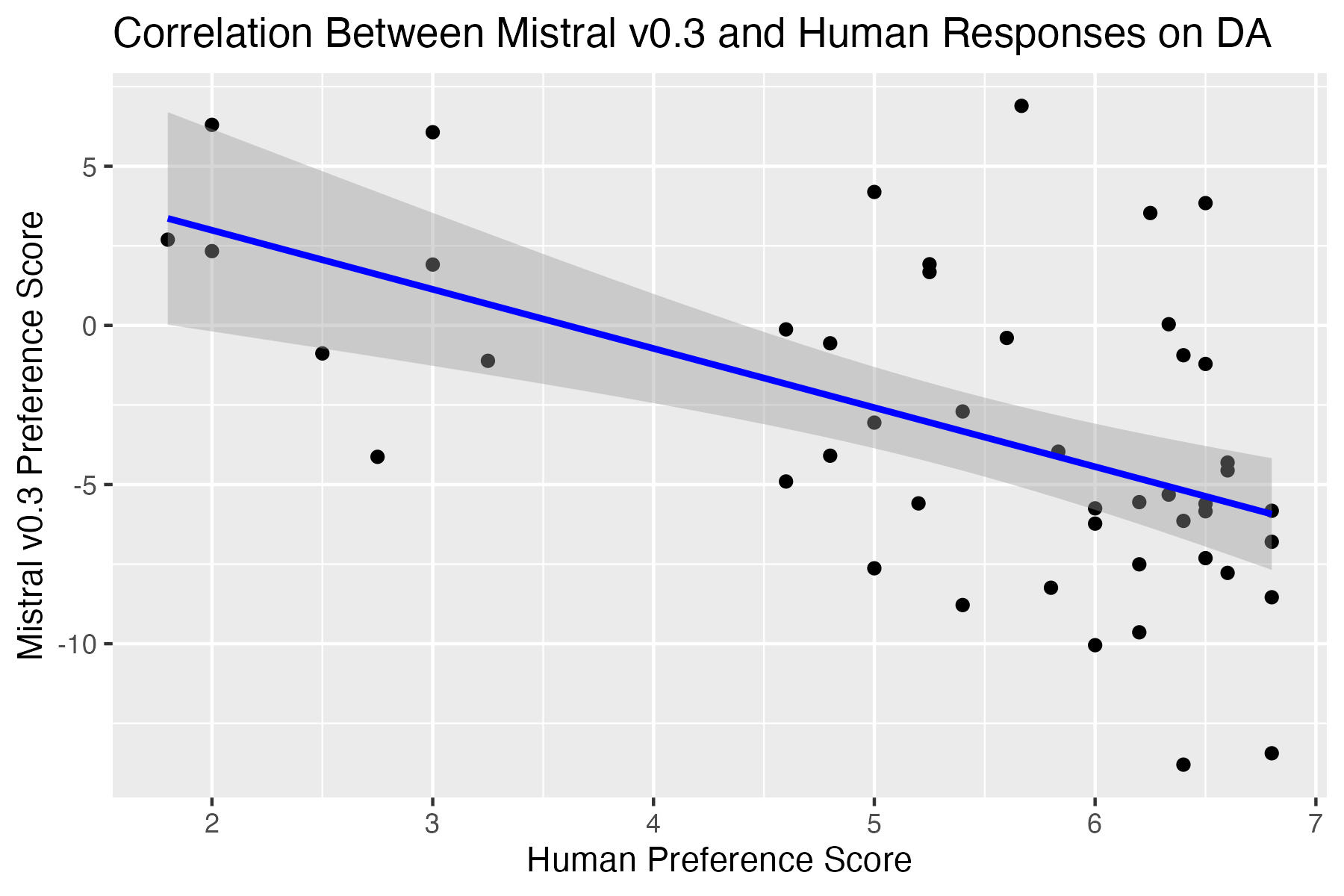}
\end{figure}
\begin{figure}[ht]
    \centering
    \includegraphics[width=\columnwidth]{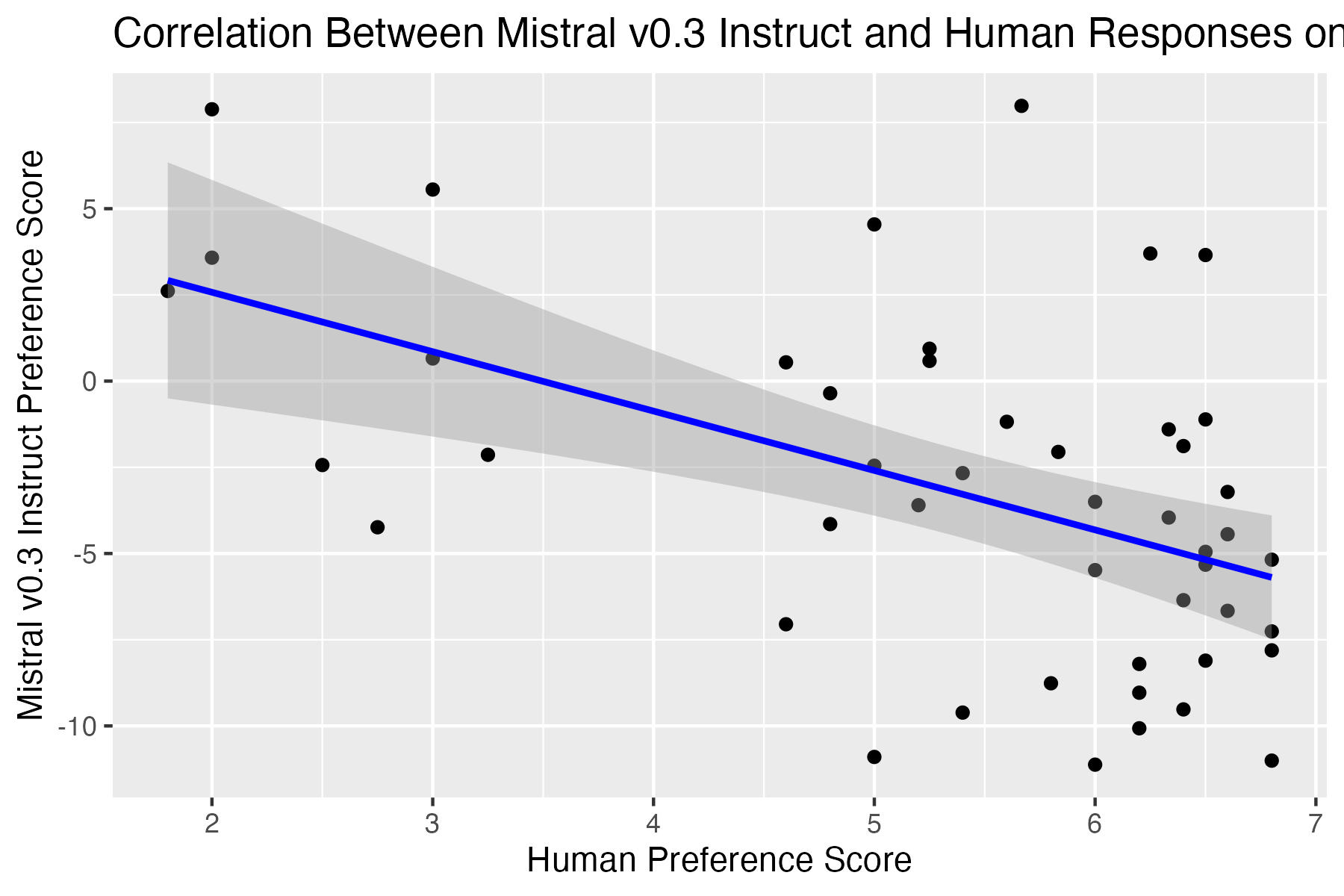}
\end{figure}
\begin{figure}[ht]
    \centering
    \includegraphics[width=\columnwidth]{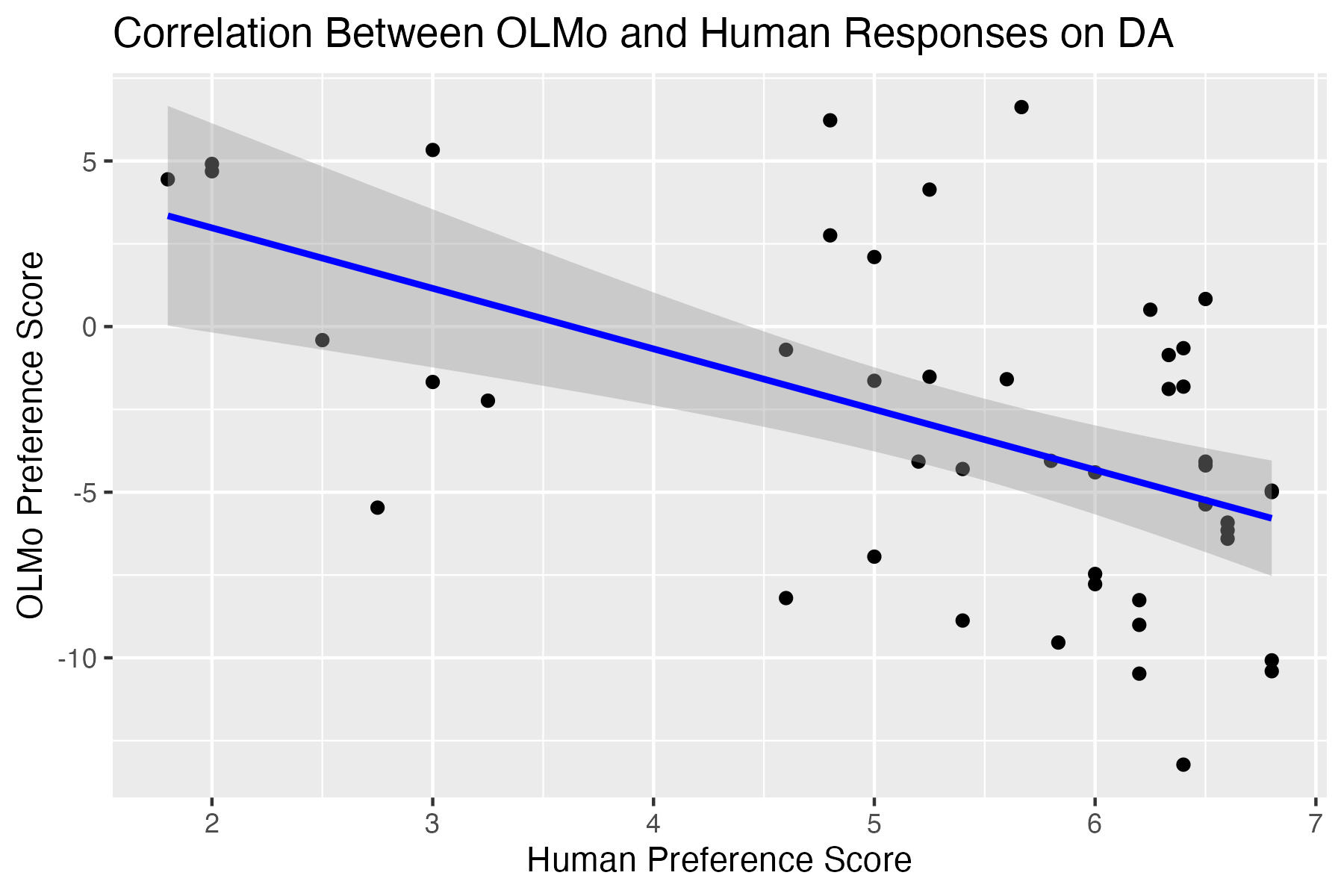}
\end{figure}
\begin{figure}[ht]
    \centering
    \includegraphics[width=\columnwidth]{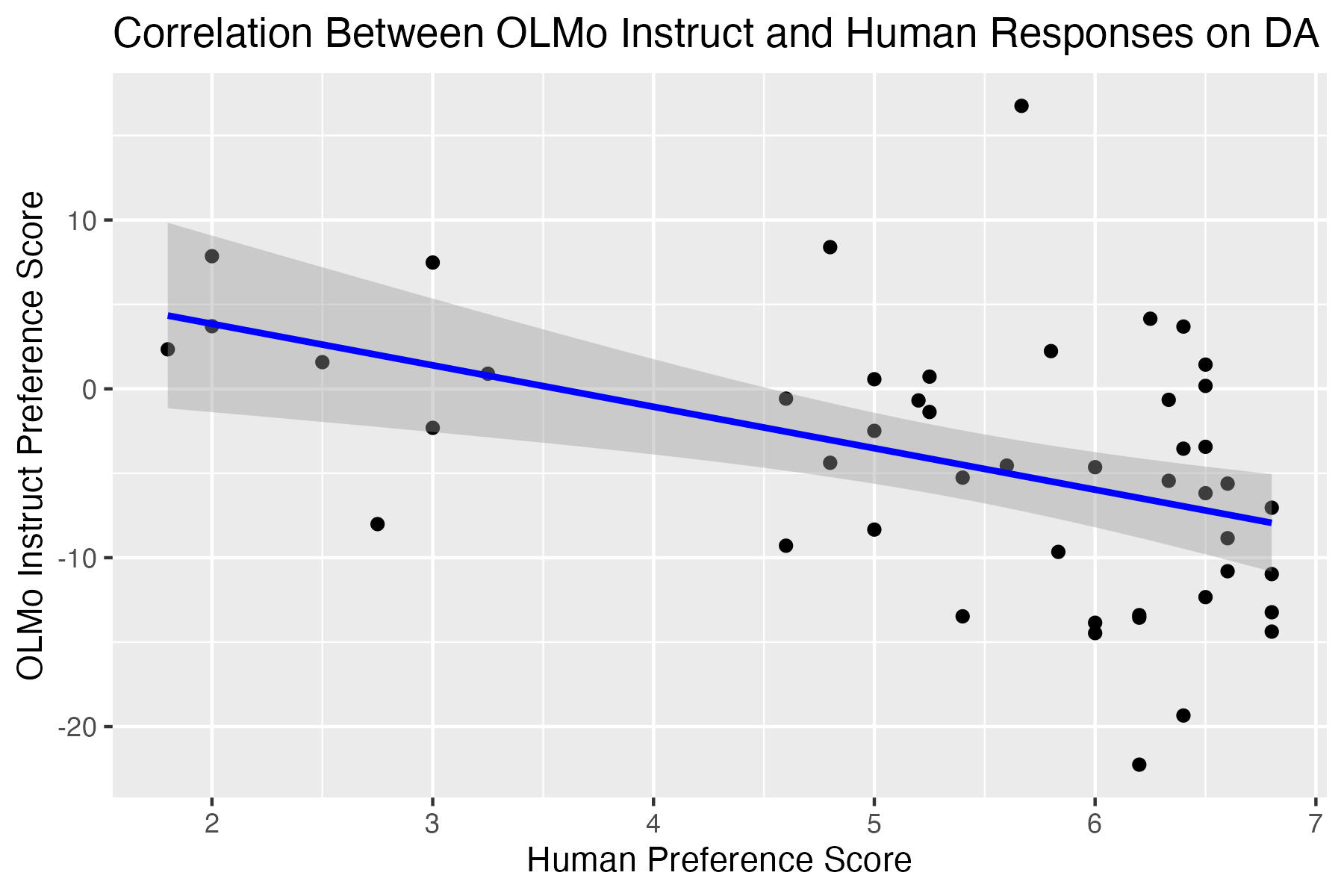}
\end{figure}
\begin{figure}[ht]
    \centering
    \includegraphics[width=\columnwidth]{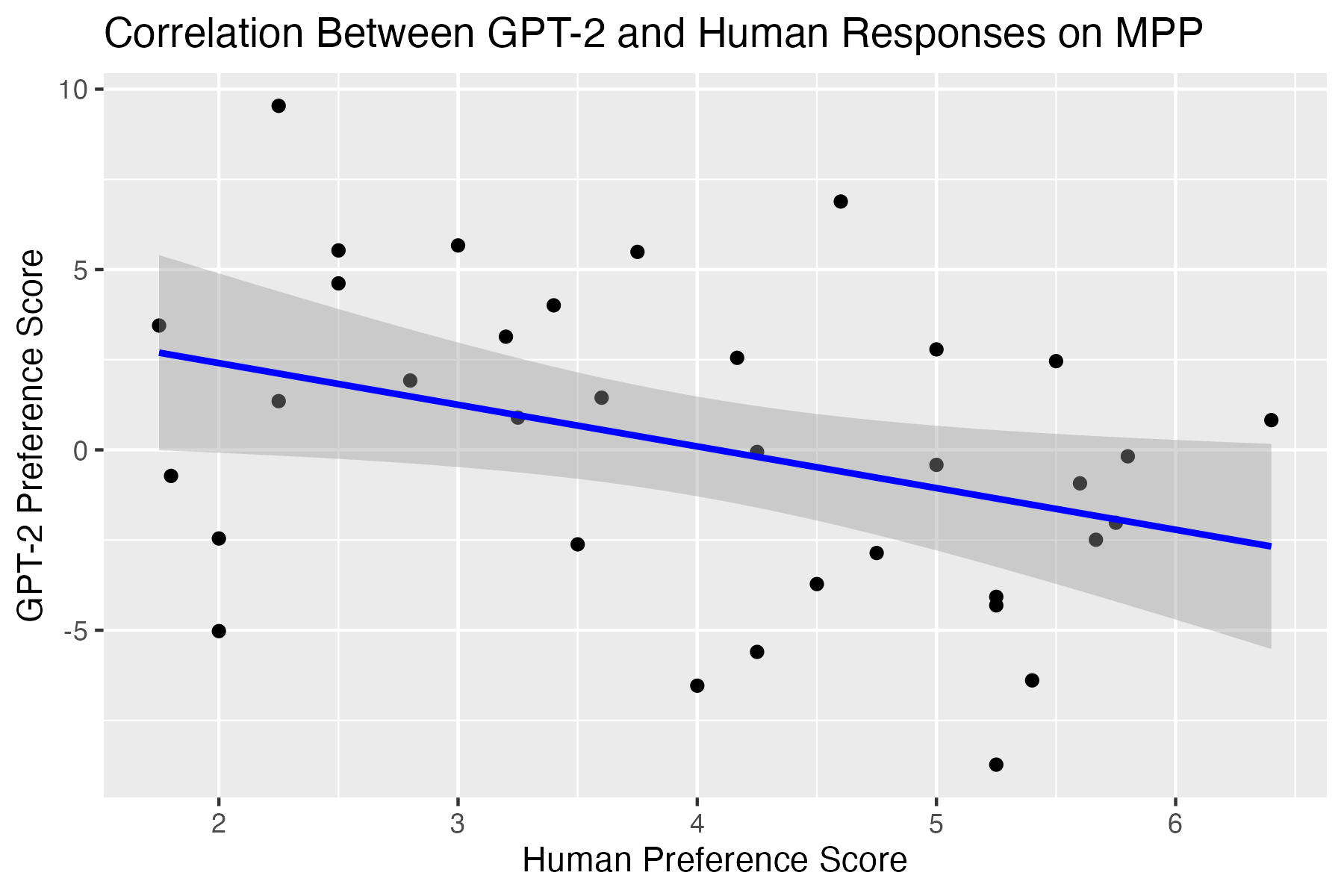}
\end{figure}
\begin{figure}[ht]
    \centering
    \includegraphics[width=\columnwidth]{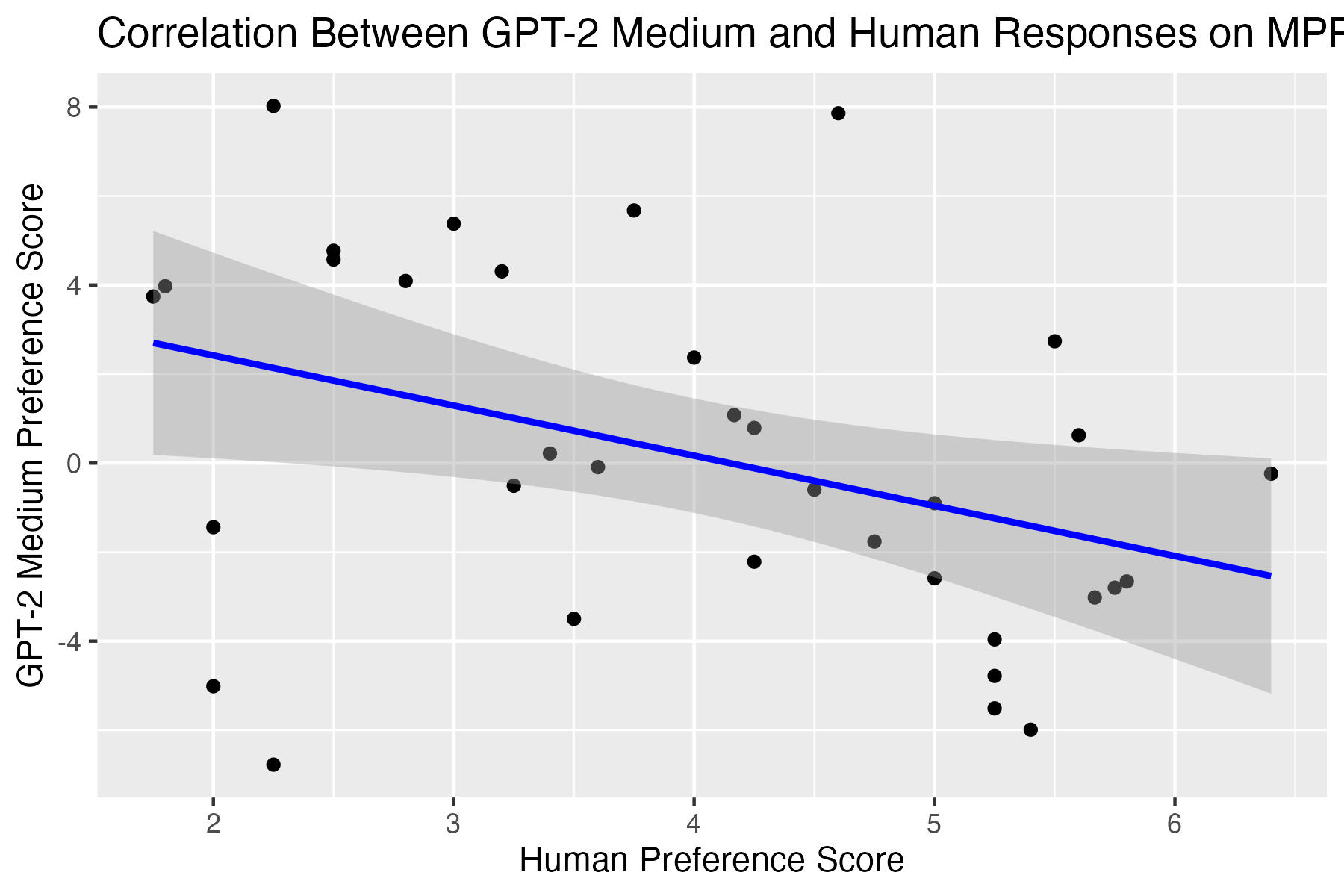}
\end{figure}
\begin{figure}[ht]
    \centering
    \includegraphics[width=\columnwidth]{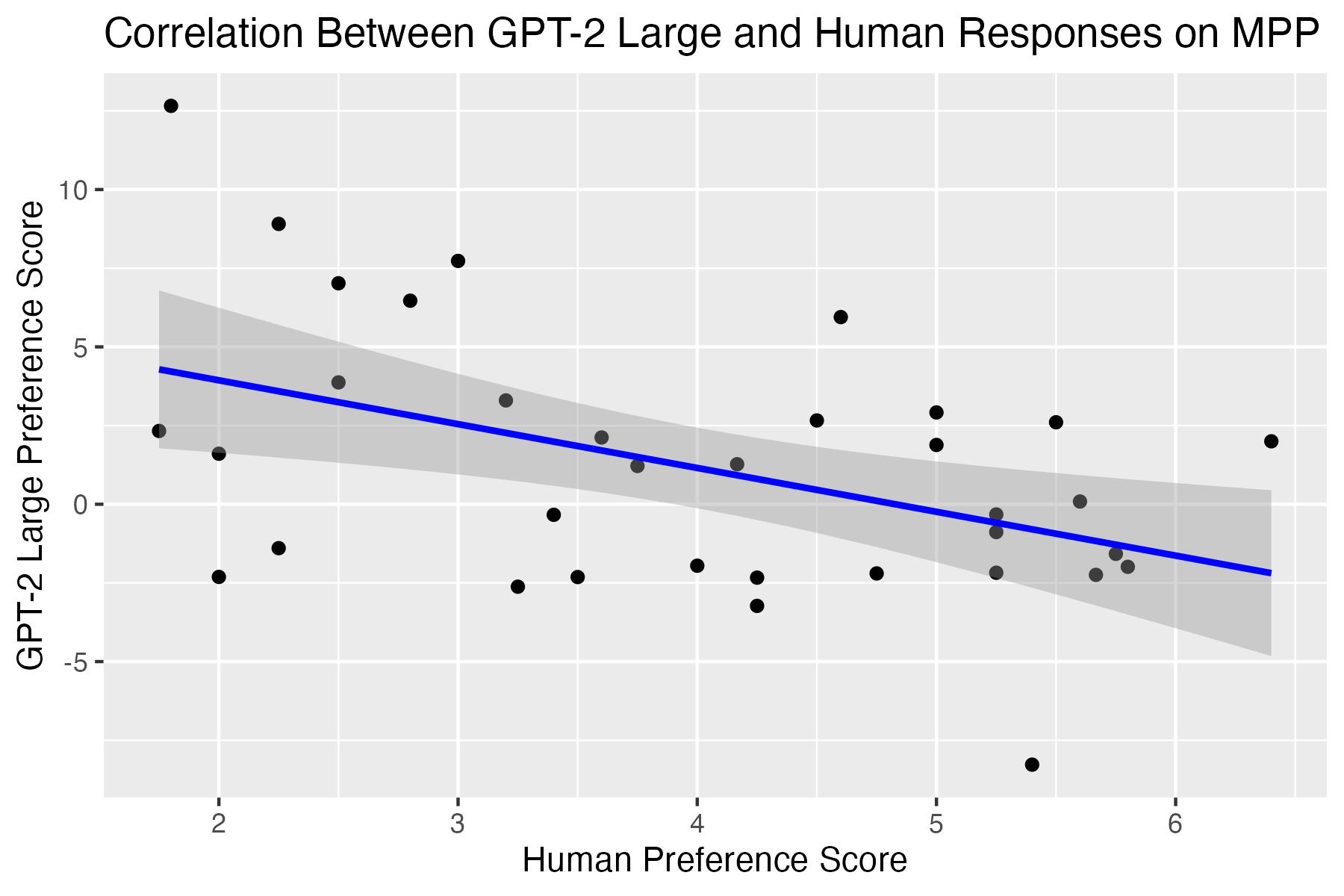}
\end{figure}
\begin{figure}[ht]
    \centering
    \includegraphics[width=\columnwidth]{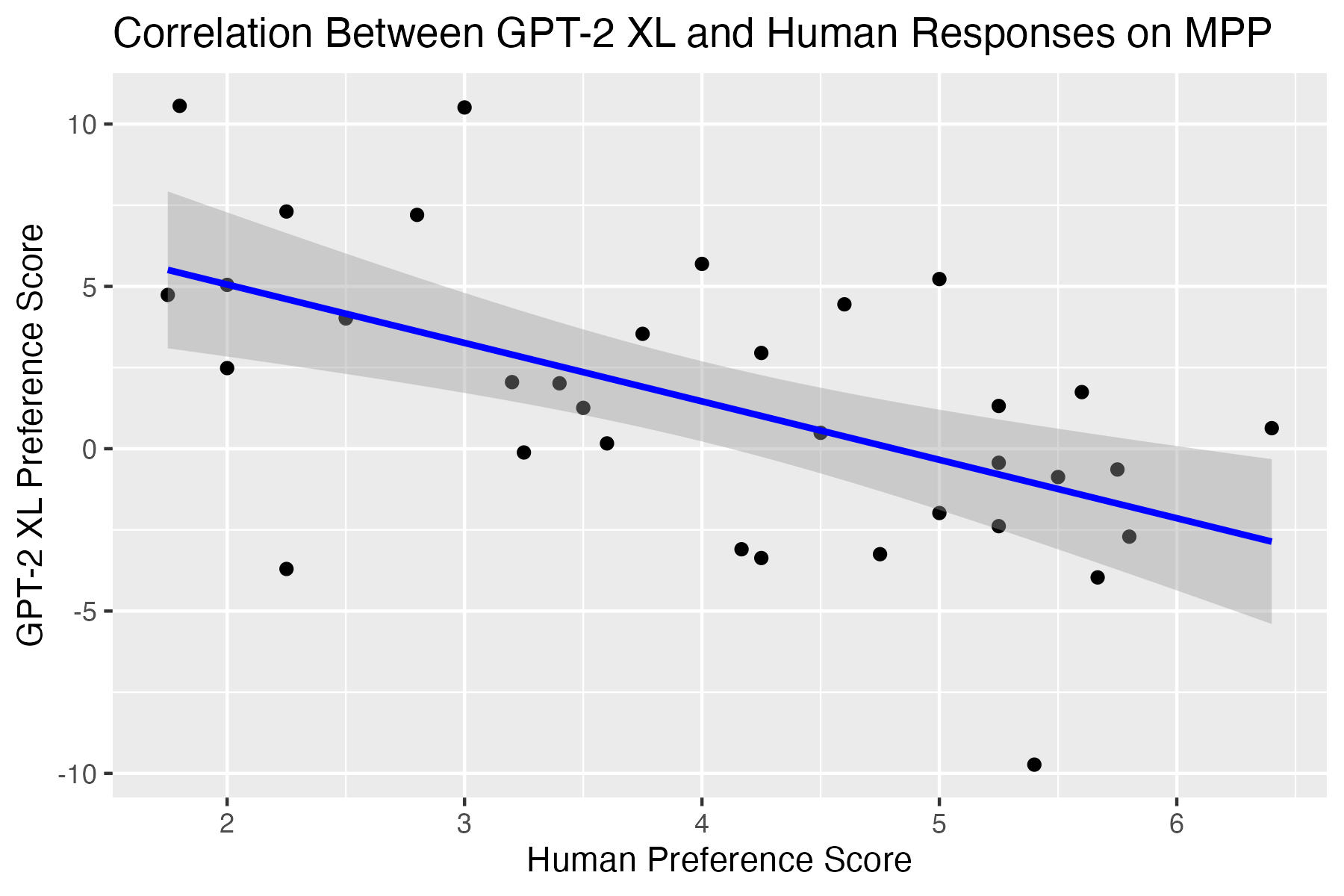}
\end{figure}
\begin{figure}[ht]
    \centering
    \includegraphics[width=\columnwidth]{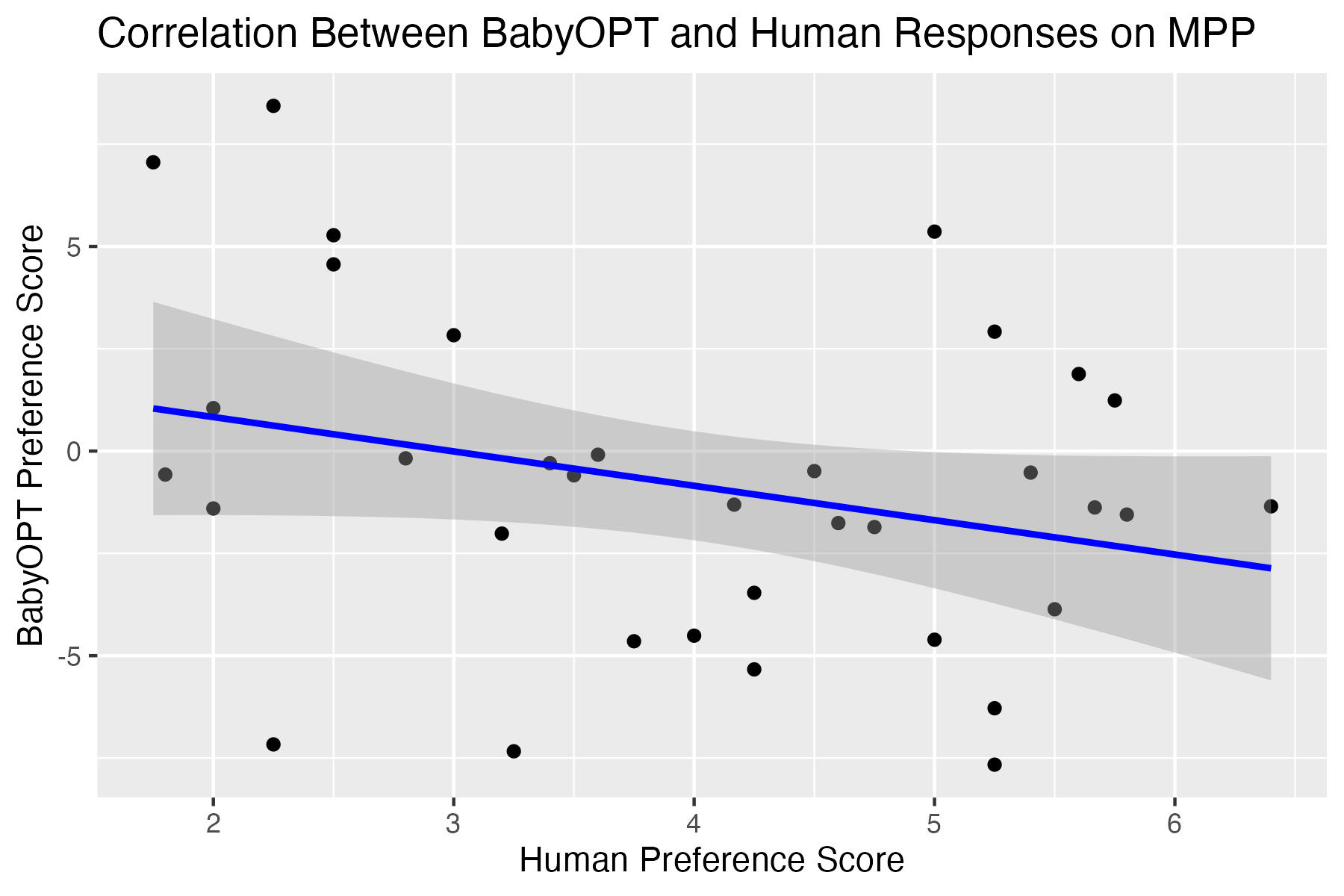}
\end{figure}
\begin{figure}[ht]
    \centering
    \includegraphics[width=\columnwidth]{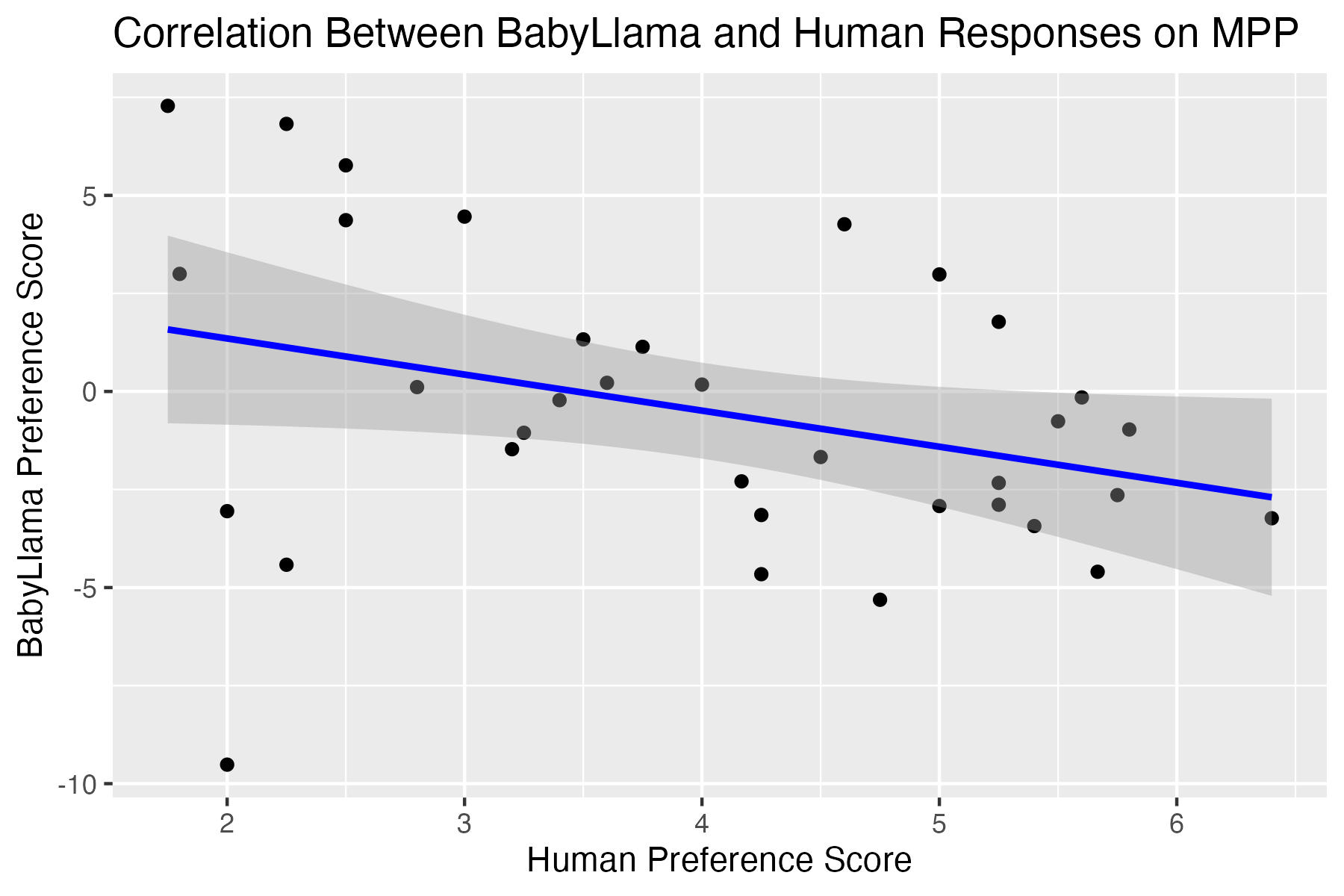}
\end{figure}
\begin{figure}[ht]
    \centering
    \includegraphics[width=\columnwidth]{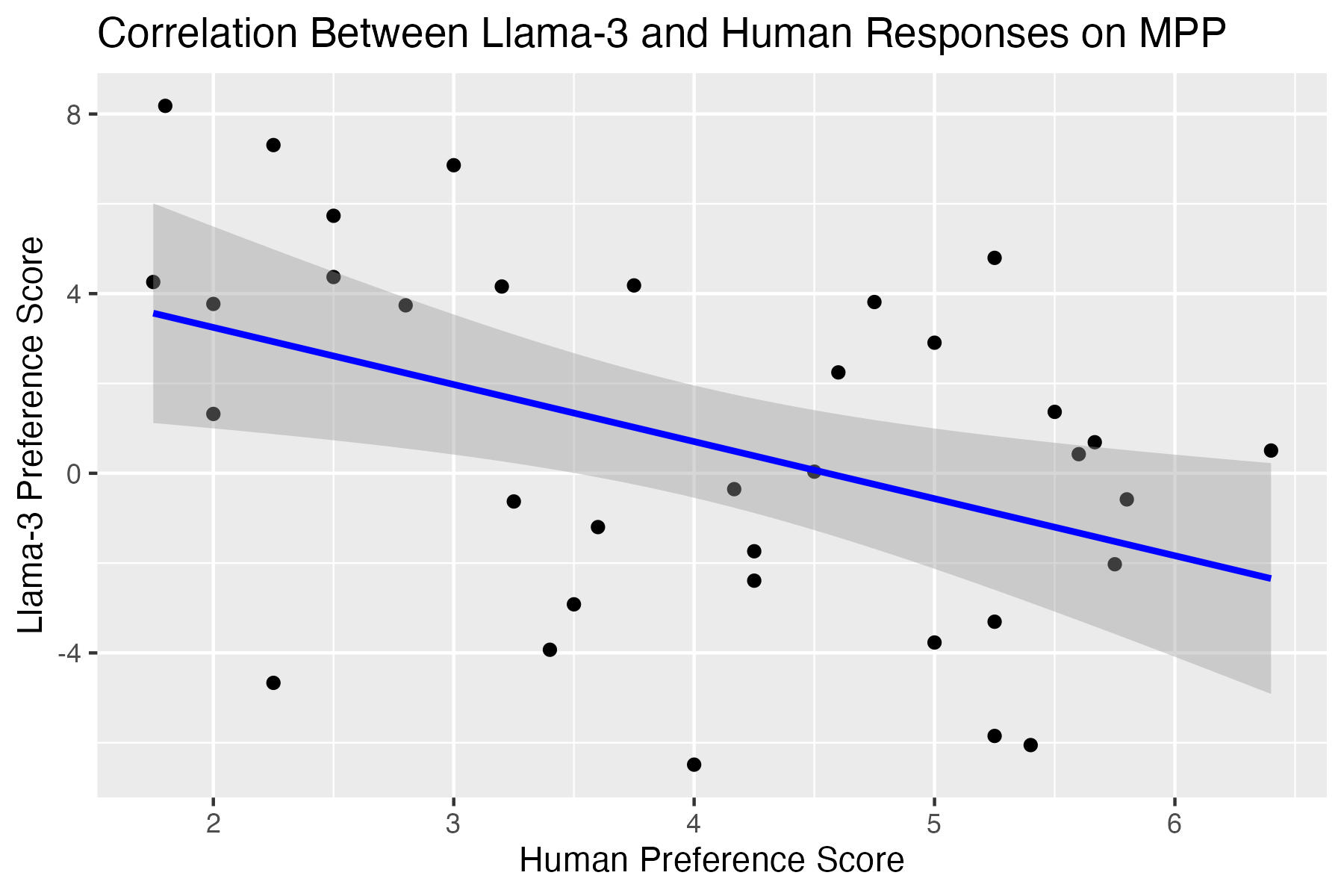}
\end{figure}
\begin{figure}[ht]
    \centering
    \includegraphics[width=\columnwidth]{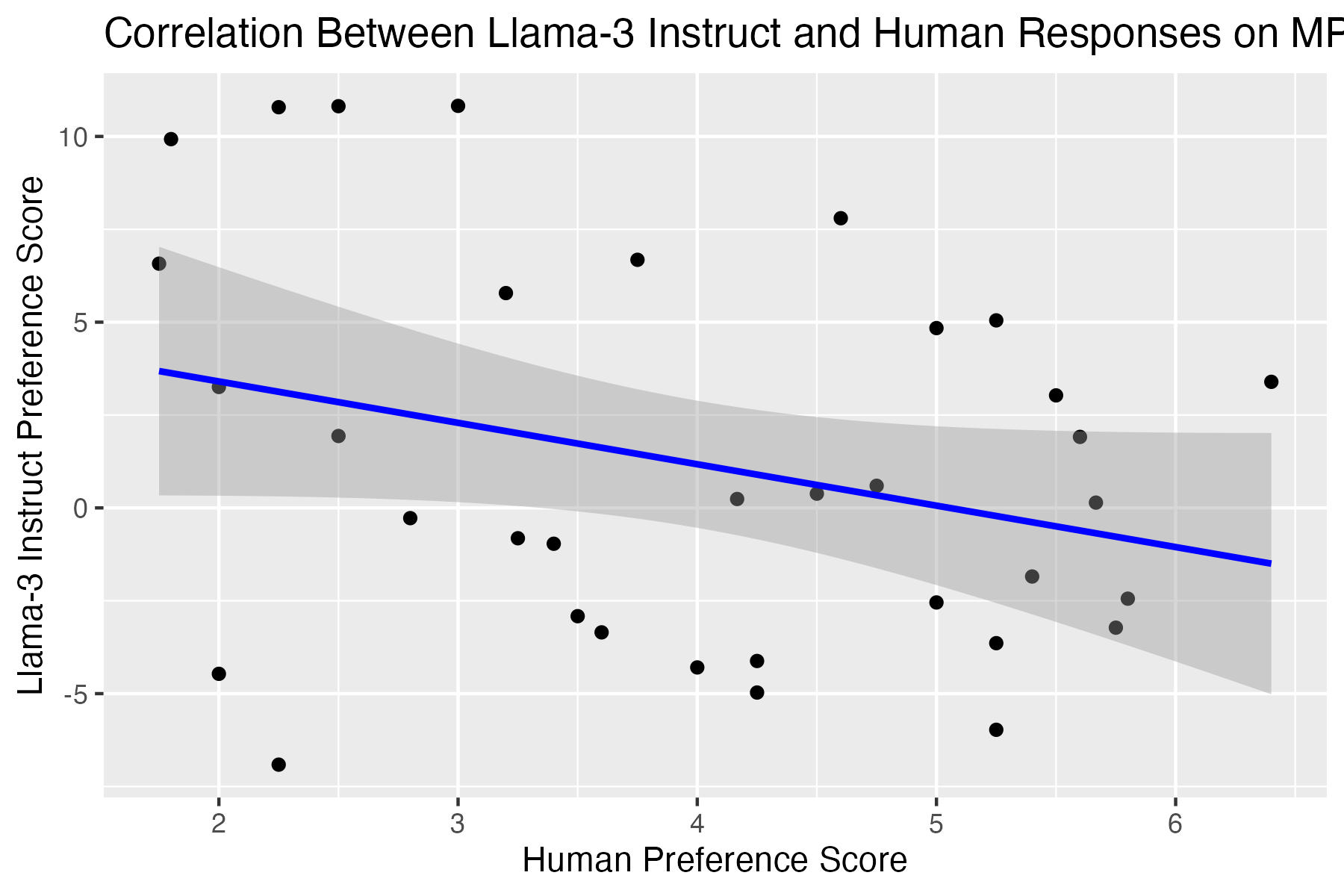}
\end{figure}
\begin{figure}[ht]
    \centering
    \includegraphics[width=\columnwidth]{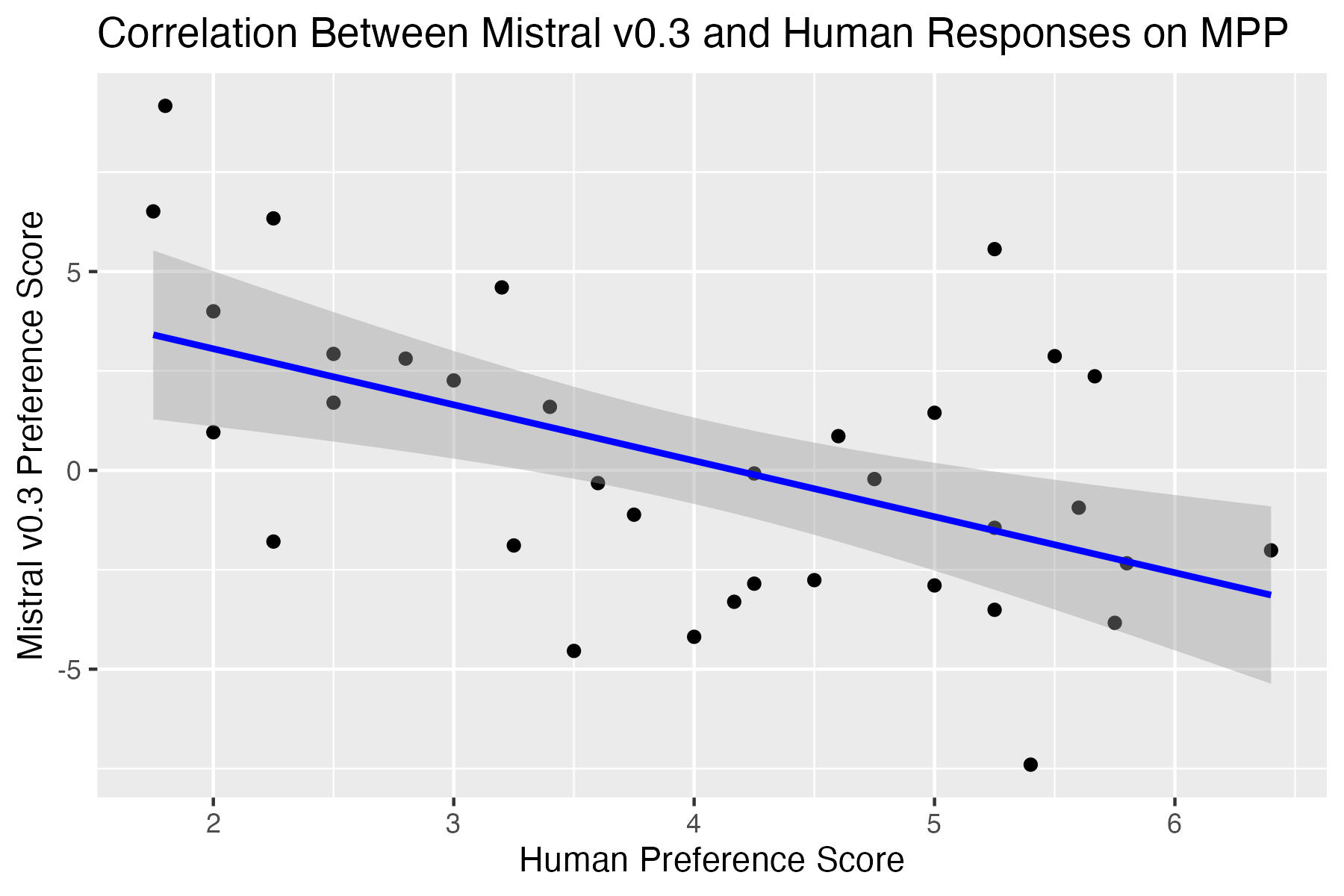}
\end{figure}
\begin{figure}[ht]
    \centering
    \includegraphics[width=\columnwidth]{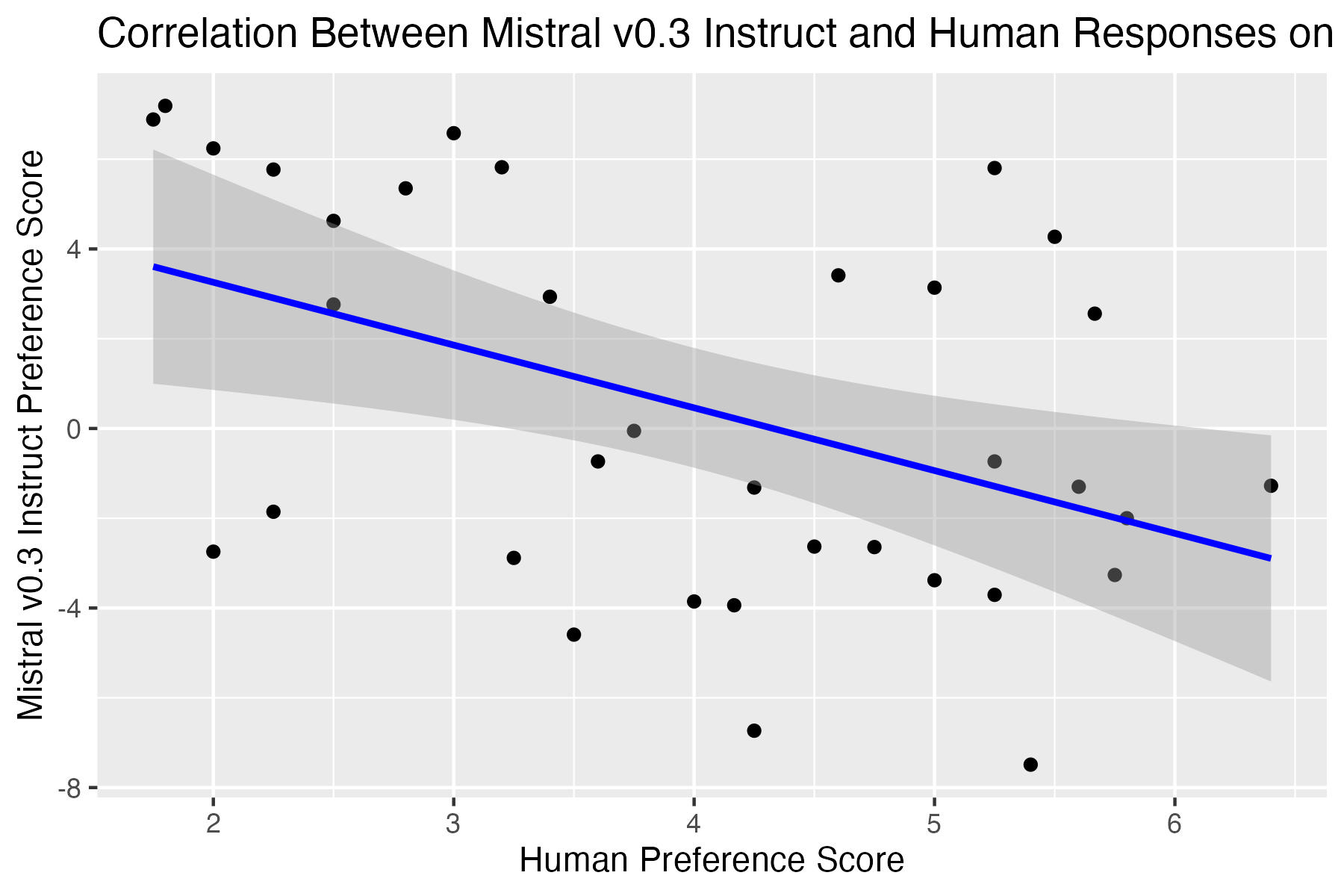}
\end{figure}
\begin{figure}[ht]
    \centering
    \includegraphics[width=\columnwidth]{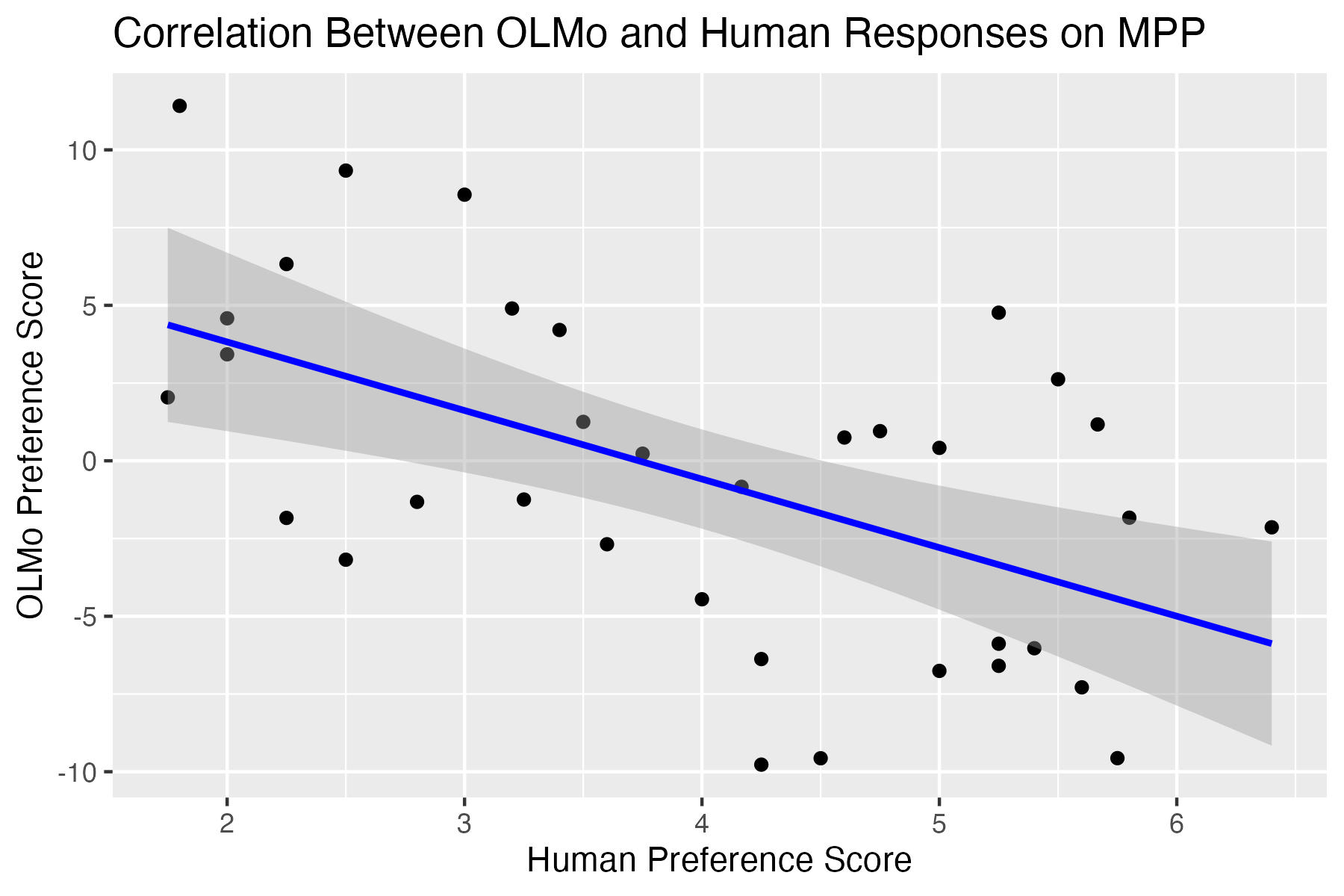}
\end{figure}
\begin{figure}[ht]
    \centering
    \includegraphics[width=\columnwidth]{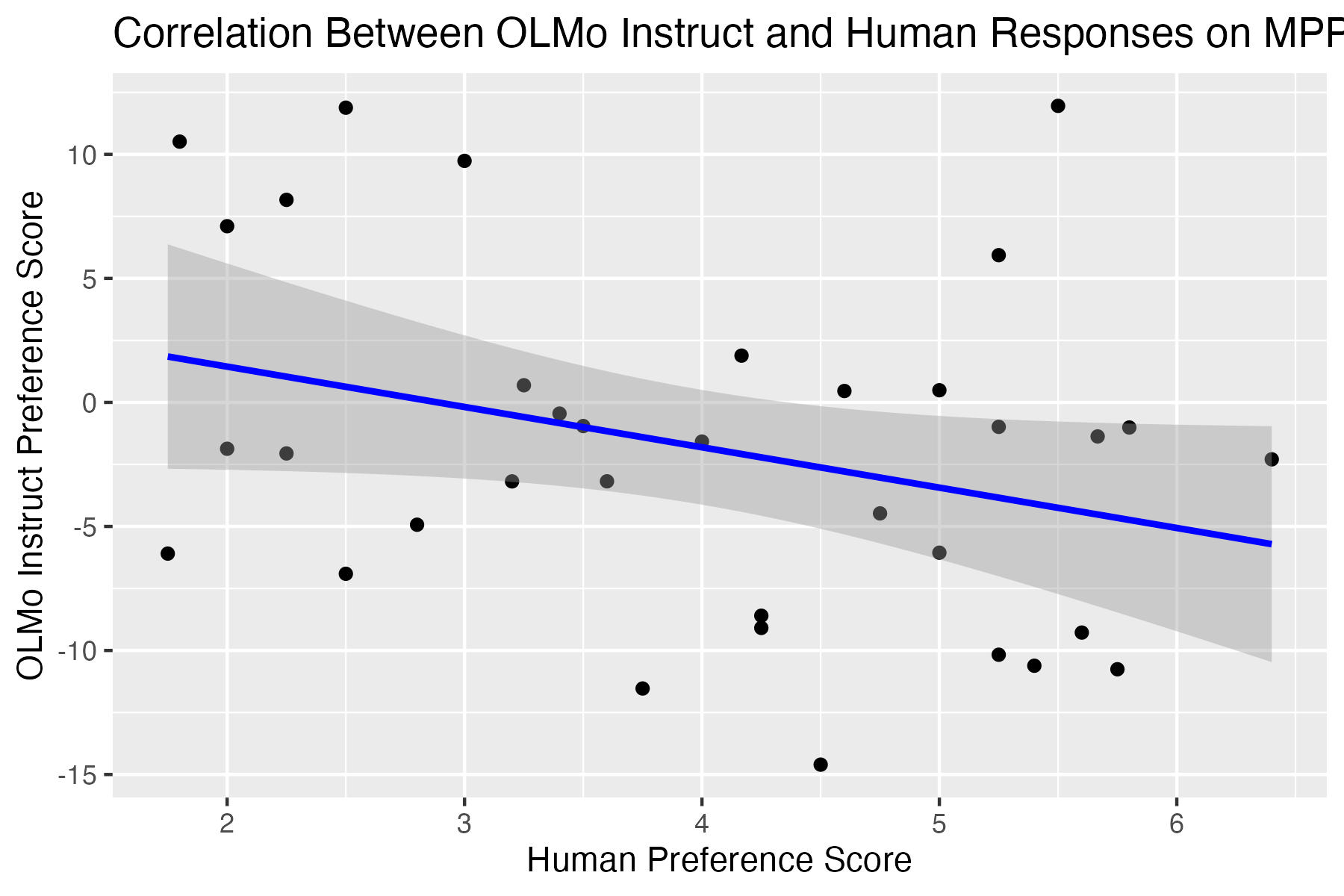}
\end{figure}

\end{document}